\documentclass{article}

\usepackage{microtype}
\usepackage{graphicx}
\usepackage{subfigure}
\usepackage{booktabs} % for professional tables

\usepackage{subfigure}
\usepackage{algorithmic}
\usepackage{algorithm}
\usepackage{graphicx}
\usepackage{subfigure}

\usepackage{xcolor}   % For coloring cells
\usepackage{caption}
\usepackage{booktabs} % For better looking tables
\usepackage{array} % 用于改进表格和数组

\usepackage{hyperref}

\usepackage[accepted]{icml2024}

\usepackage{amsmath}
\usepackage{amssymb}
\usepackage{mathtools}
\usepackage{amsthm}

\usepackage[capitalize,noabbrev]{cleveref}

\theoremstyle{plain}

\theoremstyle{definition}

\theoremstyle{remark}

\usepackage[textsize=tiny]{todonotes}

\icmltitlerunning{Understanding Time Series Anomaly State Detection through One-Class Classification}

\begin{document}

\twocolumn[
\icmltitle{Understanding Time Series Anomaly State Detection through One-Class Classification}



\icmlsetsymbol{equal}{*}

\begin{icmlauthorlist}
\icmlauthor{Hanxu Zhou}{equal,sjtu}
\icmlauthor{Yuan Zhang}{equal,sjtu}
\icmlauthor{Guangjie Leng}{equal,sjtu}
\icmlauthor{Ruofan Wang}{fudan}
\icmlauthor{Zhi-Qin John Xu}{sjtu}
\end{icmlauthorlist}

\icmlaffiliation{sjtu}{School of Mathematical Sciences, Institute of Natural Sciences, MOE-LSC
 and Qing Yuan Research Institute, Shanghai Jiao Tong University, Shanghai}
\icmlaffiliation{fudan}{School of Computer Science, Shanghai Key Laboratory of Intelligent Information Processing, Fudan University, Shanghai}

\icmlcorrespondingauthor{Hanxu Zhou}{zhouhanxu@sjtu.edu.cn}
\icmlcorrespondingauthor{Zhi-Qin John Xu}{xuzhiqin@sjtu.edu.cn}

\icmlkeywords{Machine Learning, Time Series Analysis, Anomaly Detection, One-Class Classification}

\vskip 0.3in
]



\printAffiliationsAndNotice{\icmlEqualContribution} % otherwise use the standard text.

\begin{abstract}
For a long time, research on time series anomaly detection has mainly focused on finding outliers within a given time series. Admittedly, this is consistent with some practical problems, but in other practical application scenarios, people are concerned about: assuming a standard time series is given, how to judge whether another test time series deviates from the standard time series, which is more similar to the problem discussed in one-class classification (OCC).
Therefore, in this article, we try to re-understand and define the time series anomaly detection problem through OCC, which we call 'time series anomaly state detection problem'.
We first use stochastic processes and hypothesis testing to strictly define the 'time series anomaly state detection problem', and its corresponding anomalies. Then, we use the time series classification dataset to construct an artificial dataset corresponding to the problem. We compile 38 anomaly detection algorithms and correct some of the algorithms to adapt to handle this problem. 
Finally, through a large number of experiments, we fairly compare the actual performance of various time series anomaly detection algorithms, providing insights and directions for future research by researchers. 
\end{abstract}

\section{Introduction}

%%%%% 背景介绍

Time series anomaly detection emerges as an indispensable technique across diverse sectors, playing a pivotal role in identifying deviations from established normative patterns. This method finds extensive application in areas such as financial markets, network security, industrial process monitoring, and more~\cite{hilal2022financial, kaur2017review, park2020review, fahim2019anomaly}. Leveraging the wealth of data produced by these industries for real-time surveillance of anomalous events can significantly mitigate potential losses in human, environment, and energy resources~\cite{fernando2021deep, russo2020active, jesus2021using, himeur2021artificial}. 
% For instance, within the realm of financial markets, anomaly detection is instrumental in exposing illicit activities, including market manipulation and fraud, by identifying atypical trading behaviors~\cite{hilal2022financial}. In the domain of network security, it is essential for the detection of threats like Distributed Denial of Service (DDoS) attacks~\cite{kaur2017review}, while in industrial monitoring, it facilitates the tracking of machine performance and forecasts equipment failures, thereby minimizing operational downtime \cite{park2020review, fahim2019anomaly}. In the healthcare sector, this technique assists in the early diagnosis of diseases through the analysis of patient physiological data \cite{fernando2021deep}. In the context of environmental monitoring, it is employed to detect anomalies indicative of pollution events or natural disasters \cite{russo2020active, jesus2021using}. Lastly, within energy management, anomaly detection contributes to enhanced efficiency by identifying irregular consumption patterns \cite{himeur2021artificial}.

%%%%% 算法与benchmark现状。

Given the wide application and importance of time series anomaly detection, it has received widespread academic and industrial attention and developed rapidly from six decades ago~\cite{fox1972outliers,page1957problems,tsay1988outliers,tukey1977exploratory,chang1988estimation}. 
% Pioneering efforts can be traced back to 1979, when \cite{tukey1977exploratory} introduced statistical methods for anomaly detection in time series, which was further refined by \cite{chang1988estimation}, who advocated the use of the Likelihood Ratio Test (LRT) for identifying anomalies in time series data.
Besides, as machine learning becomes more popular in various data science tasks, researchers have begun to explore the use of machine learning methods in these tasks to detect anomalies in time series data. For example, classification algorithms such as support vector machines~\cite{hearst1998support} and time series forecasting algorithms such as ARIMA~\cite{box1976time} have proven useful in identifying anomalies in time series data.
In recent years, with the rise of neural network technology, a large number of emerging anomaly detection algorithms are based on neural network architecture, which is roughly applied in three aspects: feature extraction, learning feature representations of normality, and end-to-end anomaly score learning \cite{pang2021deep}. 
% The advantage of neural networks lies in their ability to learn expressive representations of complex time series, such as multidimensional data with spatial and temporal characteristics. 
% In time series anomaly detection, the application of neural networks can be roughly divided into three aspects: feature extraction, learning feature representations of normality, and end-to-end anomaly score learning \cite{pang2021deep}. 
% Some benchmarks and reviews have also compiled the above time series anomaly detection methods in detail \cite{braei2020anomaly, pang2021deep}. 

Most of the articles on anomaly detection algorithms mentioned above are concerned with identifying individual observations that are significantly deviated from other observations in the dataset, that is, outliers, and often called `Outlier Detection', `Outlier Analysis' or `Anomalous Pattern Detection'~\cite{das2008anomaly, kim2020anomaly}, which is an important task in time series anomaly detection and can be considered as a function $\phi_{\text{binary}}$ \cite{braei2020anomaly},
\begin{equation}
\begin{aligned}
& \phi_{\text {binary}}: \mathbb{R}^{\omega} \rightarrow\{\text {normal, anomaly}\} \\
& \phi_{\text {binary}}(x) \mapsto 
\begin{cases}\text {anomaly, } & \text { if } \phi_{\text {score }}(x) = \gamma > \delta \\
\text {normal, } & \text { otherwise }
\end{cases}
\end{aligned}
\end{equation}
where $\phi_{\text{score}}$ is an anomaly detection algorithm, $\gamma$ is the anomaly score, $\delta \in \mathbb{R}$ is the threshold to determine whether a point is anomaly, $x \in X \subseteq \mathbb{R}^{\omega}$, $X$ is the dataset, and $\omega$ is the width of the sliding window. The objective of $\phi_{\text{binary}}$ is to detect rare data points that deviate remarkably from the general distribution of the dataset~\cite{braei2020anomaly}.

% Besides the above-mentioned "Outlier Detection", "Time series Change Point Detection" \cite{aminikhanghahi2017survey} is another notable sub-problems of time series anomaly detection with practical significance. This technique focuses on the identification of structural change points within time series data, thereby facilitating the recognition of potentially significant events, alterations in systemic behavior, or shifts in prevailing trends. 
% The utility of time series change point detection extends across a diverse array of fields, encompassing applications such as financial market analysis, monitoring of industrial production processes, and the study of climatic transformations \cite{reeves2007review}, among others.

However, in addition to the `Outlier Detection' discussed above, there's another kind that's widely used in industrial manufacturing but hasn't received much attention, which can be briefly described as follows: given a standard time series, how to determine if another test time series deviates from this standard time series? Here we temporarily call this problem the `Time Series Anomaly State Detection Problem'. It's a way of looking at time series anomaly detection from a one-class classification perspective.

The `time series anomaly state detection' has a wide range of practical applications. For example, in the field of mechanical equipment inspection, it is used to systematically monitor and evaluate the operating status of mechanical systems to ensure their optimal operation. Similarly, within the scope of power system monitoring, it plays a key role in the identification and analysis of grid anomalies, including harmonic distortion, voltage fluctuations and frequency irregularities, which indicates potential system issues or instability within the grid infrastructure. Furthermore, in the domain of wireless communications, it is pivotal for the surveillance of signal interference and anomalies in signal strength, a key aspect in ensuring the reliability of communication networks.

Despite its substantial utility in industrial applications, `time series anomaly state detection problem' has not yet captured significant attention within the computer science community. This is reflected in the relatively limited research and literature dedicated to the development of specialized algorithms for addressing this challenge. 
Therefore, a purpose of our paper is to introduce the problem to the field of computer science and stimulate discourse within the research community. This effort will be achieved through careful elaboration and precise definition of the problem and proposing a comprehensive framework for solving it. The main contributions of our study can be summarized as follows:

\begin{itemize}

\item \textbf{Introduction to a conceptual framework:}
Our study introduces a novel conceptual framework by rigorously defining the `time series anomaly state detection problem' as well as its corresponding anomaly with mathematical tools, and then propose an implementation framework that can be carried out concretely.

\item \textbf{Synthetic dataset construction:}
Considering that the definition of this problem is different from traditional outlier detection, a general dataset cannot be used. Therefore, this paper proposes a dataset construction method: generate data from a time series classification dataset, and applies it to experimental analysis.

\item \textbf{Empirical Evaluation and Comparative Analysis:}
Finally, our study culminates in an extensive empirical evaluation, in which the performance of 38 algorithms modified for our task is unbiasedly compared within our established framework. This comparative analysis sheds light on the relative strengths and weaknesses of different algorithms, as well as their time efficiency and stability, thereby providing valuable insights for future research and applications in this area.

% \textbf{Maybe a paragraph for introducing the arrange of the article}

\end{itemize}

\section{Related Work}

% \subsection{Time Series Anomaly Detection Algorithms}
\label{relatedwork:timeseries}

There are various algorithms for detecting anomalies in time series, which can be roughly divided into four categories based on their working mechanisms, namely, forecast-based, reconstitution-based, statistical-model-based, and proximity-based algorithms. 

\textbf{Forecast-based algorithms} refer to ones, wherein predictive models are trained to anticipate future values first, and subsequently, identify the anomaly by assessing the disparity between the predicted values and the actual observed values. 
Its representative algorithms include polynomial-based models, LinearRegreesion(LR), and neural-networks-based models, fully connected neural networks (FCNN)~\cite{back1991fir}, convolutional neural networks (CNN)~\cite{munir2018deepant}, recurrent neural network (RNN)~\cite{connor1994recurrent}, gated recurrent unit (GRU)~\cite{chung2014empirical}, and long short-term memory (LSTM)~\cite{hundman2018detecting,hochreiter1997long}.

\textbf{Reconstitution-based algorithms} aim to learn a model that captures the latent structure of the given time series data and generate synthetic reconstitution, and then discriminate the anomaly by reconstitution performance under assumption that outliers cannot be efficiently reconstructed from the low-dimensional mapping space. 
Reconstitution-based algorithms can be mainly divided into two categories: low-dimensional feature space reconstitution, including Kernel Principal Component Analysis (KPCA)~\cite{hoffmann2007kernel}, Auto-Encoder (AE)~\cite{sakurada2014anomaly}, Variational Auto-Encoder (VAE)~\cite{kingma2013auto}, and Transformer, and adversarial generation models, including Time Series Anomaly Detection with Generative Adversarial Networks (TanoGan)~\cite{bashar2020tanogan} and Time Series Anomaly Detection Using Generative Adversarial Networks (TadGan)~\cite{geiger2020tadgan}.

\textbf{Statistical-model-based algorithms} assume that data is generated by a certain statistical distribution, and use statistical models to perform hypothesis testing to determine which data does not meet the model assumptions, thereby detecting outliers. 
Generally, it can be divided into two categories, namely, methods based on empirical distribution functions, including Copula-Based Outlier Detection (COPOD)~\cite{li2020copod}, Empirical Cumulative Distribution-Based Outlier Detection (ECOD)~\cite{li2022ecod}, Histogram-based outlier detection (HBOS)~\cite{birge2006many}, Lightweight on-line detector of anomalies (LODA)~\cite{pevny2016loda}, and methods based on theoretical distribution functions, including Gaussian Mixture Models (GMMs)~\cite{aggarwal2015theoretical}, Kernel density estimation (KDE)~\cite{latecki2007outlier}, Mean Absolute Differences (MAD)~\cite{iglewicz1993volume} Algorithm, Outlier Detection with Minimum Covariance Determinant (MCD)~\cite{rousseeuw1999fast}.

\textbf{Proximity-based algorithms} for anomaly detection are techniques that rely on using various distance measures to quantify similarity or proximity between data points, including density, distance, angle and dividing hyperspace, and then identify anomalies, under the assumption that distribution of abnormal points is different from that of normal points, causing the similarity or proximity low. 
Density-based algorithms includes Cluster-Based Local Outlier Factor (CBLOF)~\cite{he2003discovering}, Connectivity-Based Outlier Factor (COF)~\cite{tang2002enhancing},  Local Outlier Factor (LOF)~\cite{breunig2000lof}, Quasi-Monte Carlo Discrepancy outlier detection (QMCD)~\cite{fang2001wrap}; Distance-based methods includes Cook's distance outlier detection (CD)~\cite{cook1977detection}, Isolation-based anomaly detection using nearest-neighbor ensembles (INNE)~\cite{bandaragoda2018isolation}, K-means outlier detector (Kmeans)~\cite{chawla2013k}, K-Nearest Neighbors (KNN)~\cite{ramaswamy2000efficient}, Outlier detection based on Sampling (SP)~\cite{sugiyama2013rapid}, Principal Component Analysis algorithm (PCA)~\cite{shyu2003novel}, Subspace Outlier Detection (SOD)~\cite{kriegel2009outlier}, Stochastic outlier selection (SOS)~\cite{janssens2012stochastic}; Angle-based methods include Angle-based outlier detection (ABOD)~\cite{kriegel2008angle}; Dividing-hyperplane-based methods includes Deep One-Class Classification for outlier detection (DeepSVDD)~\cite{ruff2018deep}, Isolation forest (IForest)~\cite{liu2008isolation}, One-class SVM (OCSVM)~\cite{scholkopf2001estimating}, One-Classification using Support Vector Data Description (SVDD)~\cite{tax2004support}.

However, algorithms mentioned above mainly focus on problems related to outlier detection. 
Therefore, we hope to provide a basic and complete paradigm for solving `time series anomaly state detection problem' by strictly defining the  problem and conducting detailed experimental analysis.
Our goal is not only to solve the problem, but also strive to build a solid academic foundation and aim to provide a pioneering framework that guides and enhances future research efforts in this specialized area.

\section{Time Series Anomaly State Detection}
\subsection{Mathematical Definition Of The Problem}
\label{sec:mathematicaldefinition}
\textbf{Stochastic Process:}
Suppose $E$ is a random experiment and the sample space is $S=\{e\}$, if for each sample point $e \in S$, there always exists a time function $X(t,e)$ and $t \in T $ corresponds to it. Then in this way, the family of time $t$ functions $\{ X(t,e),t\in T\}$ obtained for all $e \in S$ is called a stochastic process, abbreviated as $\{ X(t),t \in T\}$. Each function in the family is called a sample function of this Stochastic process. $T$ is the variation range of parameter $t$, which is called parameter set.

\textbf{Time Series:} 
A time series $\{X(t),t \in \mathbb{N} \}$  is defined as a stochastic process with discrete time points. $x(t)$ represent the observed value of $X(t)$, and call $\{x(t),t \in \mathbb{N}\}$ an orbit of $\{X(t),t \in \mathbb{N}\}$. 

\textbf{Problem (Time Series Anomaly State Detection through One-Class Classification):} Given a standard time series dataset of inliers $\mathcal{D} = \{  \Tilde{x}_j (t), t \in [\Tilde{N}_j]  \}_{j=1}^{m_0}$, containing $m_0$ independently and identically distributed (i.i.d) orbits $\{ \Tilde{x}_j(t) \}_{t=1}^{\Tilde{N}_j}, \Tilde{x}_j(t) \in \mathbb{R}^{d_0}$ of a standard stochastic process $\{X_0(t),t \in \mathbb{N}\}$ (a set of standard stochastic process $\{X_{p}(t),t \in \mathbb{N}\}_{p=1}^{n_p}$), and an unseen test time series dataset $\mathcal{U}= \{  \hat{x}_j (t), t \in [\hat{N}_j]  \}_{j=1}^{m_1}, \hat{x}_j (t) \in \mathbb{R}^{d_0}$ and its segment subset $\mathcal{U}_{\mathcal{S}} = \{  S_j^k \triangleq \{  \hat{x}_j (t), t \in [r_k, s_k]  \} \}_{j\in [1,m_1],k\in \mathbb{N}}$, where $S_j^k$ is segment intercepted from the test time series orbit, the goal of time series anomaly detection from OCC problem is to identify the outlier segment set $ \mathcal{O} = \{ S_j^k  \subset  \{  \hat{x}_j (t), t \in [\hat{N}_j]  \} \in \mathcal{U}: S_j^k \nsim \{X_0(t),t \in \mathbb{N}\}  \}$.

In order to deal with this time series anomaly state detection problem, we can consider the following mathematical framework. First, employ a pre-processing function $\mathcal{P}$ to extract time information features $\mathcal{F}_{standard} \triangleq \mathcal{P}(\mathcal{D})$ and $ \mathcal{F}_{test} \triangleq \mathcal{P}(\mathcal{U_{\mathcal{S}}})$. And then, employ the detection algorithm $\mathcal{M}$ and standard pre-processed dataset $\mathcal{F}_{standard}$ to learn a anomaly score function $A_{\mathcal{M}}(\cdot)$. Next, use the learned anomaly score function $A_{\mathcal{M}}(\cdot)$ to infer an anomaly score $A_{\mathcal{M}}(\mathcal{P}(S_j^k))$ to assess whether $S_j^k $ is an outlier. Generally, it is assumed that the larger the anomaly score $A_{\mathcal{M}}(\mathcal{P}( S_j^k ))$, the more likely the corresponding $ S_j^k $ is an outlier. Therefore, the decision set (detected outlier segment set) is selected as: $ \hat{\mathcal{O}}_{\mathcal{M}} = \{ S_j^k  \subset  \{  \hat{x}_j (t), t \in [\hat{N}_j]  \}  \in \mathcal{U}:  A_{\mathcal{M}}(\mathcal{P}(  S_j^k  )) \geq L_{\mathcal{M}} \}$, where $ L_{\mathcal{M}} $ is a threshold associated with detector $\mathcal{M}$. Denote the subset of $\mathcal{U}_{\mathcal{S}}$ containing all inliers as $ \mathcal{O}^c = \{ S_j^k  \subset  \{  \hat{x}_j (t), t \in [\hat{N}_j]  \} \in \mathcal{U}: S_j^k \sim \{X_0(t),t \in \mathbb{N}\}  \}$, and $\mathcal{O} \cup \mathcal{O}^{c} = \mathcal{U} $.

Taking a multiple testing perspective, the null hypothesis is that a new time series segment is an inlier, while the alternative asserts that it is an outlier. That is, the time series anomaly state detection from one-class classification problem is translated into a multiple testing problem as follows,
\begin{equation}
\begin{aligned}
& \mathbb{H}_{0,(j,k)}: S_j^k \sim \{X_0(t),t \in \mathbb{N}\} \quad v.s. \\ 
& \mathbb{H}_{1,(j,k)}: S_j^k \nsim \{X_0(t),t \in \mathbb{N}\} \ , \ j = 1,\dots,n, \ k \in \mathbb{N}
\end{aligned}
\end{equation}

\begin{figure*}[ht]
    \centering
    \subfigure
    {\includegraphics[width=0.75\textwidth]{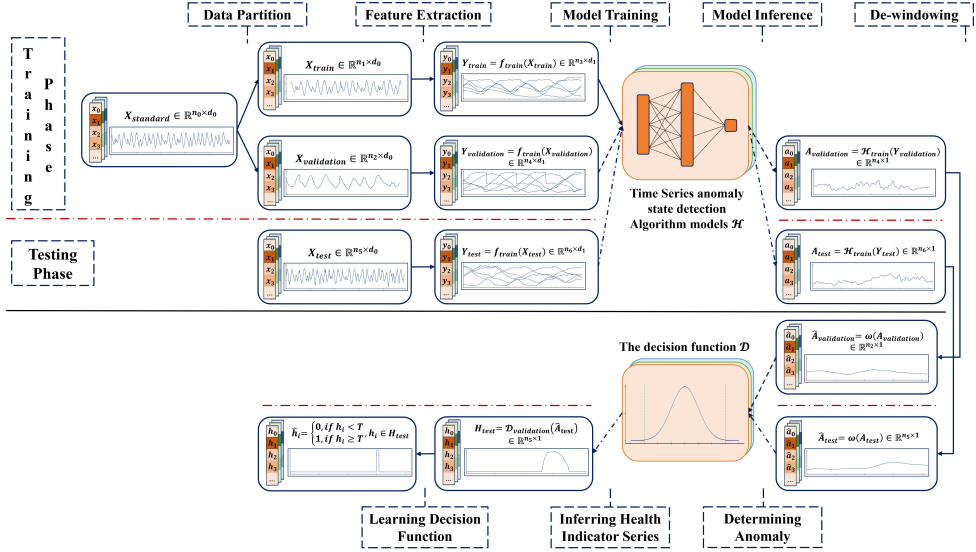}}
    \caption{Overview of the implementation frame. The entire framework is divided into a training phase and a testing phase. The training phase is used to learn the characteristics of standard time series, while the testing phase is used to infer anomalies in the test time series.
    \label{fig:traningandtesting}}
\end{figure*} 

\subsection{Implementation framework of time series anomaly state detection}

In the preceding subsection, we utilized knowledge of stochastic processes to define time series anomaly state detection and presented a mathematical framework for addressing it. However, this framework, while theoretically robust, is abstract and not applicable in practice. Therefore, in this subsection, we shift our focus to practical application, proposing a framework that can be implemented to solve this problem in real-world scenarios. Specifically, our implementation process are divided into two distinct phases: the training phase and the testing phase, as shown in Fig.~\ref{fig:traningandtesting}.

In the training phase, as delineated in Fig.~\ref{fig:traningandtesting} and Algorithm~\ref{standard training procedure} in Appendix~\ref{sec:concreteoperation}, the process primarily encompasses several key steps: data partition, feature extraction, model training, model inference, de-windowing and learning decision function. While, in the testing phase, also as outlined in Fig.~\ref{fig:traningandtesting} and Algorithm~\ref{standard testing procedure} in Appendix~\ref{sec:concreteoperation}, the procedure mainly includes: feature extraction, model inference, de-windowing, inferring health indicator series, and determining anomaly.

% It's worth noting that validation time series data $X_{train}$ is important here, because when the algorithm function $h$ learns $Y_{train}$, then naturally $A_{train}$ will be smaller then the anomaly score of other inliers. Therefore, $A_{validation} $ will be closer to the true distribution of the standard time series data under the algorithm function $h^{*}$, which is more consistent with the reality.

Based on the implementation framework proposed above, we further conduct an intuitive analysis and define the outliers under this framework in Appendix \ref{sec:intuitiveanalysis}.

\section{Preparation for Experiments}
\label{Sec:Dataset}
\subsection{Construction of Dataset}

There are many open source anomaly detection data sets on the market, including open source nature data
~\cite{lavin2015evaluating,paparrizos2022tsb,bhatnagar2021merlion} and artificial ones ~\cite{lavin2015evaluating, han2022adbench, lai2021revisiting}. However, they serve the task of outlier detection, where anomalies in these datasets are often departure points that deviate from the overall distribution within the time series, which cannot be applied to our task.

% Therefore, when considering the time series anomaly state detection tasks under OCC defined in this article, these datasets cannot be used, as the realAWSCloudewatch  example shown in Sec \ref{DFO}. Thus, in response to this current situation, we have to choose to construct artificial data ourselves:

\subsubsection{Basic guidelines for data construction}

As we defined the task in Sec.~\ref{sec:mathematicaldefinition}, we consider a segment as an outlier if it does not appear in standard time series. Therefore, an intuitive idea is to first find a suitable time series baseline, which follows a set of certain given stochastic process, and then divide it into standard time series and pre-test time series, where outliers are injected into the pre-test time series to form test time series. And, we believe that three basic requirements must be met: (i) The baseline has a certain periodicity. This ensures that the normal part of the ``test time series'' should have appeared in the training data; if there is no periodicity at all, then the ``test time series'' is likely to be abnormal at all. (ii) There is a certain amount of "noise". We hope that each period has a certain similarity, but not exactly the same, which is more consistent with the real working conditions; (iii) There is more effective information than "noise". That is, we hope that the baseline has a high signal-to-noise ratio.

\subsubsection{Build datasets from categorical time series data}

Based on the intuitive idea and requirements in the previous subsection, using the time series classification dataset to construct an time series anomaly state detection dataset seems to be suitable, for in time series classification datasets, data belonging to one category tends to be relatively close, while there are certain differences between data in different categories. Therefore, if we use a certain category in the classification datasets as the baseline and then appropriately insert another category into the test data as anomaly data, then we could create time series anomaly state detection dataset that meets our requirements in previous subsection. To ensure the quality of the constructed dataset, we hope that the normal part of the test time series data is close to the standard time series data, while the abnormal part has a certain distance from the standard time series data. Finally, to satisfy the discussion above demands, we construct the artificial dataset as Algorithm~\ref{BuildFromCategoricalData}. In order to ensure the quality of the dataset, we use algorithms and data to check each other: if the AUC-ROC of the results of all algorithms on a dataset is below 0.8, then we will remove this artificial data from the artificial dataset.

\begin{algorithm}[!h]
\caption{Build datasets from categorical time series data}
\renewcommand{\algorithmicrequire}{\textbf{Input:}}
\renewcommand{\algorithmicensure}{\textbf{Output:}}
\begin{algorithmic}[1]
\label{BuildFromCategoricalData}

\REQUIRE The data from a given classification dataset $\{ (X_i,y_i)\}_{i=1}^{N}, X_i \in \mathbb{R}^{m}, y \in \mathbb{ R}$, and the shape-based distance (sbd) distance $d(x,y)$.

\STATE Pick out the $k$-th type of data $\{ (X_i,y_i)\}_{y_i=k}$

\STATE Perform k-means clustering with sbd on $\{ (X_i,y_i)\}_{y_i=k}$ to ensure that similar data can be found even in one category and ensure the consistency of the baseline.

\STATE Select the most concentrated category $\{ (X_i,y_i)\}_{y_i=\hat{k}}$ in 2 as baseline and randomly shuffle it.

\STATE Calculate the average distance $\{ d(x_i,y_i) \}_{y_i \neq k }$ from each data in other categories to the baseline $\{ (X_i,y_i)\}_{y_i=\hat{k}}$, respectively.

\STATE Extract the first half of the baseline dataset $\{ (X_i,y_i)\}_{y_i=\hat{k}}$, and sequentially concatenate these elements to construct the standard time series, denoted as $X_{standard}$. Subsequently, based on the average distance $\{ d(x_i,y_i) \}_{y_i \neq k}$, incorporate suitable data from categories other than the $k$-th into the second half of $\{ (X_i,y_i)\}_{y_i=\hat{k}}$ randomly. These are then sequentially concatenated to form the test time series, labeled as $X_{test}$.

\end{algorithmic}
\end{algorithm}

\subsection{The Time Series Anomaly State Detection Algorithms Under One-Class Classification}
\label{Subsec:algorithms}

In this paper, we look at time series anomaly state detection from the perspective of One-Class Classification, therefore, we have modified anomaly detection algorithms so that they can maintain the ability to handle traditional outlier detection tasks, while also being able to handle the time series anomaly state detection tasks we defined. 
We have compiled a comprehensive algorithm collection with 38 time series anomaly detection methods, which could be categorized into four types based on their working mechanisms, i.e., forecast-based algorithms, reconstitution-based algorithms, statistical-model-based algorithms, and proximity-based algorithms. For the introduction of these algorithms, we have described them in Sec. \ref{relatedwork:timeseries}, and provide more information about them in Appendix~\ref{sec:algorithmslist}.

%%%%%%%%%%%%%%%%%%%%%%%%%%%%%%%%
\begin{table*}[htbp]
\renewcommand\arraystretch{1.1}
\setlength\tabcolsep{3.0pt}
\centering
\caption{Average value of eight representative accuracy evaluation measures for all the thirty-eight algorithms.}
\scalebox{0.90}{
\label{table:averagevalue}
\fontsize{6pt}{4.0pt}\selectfont
\begin{tabular}{|>{\centering\arraybackslash}m{2.5cm}||>{\centering\arraybackslash}m{1.5cm}|>{\centering\arraybackslash}m{1.5cm}|>{\centering\arraybackslash}m{1.5cm}|>{\centering\arraybackslash}m{1.5cm}|>{\centering\arraybackslash}m{1.5cm}|>{\centering\arraybackslash}m{1.5cm}|>{\centering\arraybackslash}m{1.5cm}|>{\centering\arraybackslash}m{1.5cm}|}
\toprule[0.8pt]
Algorithms & Precision & Recall & F1 & Range-F1 & AUC-ROC & AUC-PR & VUS-ROC & VUS-PR \\
\midrule
ABOD & 0.183 & 0.200 & 0.158 & 0.174 & 0.797 & 0.775 & 0.664 & 0.650  \\ 
AE & 0.743 & 0.729 & 0.613 & 0.655 & 0.907 & 0.802 & 0.785 & 0.680  \\ 
CBLOF & 0.799 & 0.631 & 0.562 & 0.656 & 0.879 & 0.773 & 0.771 & 0.664  \\ 
CD & 0.378 & 0.197 & 0.165 & 0.253 & 0.596 & 0.583 & 0.514 & 0.510  \\ 
CNN & 0.739 & 0.690 & 0.567 & 0.633 & 0.876 & 0.741 & 0.757 & 0.629  \\ 
COF & 0.815 & 0.591 & 0.543 & 0.650 & 0.858 & 0.724 & 0.765 & 0.624  \\ 
COPOD & 0.412 & 0.220 & 0.207 & 0.297 & 0.678 & 0.425 & 0.572 & 0.359  \\ 
DeepSVDD & 0.754 & 0.807 & 0.644 & 0.680 & 0.913 & 0.764 & 0.816 & 0.677  \\ 
ECOD & 0.452 & 0.247 & 0.232 & 0.341 & 0.686 & 0.428 & 0.580 & 0.370  \\ 
FCNN & 0.813 & 0.696 & 0.596 & 0.663 & 0.881 & 0.753 & 0.776 & 0.653  \\ 
GMM & 0.683 & 0.672 & 0.553 & 0.614 & 0.861 & 0.684 & 0.751 & 0.588  \\ 
GRU & 0.581 & 0.579 & 0.463 & 0.508 & 0.822 & 0.628 & 0.696 & 0.524  \\ 
HBOS & 0.721 & 0.488 & 0.448 & 0.549 & 0.810 & 0.686 & 0.692 & 0.583  \\ 
IForest & 0.730 & 0.520 & 0.474 & 0.568 & 0.838 & 0.707 & 0.710 & 0.595  \\ 
INNE & 0.708 & 0.581 & 0.505 & 0.577 & 0.844 & 0.740 & 0.732 & 0.646  \\ 
KDE & 0.812 & 0.776 & 0.641 & 0.690 & 0.921 & 0.811 & 0.792 & 0.688  \\ 
KMeans & 0.818 & 0.689 & 0.614 & 0.685 & 0.895 & 0.803 & 0.795 & 0.700  \\ 
KNN & 0.798 & 0.799 & 0.686 & 0.719 & 0.949 & 0.857 & 0.847 & 0.749  \\ 
KPCA & 0.602 & 0.724 & 0.560 & 0.574 & 0.883 & 0.705 & 0.765 & 0.606  \\ 
LinearRegression & 0.450 & 0.459 & 0.352 & 0.403 & 0.746 & 0.501 & 0.645 & 0.422  \\ 
LODA & 0.693 & 0.466 & 0.395 & 0.510 & 0.785 & 0.640 & 0.664 & 0.565  \\ 
LOF & 0.750 & \textbf{0.873} & 0.713 & 0.701 & 0.953 & 0.851 & 0.865 & 0.762  \\ 
LSTM & 0.578 & 0.621 & 0.485 & 0.527 & 0.827 & 0.649 & 0.713 & 0.547  \\ 
MAD & 0.582 & 0.357 & 0.320 & 0.406 & 0.752 & 0.581 & 0.616 & 0.502  \\ 
MCD & 0.511 & 0.472 & 0.366 & 0.435 & 0.751 & 0.537 & 0.631 & 0.441  \\ 
MSD & 0.493 & 0.290 & 0.259 & 0.341 & 0.704 & 0.510 & 0.583 & 0.450  \\ 
OCSVM & 0.466 & 0.153 & 0.161 & 0.311 & 0.614 & 0.387 & 0.533 & 0.355  \\ 
PCA & 0.275 & 0.066 & 0.085 & 0.177 & 0.566 & 0.332 & 0.499 & 0.314  \\ 
QMCD & 0.053 & 0.443 & 0.074 & 0.074 & 0.654 & 0.663 & 0.598 & 0.595  \\ 
RNN & 0.600 & 0.636 & 0.507 & 0.554 & 0.841 & 0.667 & 0.729 & 0.571  \\ 
Sampling & \textbf{0.841} & 0.818 & \textbf{0.718} & \textbf{0.762} & \textbf{0.956} & \textbf{0.873} & 0.858 & \textbf{0.769}  \\ 
SOD & 0.771 & 0.678 & 0.567 & 0.630 & 0.875 & 0.714 & 0.752 & 0.607  \\ 
SOS & 0.510 & 0.825 & 0.535 & 0.476 & 0.955 & 0.851 & \textbf{0.883} & 0.740  \\ 
SVDD & 0.642 & 0.421 & 0.362 & 0.491 & 0.757 & 0.567 & 0.640 & 0.513  \\ 
TadGan & 0.804 & 0.582 & 0.523 & 0.622 & 0.843 & 0.736 & 0.733 & 0.639  \\ 
TanoGan & 0.446 & 0.176 & 0.193 & 0.313 & 0.651 & 0.529 & 0.541 & 0.481  \\ 
Transformer & 0.675 & 0.510 & 0.441 & 0.519 & 0.806 & 0.620 & 0.685 & 0.531  \\ 
VAE & 0.691 & 0.559 & 0.481 & 0.541 & 0.831 & 0.692 & 0.698 & 0.576  \\ 
\bottomrule[1pt]
\end{tabular}
}
\end{table*}
%%%%%%%%%%%%%%%%%%%%%%%%%%%%%%%%

\subsection{Accuracy Evaluation Measures}

Many accuracy evaluation measures have been proposed to quantitatively evaluate the detection performance of different anomaly detection algorithms and help select the optimal model through comparison. In this article, we will mainly focus on four threshold-based measures, Precision, Recall, F1-score, and Range F1-score~\cite{tatbul2018precision}, and four threshold-independent measures, the area under (AUC) the receiver operating characteristics curve (AUC-ROC) and the precision-recall curve (AUC-PR), and the volume under the surface (VUS) for the ROC surface (VUS-ROC) and PR surface (VUS-PR) ~\cite{paparrizos2022volume}.
Considering that threshold-independent measures are generally better than threshold-based ones in terms of robustness, separability and consistency~\cite{paparrizos2022volume}, we will conduct more analyzes based on them in this article.

\subsection{Methods to evaluate the difficulty of TSADBench dataset}

In addition to accuracy evaluation measures, the quantification of the difficulty coefficient of the dataset should also be considered, which helps to conduct a more detailed evaluation of model performance. A comprehensive understanding of the properties of the dataset can not only enrich the comparative analysis of algorithms, such as robustness, but can also guide the selection and calibration of models in practical applications.

Previous articles have discussed several indicators to measure the difference in distribution of normal points and abnormal points, including Relative Contrast (RC)~\cite{he2012difficulty}, Normalized clusteredness of abnormal points (NC)~\cite{emmott2013systematic}, and Normalized adjacency of normal/abnormal cluster (NA)~\cite{paparrizos2022tsb}, which we elaborate on in Appendix~\ref{sec:rcncna}.

However, the three indicators RC, NC and NA are all constructed for traditional outlier analysis, and are not applicable to the dataset we constructed. Therefore, inspired by the previous work and combined with the characteristics of our task, we proposed the k-nearest-neighbors (knn) normalized clusteredeness (KNC) of abnormal points method to measure the difference between the distributions of normal points and abnormal points, and use it to evaluate the difficulty of the our dataset. Suppose that $s$ represents time series, $D(s_i,s_j)$ denote the SBD distance between two series $s_i$ and $s_j$, $S$ is the whole set, $\mathcal{S}_{std}$ denotes the set of standard sequences, $\mathcal{S}_{nor}$ and $\mathcal{S}_{abn}$ denotes the set of normal and anomalous sequences in test sequences, and $D^{k}_{mean}(x,\mathcal{S}) = \frac{1}{k} \sum_{s_i \in \text{knn to x}, s_i \in \mathcal{S} } D(x,s_i)$ is the mean SBD distance between time series $x$ and its k-nearest-neighbors in the set $\mathcal{S}$, then we define KNC as the ratio of the average k-nearest-neighbor SBD distance between the anomalous sequences and standard set to that between the normal sequences and standard set,

\begin{equation}
\begin{aligned}
KN_c = \frac{\mathbb{E}_{s \in S_{\text{ano}}} \left[D^{k}_{mean}(s,\mathcal{S}_{std})\right]}
{\mathbb{E}_{s \in S_{\text{nor}}} \left[D^{k}_{mean}(s,\mathcal{S}_{std})\right]}
\end{aligned}
\end{equation}

The K-nearest-neighbor here is critical, because we do not require the data in the test data set to be close to all the data in the standard data set, but only require that there is data in the standard data set that is close to it.

\section{Experiments}

\subsection{Experiments setup}

\subsubsection{Feature Extraction}

\textbf{1) Feature extraction: } For each dataset, we first normalize the data between $[0,1]$. Then we set the window lengths to 32, 64, 128, 256, the length based on~\cite{imani2021multi}, the length based on Auto-correlation Function, and the length based on Fourier transform, respectively, with stride equals to 1 . 

\textbf{2) Algorithms: } Here, we have implemented all 38 algorithms mentioned in Sec. \ref{Subsec:algorithms}, and the key parameters of each algorithm have been tried within the appropriate range.

\textbf{3) De-windowing function $\omega$: }
Given the original time series $X \in \mathbb{R}^{n_x \times d_x}$ and the corresponding anomaly score $A \in \mathbb{R}^{n_y \times 1}$, then modified anomaly score $\hat{A}=w(A)=(\hat{a}_1,\hat{a}_2,\dots,\hat{a}_{n_x}) \in \mathbb{ R}^{n_x \times 1}$ is defined as follows:

For a given $i \in [n_x]$, assume that $\mathcal{S}_i$ is the serial number set of $A$ which contains the information of $x_i$, then we have $\hat{a}_i = \frac{1}{|\mathcal{S}_i|}\sum_{j \in \mathcal{S}_i} a_j $.

\textbf{4) Decision function $\mathcal{D}$ and decision threshold $T$: }
Given the validation modified anomaly score $\hat{A}_{v}$, the decision function $\mathcal{D}$ is modified by the integral cumulative distribution function of the standard normal distribution and defined as follows.
Let $\mu$ be the mean of $\hat{A}_{v}$, and $\sigma$ be the variance of $\hat{A}_{v}$, then
$\mathcal{D}(\hat{a}) = \max((\operatorname{erf}((\hat{a}-\mu-\sigma)/(\sqrt{2}\sigma))-0.5)*2,0), \ \operatorname{erf}(x)=\frac{2}{\sqrt{\pi}} \int_0^x e^{-t^2} dt $, with the decision threshold $T = 1 - (1- \operatorname{erf}(2/\sqrt{2})) / 2$

The decision function $\mathcal{D}$ here is essentially calculating the significance level $\alpha$ of $\hat{a}$ under a given modified test anomaly score $\hat{A}_{t}$. The transformation operations consider that only when $\hat{a}$ and $\mu$ reaches a certain distance, then $\hat{a}$ begin to be viewed as an outlier. We hope that when $\mathcal{D}(\hat{a})>0$, it means that an abnormality may occur. The decision threshold $T$ is chosen as $2\sigma$ because when calculating the decision function $\mathcal{D}$, $\hat{a}$ has already been subtracted by a $\sigma$.

\subsection{Overall experimental results}

\begin{figure}[htb]
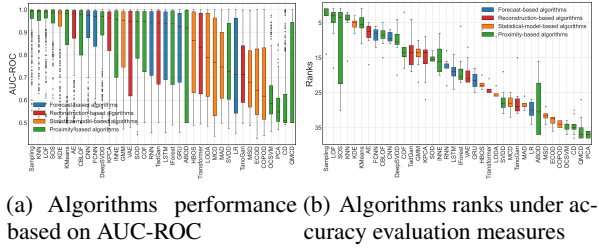

    \centering
    % \subfigure[F1-score]{\includegraphics[width=0.225\textwidth]{figures/Sec6/boxplot-ranks-f1-score.pdf}}
    \subfigure[Algorithms performance based on AUC-ROC]
    {\includegraphics[width=0.225\textwidth]{figures/Sec6/boxplot-ranks-auc-roc.pdf}}
    % \subfigure[VUS-ROC]{\includegraphics[width=0.225\textwidth]{figures/Sec6/boxplot-ranks-vus-roc.pdf}}
    \subfigure[Algorithms ranks under accuracy evaluation measures]{\includegraphics[width=0.225\textwidth]{figures/Sec6/boxplot-ranks-of-all-evalution.pdf}}
    
    \caption{The performance of different algorithms on all 38 algorithms, sorted in descending order based on median. In (a), the vertical axis represents AUC-ROC, and in (b), the vertical axis represents the ranks of the chosen evaluation measures. Colors indicate the type of anomaly detection algorithms.
    \label{fig:aucrocandranks}}
\end{figure} 

We first conduct 38 algorithms on the artificial dataset generated by categorical dataset, which contains about 369 time series with labeled anomalies. Table~\ref{table:averagevalue} presents the average value of eight representative accuracy evaluation measures (i.e., precision, recall, f1-score, range-f1-score, auc-roc, auc-pr, vus-roc, vus-pr) for all the thirty-eight algorithms across the dataset. Fig.~\ref{fig:aucrocandranks}(a) presents the average AUC-ROC across the entire datasets and algorithms. The results of other accuracy evaluation measures could be found in the Appendix~\ref{sec:moreperformancebox}. Furthermore, we calculate the ranks of all algorithms under the eight representative accuracy evaluation measures across the dataset, and draw a boxplot of the ranks of each algorithm under these measures, as shown in Fig.~\ref{fig:aucrocandranks}(b).  
From this initial inspection, three methods seem to perform well: Sampling, LOF and KNN. According to the classification of anomaly detection algorithms, the best-performing forecast-based algorithm is FCNN, the best-performing reconstitution-based algorithm is the FCNN algorithm, the best-performing statistical-model-based algorithm is the KNN algorithm, and the best-performing proximity-based algorithm is the Sampling algorithm. Through this ranking, we have initially observed two very interesting phenomena: (1) In fact, the algorithms with the best results are not algorithms based on neural networks, but some traditional distance-based algorithms; (2) In prediction-based algorithms with neural networks, the best performing one is the simplest FCNN, rather than ones with special structures.

\begin{figure}[htb]
    \centering

    % \subfigure[F1-score]{\includegraphics[width=0.45\textwidth]{figures/Sec6/cd-ranks-f1-score.pdf}}
    \subfigure
    {\includegraphics[width=0.45\textwidth]{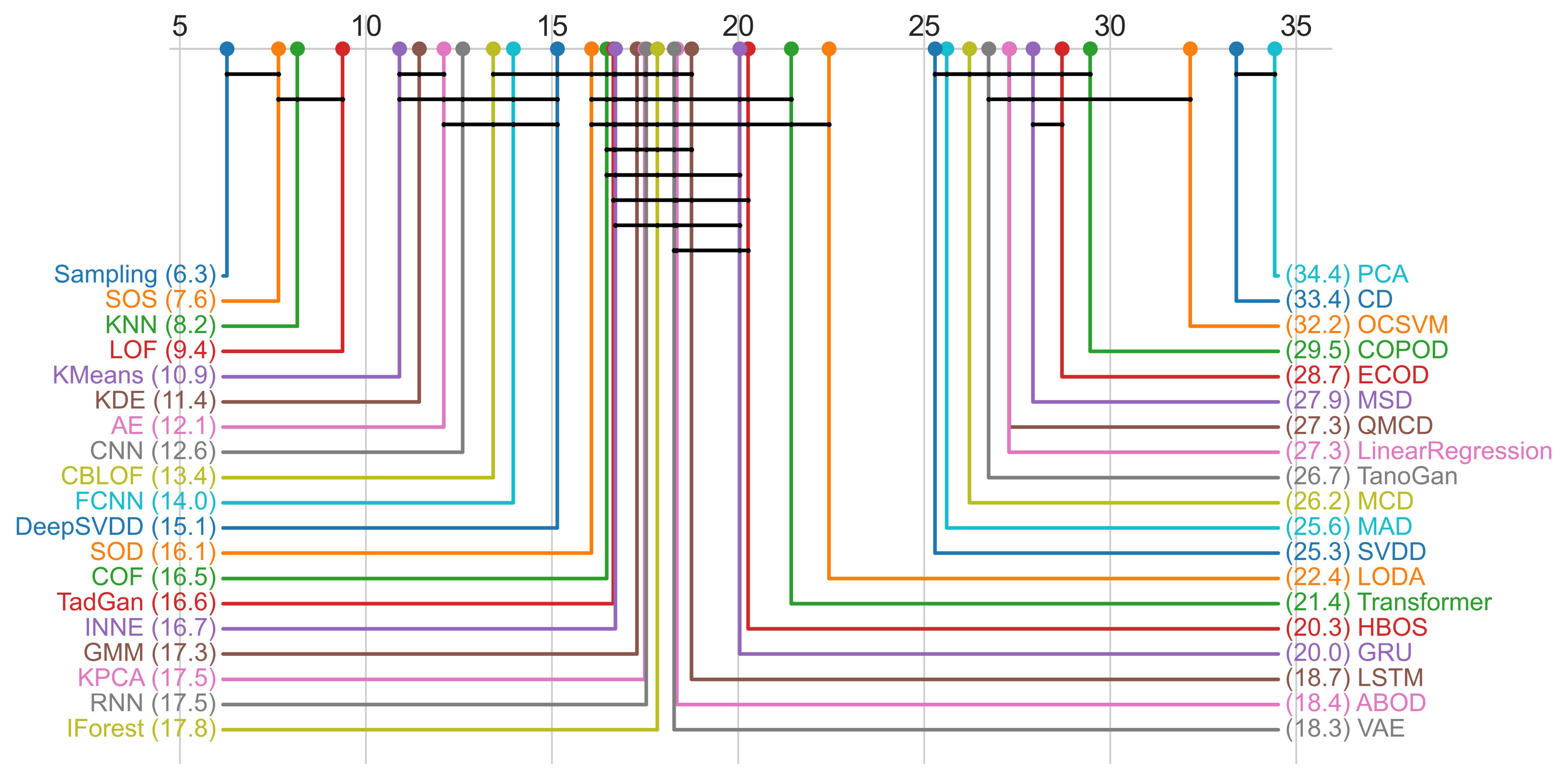}}
    
    \caption{Critical difference of 38 algorithms based on AUC-ROC
    \label{fig:criticaldifferece}}
\end{figure}

To better understand the ranks of the algorithms, we employed a Critical Difference (CD) diagram to illustrate and evaluate the comparative performance of various anomaly detection algorithms across the entire datasets. The CD diagram reveals the relative performance of the algorithms and their statistical significance of differences. Algorithms are arrayed according to their average ranks across all datasets, with the significance of performance differences ascertained through the Friedman test followed by Wilcoxon signed-rank test [113].
From the critical difference diagram in Fig.~\ref{fig:criticaldifferece} of AUC-ROC, we observe that the 'Sampling' algorithm (with an average rank of 6.1) significantly outperforms the rest of the algorithoms significantly except 'SOS' algorithm. This result is consistent with the results of each algorithm in Fig.~\ref{fig:aucrocandranks}(a). Critical difference graphs for more accuracy evaluation measures can be found in Appendix~\ref{sec:moreperformancecd}.

\begin{figure}[htb]
    \centering
    % 第一列
    \begin{minipage}{.225\textwidth}
        \centering
        \subfigure[F1-score]{\includegraphics[width=\linewidth]{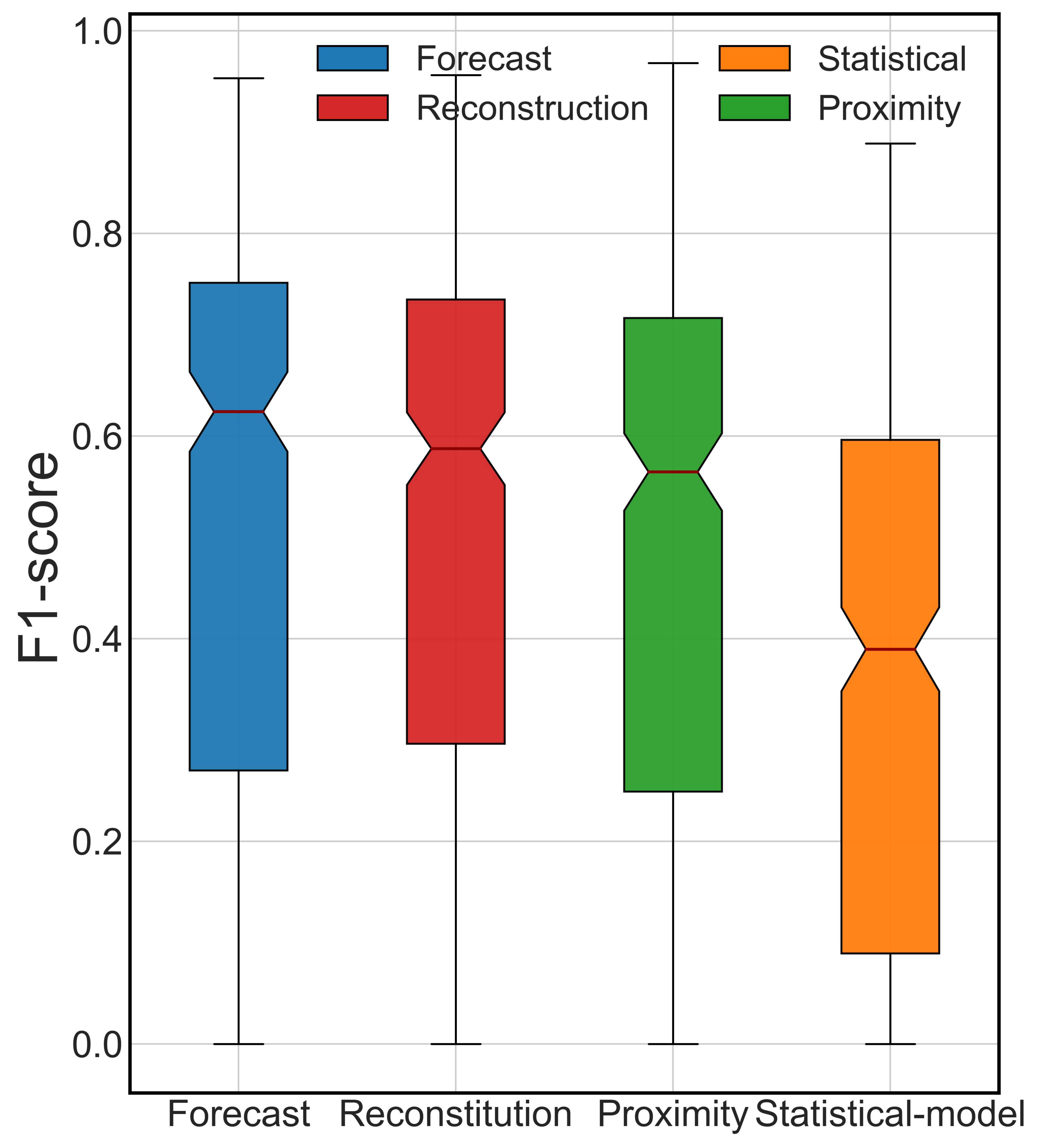}}
    \end{minipage}%
    % 第二列
    \begin{minipage}{.225\textwidth}
        \centering
        \subfigure[AUC-ROC]{
        \includegraphics[width=\linewidth]{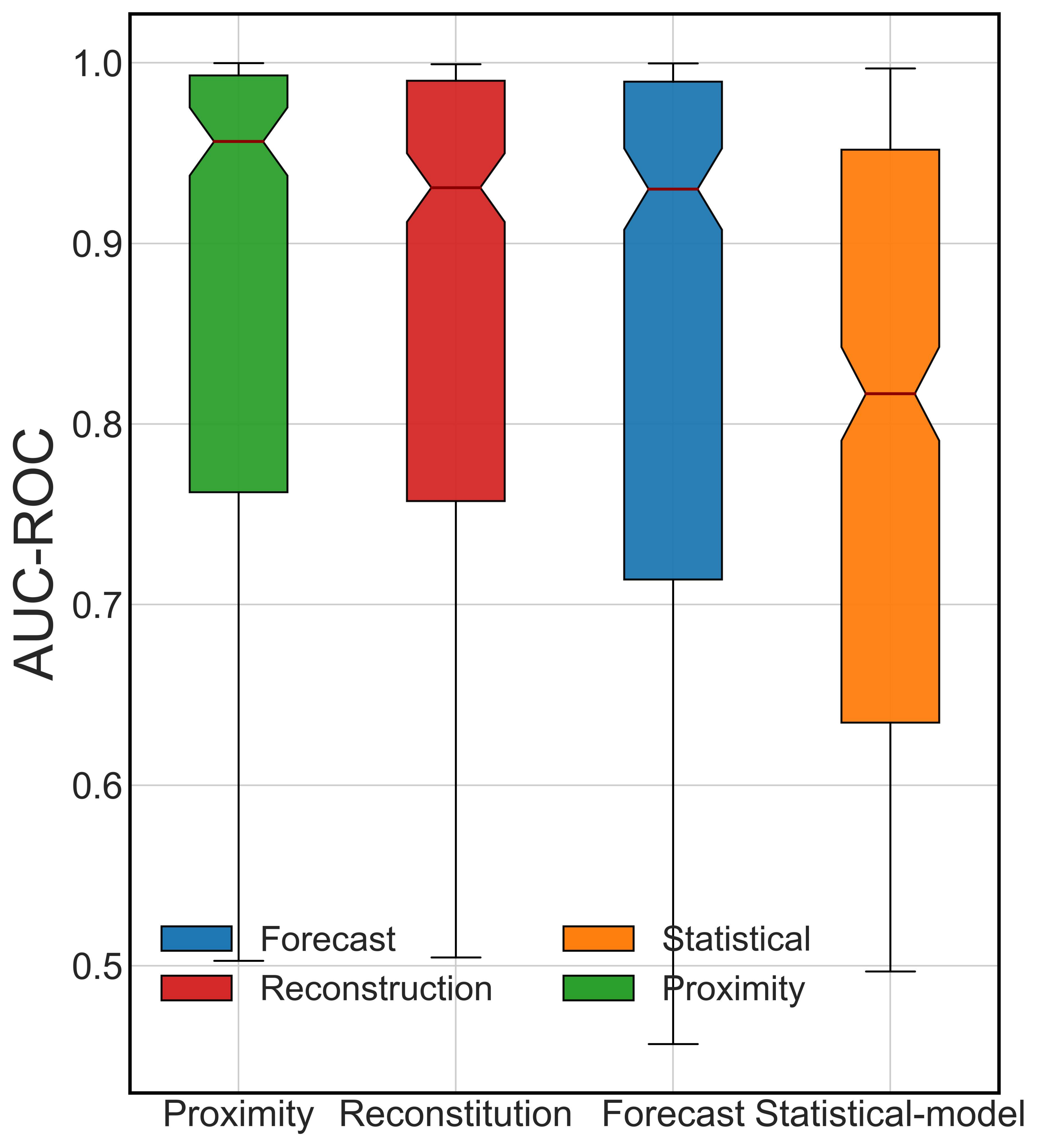}}
    \end{minipage}
    % 第三列
    \begin{minipage}{.45\textwidth}
        \centering
        \subfigure[Critical Difference under F1-score]{
        \includegraphics[width=\linewidth]{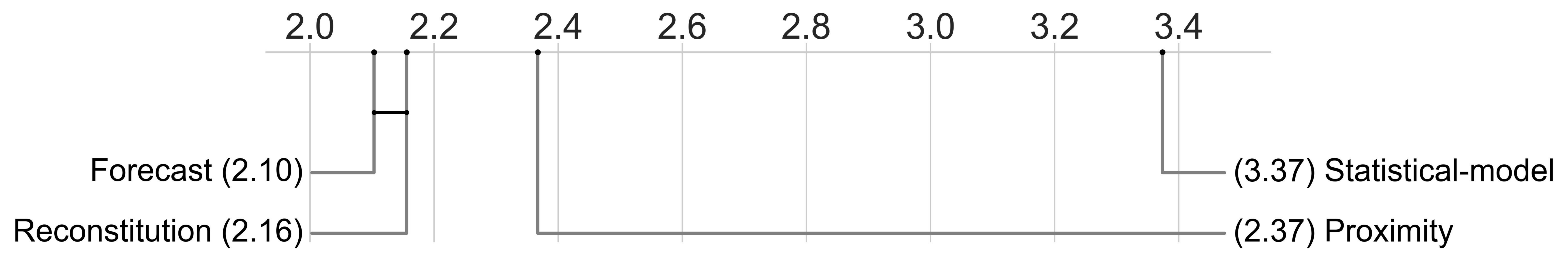}}
        \subfigure[Critical Difference under AUC-ROC]{
        \includegraphics[width=\linewidth]{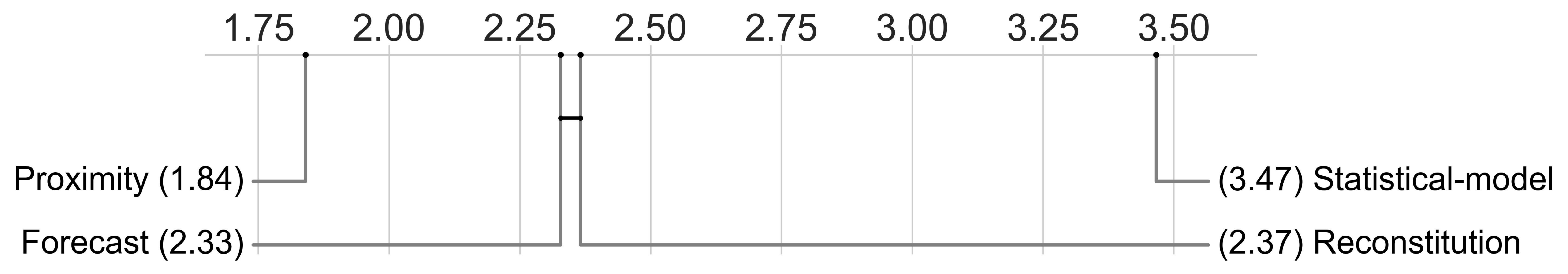}}
    \end{minipage}
    \caption{The performance of four types of anomaly detection algorithms under the AUC-ROC and F1 accuracy evaluation measures. (a,b) depict the performance of the four algorithms on the entire dataset in box-plots, while (c,d) further utilize CD diagram to reflect whether there are significant differences between them.}
    \label{fig:fourtypes}
\end{figure}

\subsection{Comparison among four different based algorithms}

Previously, we demonstrated the performance across different anomaly detection algorithms, whose results vary significantly. In this section, we will illustrate the analysis under the four types of algorithms , i.e., forecast-based algorithms, reconstitution-based algorithms, statistical-model-based algorithms, and proximity-based algorithms.

\begin{figure}[ht]
    \centering
    % \subfigure[rank of median]{\includegraphics[width=0.30\textwidth]{pic/boxplot_rank_results/Rank of median of methods based on types box.png}}
    
    \subfigure[Value of algorithms under eight evaluation measures]
    {\includegraphics[width=0.225\textwidth]{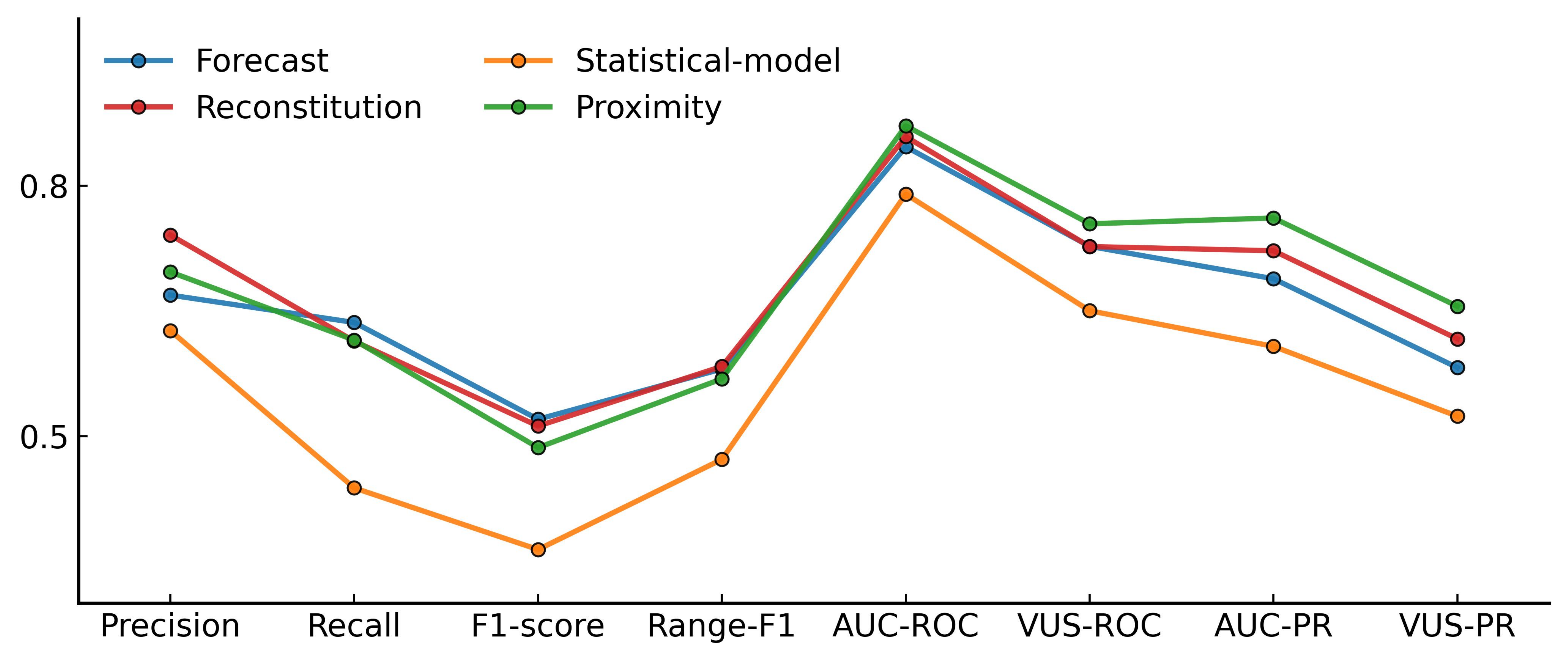}}
    \subfigure[Ranks of algorithms under eight evaluation measures]
    {\includegraphics[width=0.225\textwidth]{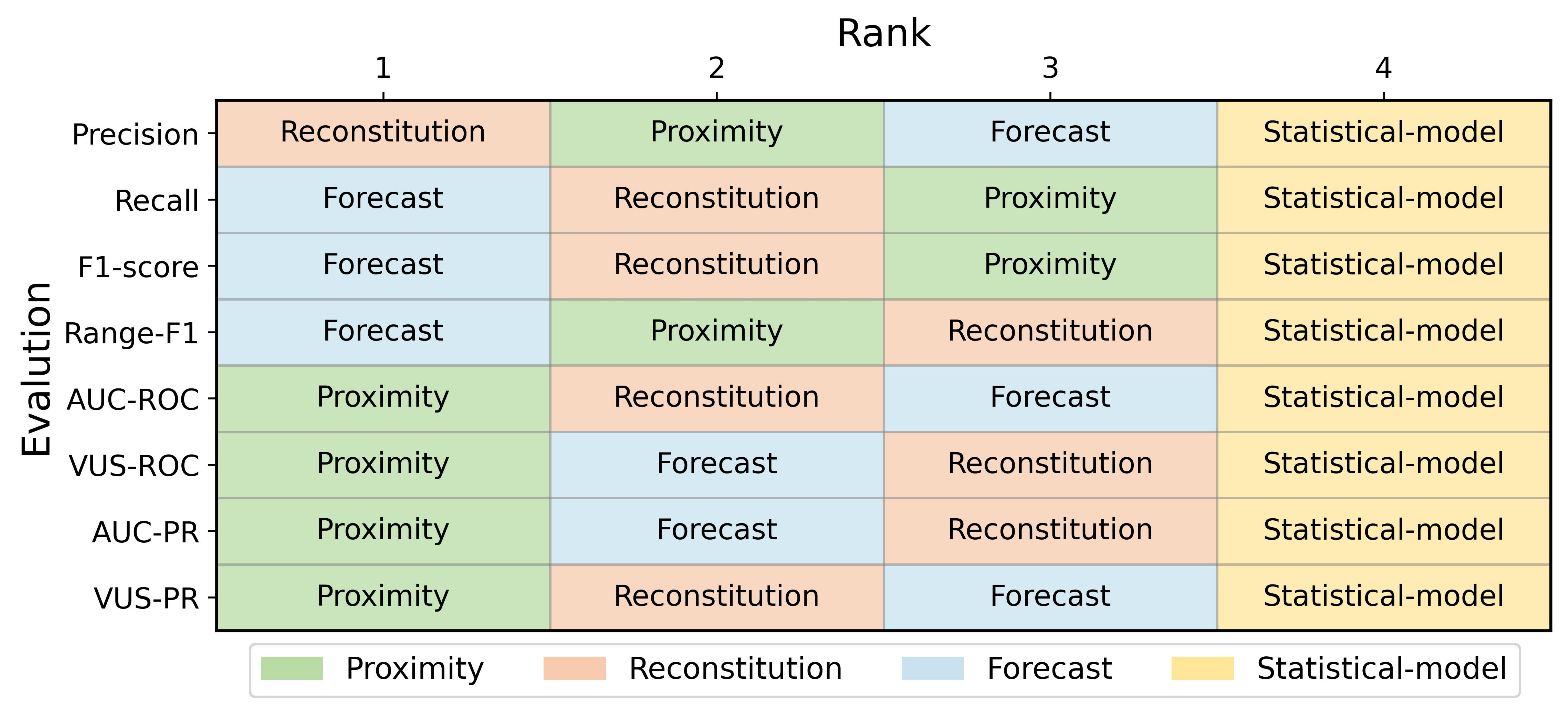}}

    \subfigure[Critical Difference of algorithms of the average ranks.]{\includegraphics[width=0.45\textwidth]{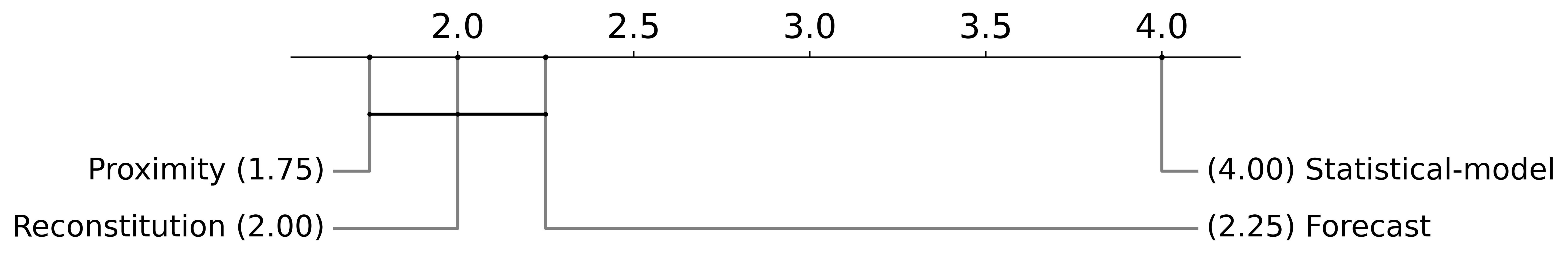}}
    \caption{Performance of four kinds of anomaly detection algorithms under 8 different anomaly metrics. (a) depicts the specific values of four types of algorithms under different evaluation measures, (b) depicts the specific ranks between them, and (c) is concerned with whether there are significant differences between them.
    \label{fig:ranksof4type}}
\end{figure} 

First, we consider the overall situation of four types of algorithms under different accuracy evaluation measures. Specifically, for a given evaluation measure, we first consider the performance of four types of algorithms on each sub-data, take the median of each category of algorithms as the evaluation score for the data, and finally calculate the average scores of all sub-data, as the final score. 
We found that under F1-score, there is no significant difference between the reconstitution and forecast algorithms, which are all significantly better than proximity and statistical-model algorithms, as shown in Fig.~\ref{fig:fourtypes}(a,c).
Under F1-score in Fig.~\ref{fig:fourtypes}(b,d), the proximity-based algorithm is significantly better than the other three algorithms.
The results of more evaluation indicators can be found in Appendix XX.

Considering the difference in rankings under different evaluation measures, we further compared the results of the four types of algorithms under the eight representative evaluation measures, as shown in Fig.~\ref{fig:ranksof4type}(a). As shown in Fig.~\ref{fig:ranksof4type}(b), proximity-based algorithms achieve the best ranking under the eight evaluation measures, while statistical-model-based algorithms perform the worst. At the same time, in the CD diagram in Fig.~\ref{fig:ranksof4type}(c), We can also find that proximity-based algorithms rank first with an average ranking of 1.75, followed by reconstruction-based algorithms with 2.00, prediction-based algorithms with 2.25, and statistical-model-based algorithms with 4.00.

\subsection{Comparison of time required by algorithms}

\begin{figure}[ht]
    \centering
    \subfigure[Time spent by algorithms]
    {\includegraphics[width=0.225\textwidth]{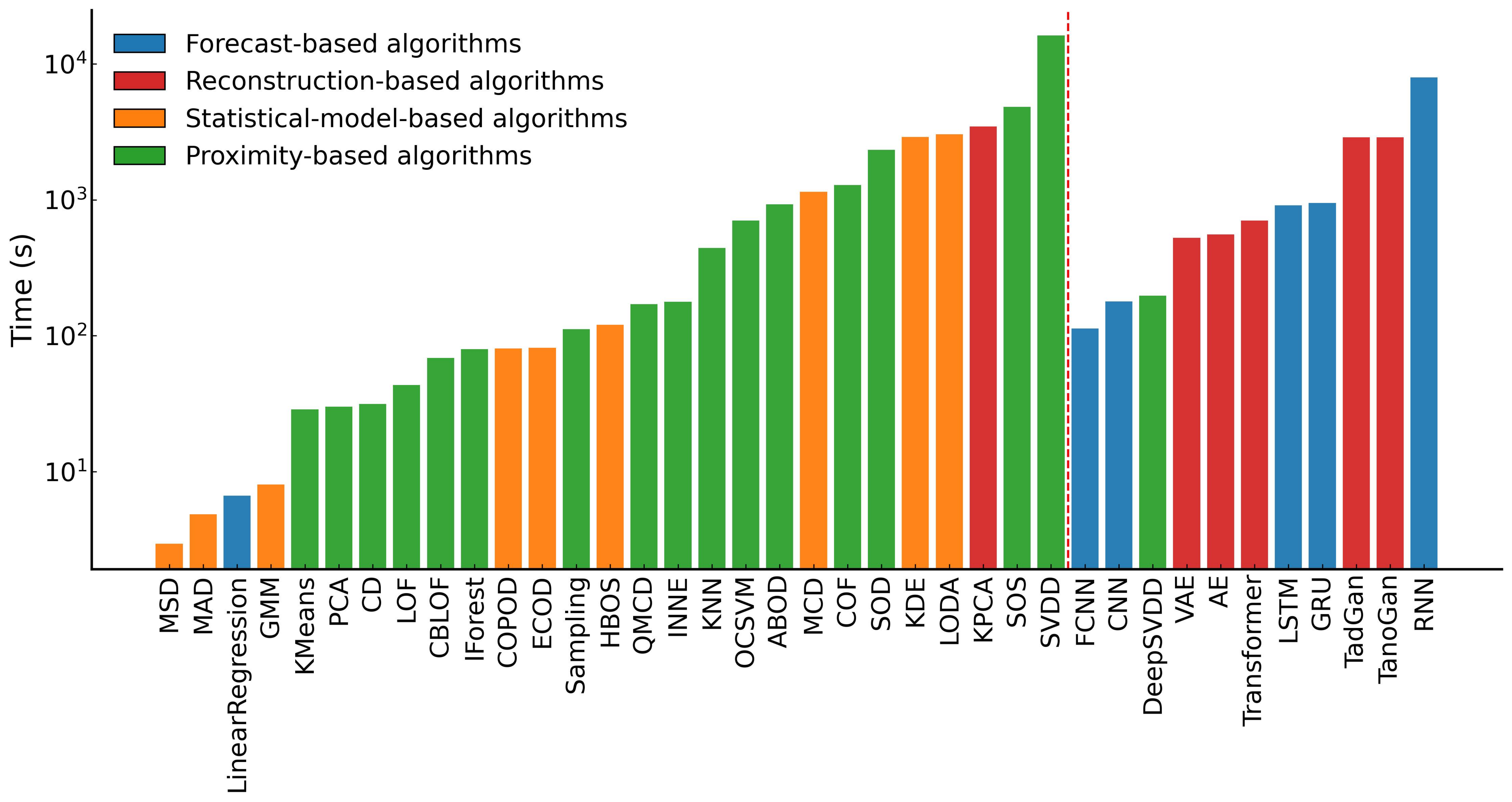}}
    \subfigure[AUC-ROC v.s. time]{\includegraphics[width=0.225\textwidth]{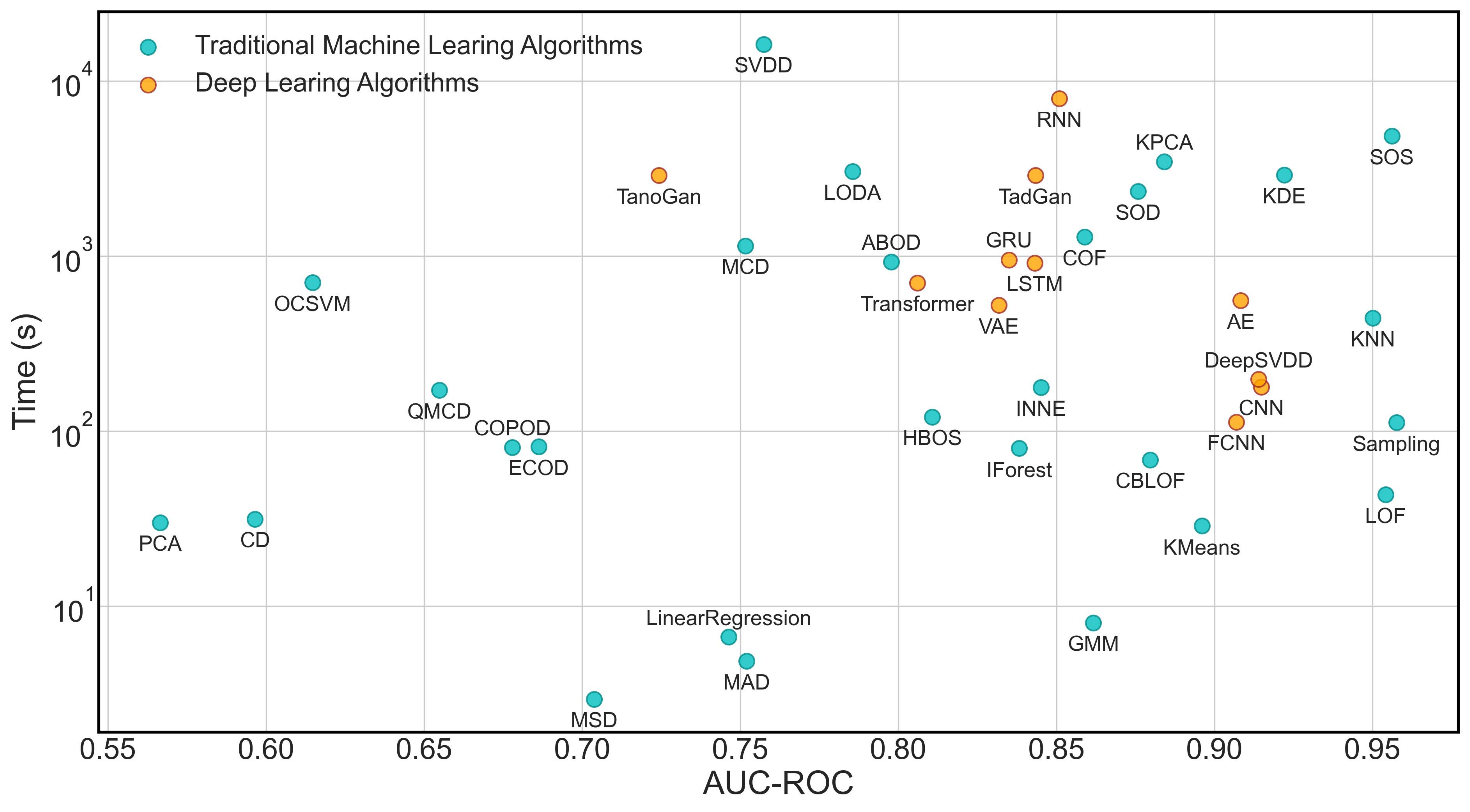}}
    \caption{Comparison of time spent by different algorithms. In (a), the left side of red dotted line represents the traditional machine learning algorithms running on CPU, while the right side represents the deep learning algorithms running on GPU. (b) presents a comparison of AUC-ROC and runtime for different algorithms.
    \label{fig:timerequired}}
\end{figure} 

We also compared the time spent by each algorithm in processing the time series anomaly state detection problem. We randomly selected 3 sets of data from the entire artificial dataset, where each set of data consists of 10 data. We then calculate the time spent on each set of data and take the mean for comparison.

For the non-neural network algorithms, we use the CPU of the same computer for calculation; and for the neural network algorithms, we use the same GPU (a 4080) for calculation. Therefore, we use a red line to distinguish them in Fig.~\ref{fig:timerequired}(a) and compare them separately.

Furthermore, in order to understand the cost-effectiveness between performance and time, we draw a time and AUC-ROC comparison graphs of each algorithm, as shown in Fig.~\ref{fig:timerequired}(b). Among all algorithms with AUC-ROC exceeding 0.95, the LOF algorithm takes the shortest time; and among all algorithms with less than 10 seconds, GMM has the highest AUC-ROC value.

\begin{figure}[ht]
    \centering
    \subfigure
    {\includegraphics[width=0.45\textwidth]{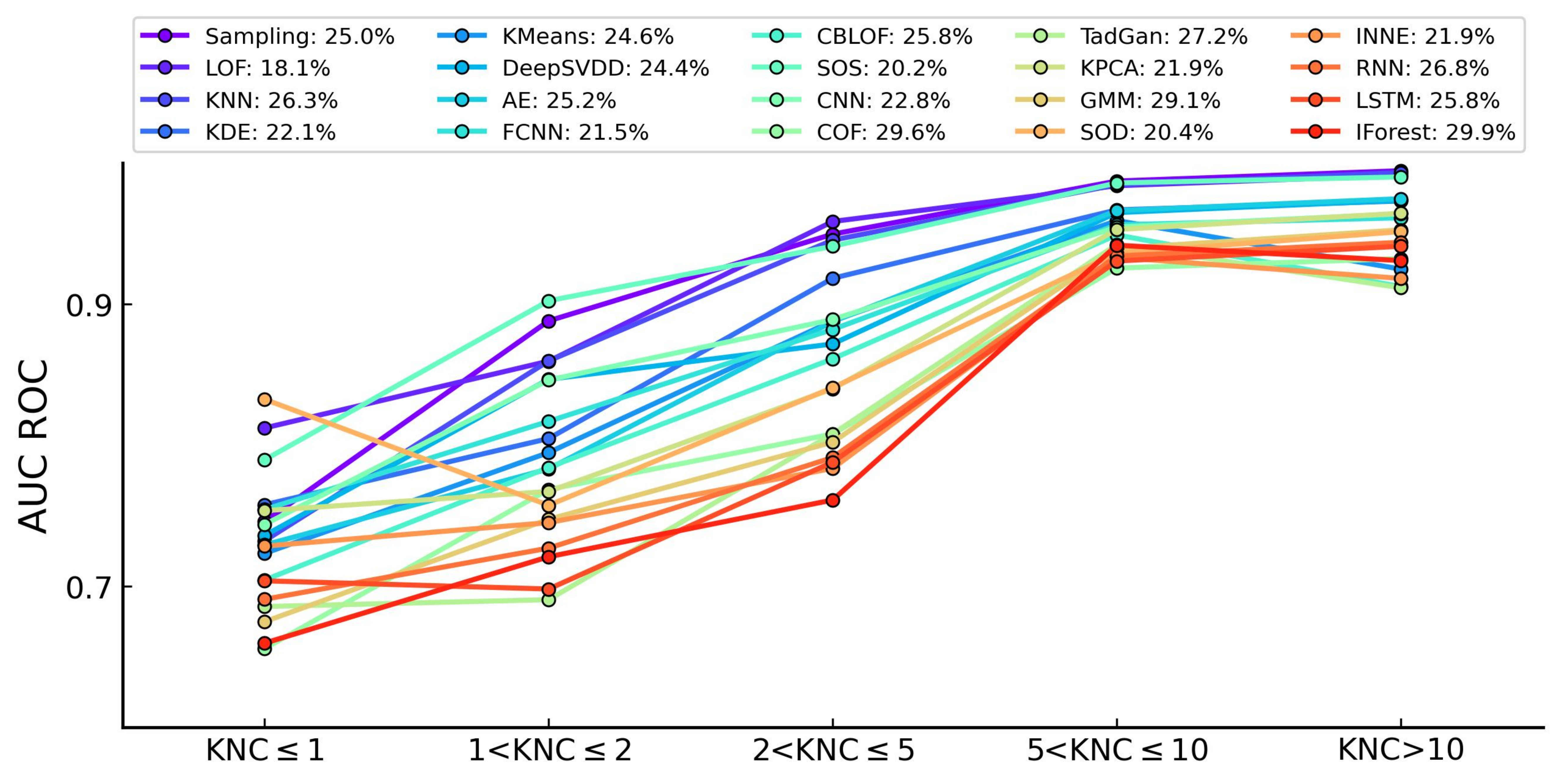}}
    \caption{The performance of top 20 algorithms on AUC-ROC under different KNC.
    \label{fig:algorithmsunderknc}}
\end{figure}

\subsection{The robustness of algorithms under KNC}

In order to observe the robustness of various algorithms, we classify the difficulty of the data set according to KNC and observe the results of each algorithm under different difficulty data.

We first select 20 best-performing algorithms under AUC-ROC and observe their change in AUC-ROC values on the difficulty of the dataset, where all the algorithm achieve a AUC-ROC value of more than 0.9 on the dataset with KNC$>$10. Furthermore, we calculate the decline ratio of each algorithm (i.e. 1 - Min/Max performed at different difficulties). We find that the two most robust algorithms are LOF with 18.1$\%$ and SOS with 20.4$\%$, while the two worst ones are IForest with 29.9$\%$ and GMM with 29.1$\%$. The decline rate of most algorithms is concentrated between 20 and 30 percent. Furthermore, we also analyze the performance and change of each algorithm under different KNC, which is elaborated in Appendix~\ref{sec:moreresultsknc}. 

\section{Acknowledgments}

This work is sponsored by the National Key R$\&$D Program of China Grant No. 2022YFA1008200 (Z. X.), the Shanghai Sailing Program, the Natural Science Foundation of Shanghai Grant No. 20ZR1429000 (Z. X.), the National Natural Science Foundation of China Grant No. 62002221 (Z. X.), Shanghai Municipal of Science and Technology Major Project No. 2021SHZDZX0102, and the HPC of School of Mathematical Sciences and the Student Innovation Center, and the Siyuan-1 cluster supported by the Center for High Performance Computing at Shanghai Jiao Tong University.

% We also discussed the difference between time series outlier analysis and anomaly state detection in Appendix XXX.


\bibliography{icml2024}
\bibliographystyle{icml2024}

%%%%%%%%%%%%%%%%%%%%%%%%%%%%%%%%%%%%%%%%%%%%%%%%%%%%%%%%%%%%%%%%%%%%%%%%%%%%%%%
%%%%%%%%%%%%%%%%%%%%%%%%%%%%%%%%%%%%%%%%%%%%%%%%%%%%%%%%%%%%%%%%%%%%%%%%%%%%%%%
% APPENDIX
%%%%%%%%%%%%%%%%%%%%%%%%%%%%%%%%%%%%%%%%%%%%%%%%%%%%%%%%%%%%%%%%%%%%%%%%%%%%%%%
%%%%%%%%%%%%%%%%%%%%%%%%%%%%%%%%%%%%%%%%%%%%%%%%%%%%%%%%%%%%%%%%%%%%%%%%%%%%%%%
\newpage
\appendix
\onecolumn

\section{Appendix for Time Series Anomaly State Detection}

\subsection{Concrete operations of training stage and testing stage for implementation framework}
\label{sec:concreteoperation}

\begin{algorithm}[!h]
\caption{Standard training procedure for time series anomaly state detection}
\begin{algorithmic}[1]
\label{standard training procedure}

\STATE \textbf{Input: } standard time series data $X_{standard} \in \mathbb{R}^{n_0 \times d_0}$, where $n_0$ is the length and $d_0$ is the dimension of features, feature extraction function $f: X \to Y$, time series anomaly state detection algorithm functions $\mathcal{H}: Y \to A$, de-windowing function $\omega: A \to H$, and decision function $\mathcal{D}$.

\STATE \textbf{Data Partition:} Partition standard time series data $X_{standard}$ into training time series data $X_{train} \in \mathbb{R}^{n_1 \times d_0}$ and validation time series data $X_{validation} \in \mathbb{R}^{n_2 \times d_0}$, where $n_0 = n_1 + n_2$.

\STATE \textbf{Feature Extraction:} Learn and extract features of training time series $X_{train}$ through feature extraction functions $f$ to obtain the training feature window data $Y_{train} = f_{train} (X_{train}) \in \mathbb{R}^{n_3 \times d_1}$, where $n_3$ is the number of feature windows and $d_1$ is the dimension of features after pre-processing. And then, Use $f_{train}$ to extract the features of validation time series data $X_{validation}$ and obtain the validation feature window data $Y_{validation} = f_{train} (X_{validation}) \in \mathbb{R}^{n_4 \times d_1}$, where $n_4$ is the number of feature windows.

\STATE \textbf{Model Training:} Use the training feature window data $Y_{train}$ to train time series anomaly state detection algorithm functions $\mathcal{H}$, and get the trained algorithm $\mathcal{H}_{train}$.

\STATE \textbf{Model Inference:} Calculate the anomaly score $A_{validation} = \mathcal{H}_{train}(Y_{validation}) \in \mathbb{R}^{n_4 \times 1}$ of the validation feature window data $Y_{validation}$ under the trained algorithm function $\mathcal{H}_{train}$. 

\STATE \textbf{De-windowing:} The de-windowing function $\omega$ is utilized to map the validation anomaly score $A_{validation}$ back to the length of the validation time series $X_{validation}$ to align with the original sequence, and get the modified validation anomaly score $ \hat{A}_{validation} = \omega(A_{validation}) \in \mathbb{R}^{n_2 \times 1}$.

\STATE \textbf{Learning Decision Function:} Use the modified validation anomaly score $ \hat{A}_{validation}$ to learn the decision function $\mathcal{D}$, and get the learned decision function $\mathcal{D}_{validation}$.

% \IF {$n < 1$}                        %条件语句
% \PRINT {Input Error}                 %打印语句
% \ELSE
%     \FOR {$i = 0$ to n}          %FOR循环结构
%     \STATE $Sum = Sum + i$\\
%     \STATE $i = i + 1$
%     \ENDFOR
% \ENDIF
% \RETURN Sum
% \STATE \textbf{Output: }

\end{algorithmic}
\end{algorithm}

\begin{algorithm}[!h]
\caption{Standard testing procedure for time series anomaly state detection}
\begin{algorithmic}[1]
\label{standard testing procedure}

\STATE \textbf{Input: } test time series data $X_{test} \in \mathbb{R}^{n_5 \times d_0}$, where $n_5$ is the length and $d_0$ is the dimension of features, trained feature extraction function $f_{train}$, trained time series anomaly state detection algorithm function $\mathcal{H}_{train}$, de-windowing function $\omega$, trained decision function $\mathcal{H}_{validation}$, and decision threshold $T$.

\STATE \textbf{Feature Extraction:} Use the trained feature extraction function $f_{train}$ to process the test time series $X_{test}$ to obtain the testing feature window data $Y_{test} = f_{train} (X_{test}) \in \mathbb{R}^{n_6 \times d_1}$.

\STATE \textbf{Model Inference:} Use the trained algorithm function $\mathcal{H}_{train}$ to calculate the test anomaly score $A_{test}= \mathcal{H}_{train}(Y_{test}) \in \mathbb{R}^{n_6\times 1} $ of the testing feature window data $Y_{test}$.

\STATE \textbf{De-windowing:} The de-windowing function $\omega$ is utilized to map the test anomaly score $A_{test}$ back to the length of the test time series $X_{test}$ to align with the original sequence, and get the modified test anomaly score $ \hat{A}_{test} = \omega(A_{test}) \in \mathbb{R}^{n_5 \times 1}$.

\STATE \textbf{Inferring Health Indicator Series:} Use the trained decision function $\mathcal{D}_{train}$ to calculate the test health indicator series $H_{test}=\mathcal{D}_{train}(\hat{A}_{test})$ of the test time series $X_{test}$ 

\STATE \textbf{Determining Anomaly:} Use decision threshold $T$ to determine whether an anomaly occurs in $X_{test}$, and finally get the healthy index $\hat{H}_{test}$. For a given point $h_i \in H_{test}$ in the test health indicator series $H_{test}$: if $h_i < T$, then the corresponding original test time series point $x_i \in X_{test}$ is considered as the inlier, and has $\hat{h}_{i}=0$; else if $h_i \geq T$, then the corresponding original test time series point $x_i \in X_{test}$ is considered as the outlier, and has $\hat{h}_{i}=1$.

\end{algorithmic}
\end{algorithm}

\subsection{Intuitive analysis for outliers under the implementation framework}
\label{sec:intuitiveanalysis}

A question of concern is, what do the anomalies we find look like in the implementation framework? When we find a point $h_0 = d_{train}(\hat{a}_0) \geq T, h_0 \in H_{test}, \hat{a}_0 \in A_{test}$, then it means that $\hat{a}_0$ is relative large among $\hat{A}_{validation}$. Therefore, $A_0 = w^{-1}(\hat{a}_0) = h^{*}(Y_0) $ is also relative large among $A_{validation}$, which infers that under trained algorithm function $h^{*}$, $Y_0$ is far away from all the feature windows in $Y_{validation}$.

In other words, if the "distance" $d_{test}$ between a test feature window $Y_0 \in Y_{test}$ and its closest training windows $Y_{train,0}$ compared to the average "distance" $\bar{d}_{validation}$ between the validation windows $Y_{validation}$ and the training windows $Y_{train}$ is far, then it will result in $a_0=h^{*}(Y_0)$ to be too large, and ultimately lead to $h_0=d_{train}(\hat{a}_0)$ being too large. Then we consider the test feature window $Y_0$ to be abnormal, and its corresponding original time series segment $X_0$ to be abnormal.

Therefore, if a pattern segment in the test time series has never appeared in the standard time series, then it will easily lead to the test feature window $Y_0 \in Y_{test}$ containing this pattern far away from the validation feature windows $Y_{validation}$ and makes the corresponding health indicator exceed the decision threshold $T$. Therefore, we define anomalies in the time series anomaly state detection problem as follows.

\textbf{Intuitive Definition for Outliers under specific implementation:} If a pattern in test time series does not appear in standard time series, then we call this pattern an anomaly.

\textbf{Formal Definition for Outliers under specific implementation:} Given a suitable threshold $\epsilon$, a distance function $d(X,Y)$ and a sequence length $n_0$, if there exists a dimension $d_0$ of a test time series data segment $ \hat{X}_{0, d_0}^{test} \in \mathbb{R}^{n_0}$, such that $ \hat{X}_{0,d_0}^{test}$ is inconsistent with any segment of this dimension $d_0$ in the standard time series data segments $\{ \hat{X}_{i,d_0}^{standard} \in \mathbb{R}^{n_0} \}_{i\in \mathbb{N}} $ with the same length $n_0$, that is, $d(\hat{X}_{0,d_0}^{test},\hat{X}_{i,d_0}^{standard})>\epsilon$ for all $i \in \mathbb{N}$ , then it is said that under the threshold $\epsilon$, the distance function $d(X,Y)$ and the sequence length $n_0$, the test data segment$ \hat{X}_{0}^{test}$  is abnormal from the standard time series data.

Note that the sequence length $n_0$ here is similar to the process of windowing, which is a very important parameter in the time series. Such a definition is also consistent with the mathematical definition in Sec.~\ref{sec:mathematicaldefinition}. Specifically, in mathematical definition, we believe that outliers are those segments that do not follow the standard random process, $ \mathcal{O} = \{ S_j^k  \subset  \{  \hat{x}_j (t), t \in [\hat{N}_j]  \} \in \mathcal{U}: S_j^k \nsim \{X_0(t),t \in \mathbb{N}\}  \}$, so in specific practice, we identify those test time series segments that are far away from the standard time series as ones not obeying the standard random process, which is very reasonable and consistent.

\newpage
\section{Appendix for preparation for experiments}

\subsection{The time series anomaly state detection algorithms list}
\label{sec:algorithmslist}

\textbf{Forecast-based methods} refer to time series anomaly detection approaches , wherein predictive models are trained to anticipate future values first, and subsequently, identify the anomaly by assessing the disparity between the predicted values and the actual observed values. This method constitutes the prevailing and prevalent technique within the realm of time series anomaly detection.
LinearRegreesion(LR) utilizes a high-dimensional linear model to act as a prediction function and assumes that each moment is linearly related to several previous moments.
Fully connected neural networks (FCNN), Convolutional neural networks (CNN), Recurrent Neural Network (RNN), Gated Recurrent Unit (GRU), and Long Short-Term Memory (LSTM) respectively use corresponding neural networks structure to serve as prediction models and analyse the nonlinear temporal correlations between data samples.
It is worth mentioning that the model design of RNN, GRU and LSTM makes them further take into account the time correlation of the data itself, which is more in line with the characteristics of time series data.

\textbf{Reconstitution-based methods} aim to learn a model that captures the latent structure of the given time series data and generate synthetic reconstructions, and then discriminate the anomaly by reconstruction performance under assumption that outliers cannot be efficiently reconstructed from the low-dimensional mapping space.
Auto-Encoder (VAE) and Variational Auto-Encoder (VAE) use the Encoder neural network to map the time series data to low-dimensional latent space and normal distribution respectively, and then use the Decoder neural network to reconstruct the original data. Kernel Principal Component Analysis (KPCA) maps data to the feature space generated by the kernel and uses the reconstruction error on the feature space to determine anomalies. Transformer uses the Transformer structure to compress and reconstruct data, whose main feature is that it has a special attention structure to consider the temporal correlation of data.
Time Series Anomaly Detection with Generative Adversarial Networks (TanoGan) and Time Series Anomaly Detection Using Generative Adversarial Networks (TadGan) are both algorithms based on adversarial generative neural networks, and use LSTM as the generator and discriminator model to gradually learn the real data distribution through adversarial generation. However, during the reconstruction process, TanoGan uses the residual loss and discriminant loss between the real sequence and its closest generated sequence to identify anomalies; while TadGan uses the cyele consistency loss of the time series and the value of its discriminator to distinguish anomalies.

\textbf{Statistical-model-based methods} assume that the data is generated by a certain statistical distribution, and use statistical models to perform hypothesis testing to determine which data does not meet the model assumptions, thereby detecting outliers.
Copula-Based Outlier Detection (COPOD)  is a non-parametric method, which obtain the empirical copula through the empirical cumulative distribution , and then estimate the tail probability of the joint distribution in all dimensions. The Empirical Cumulative Distribution-Based Outlier Detection (ECOD) method undertakes the computation of the empirical cumulative distribution for each dimension within the training dataset, and Subsequently, amalgamates the tail probabilities associated with each dimension as a means of deriving the anomaly score. Through the representation of data as a composite of Gaussian components, Gaussian Mixture Models (GMMs) have the capacity to distinguish anomalies by detecting data points that exhibit substantial deviations from the acquired distribution. Histogram- based outlier detection (HBOS) assumes feature independence and utilizes histograms to measure data point outlierness, where the inverse of the bin height serves as the outlier score. Kernel density estimation (KDE) is a non-parametric method that applies kernel smoothing to probability density estimation, that is, using the kernel as a weight to estimate the probability density function of random variables, and then calculates the density estimate of the data points and compares it with the threshold Compare to identify outliers. Lightweight on-line detector of anomalies (LODA) utilizes one-dimensional histograms constructed on sparse random projections, where anomaly score is a negative average log probability estimated from histogram on projections. Random sparse projects allow LODA to use simple one-dimensional histograms, thus processing large datasets in relatively small time complexity. Mean Absolute Differences (MAD) Algorithm determines whether an anomaly has occurred by calculating the median of the absolute deviation from the data point to the median. Outlier Detection with Minimum Covariance Determinant (MCD) is a powerful method for detecting outliers in multivariate data by estimating a robust covariance matrix and using Mahalanobis distances to quantify the outlier degree of each data point.

\textbf{Proximity-based methods} for anomaly detection are techniques that primarily rely on using various distance measures to quantify the similarity or proximity between data points, including density, distance, angle and dividing hyperplane, and then identify anomalies, under the assumption that the distribution of abnormal points is different from that of normal points, causing the similarity is low. Angle-based outlier detection (ABOD) considers the variance of an observation's weighted cosine score relative to all its neighbors as its outlying score. Cluster-Based Local Outlier Factor (CBLOF) first use a clustering algorithm to divide the data into large clusters and small clusters, and then calculate the distance from each point to the nearest large cluster as anomaly score. Cook's distance outlier detection (CD) measure the influence of observations on a linear regression, and identify instances with large influence as outliers. Connectivity-Based Outlier Factor (COF) calculates the ratio of the average chaining distance of an oberservation to the average chaining distance of its k-th nearest neighbor as the anomaly score. Deep One-Class Classification for outlier detection (DeepSVDD)  minimize the volume of hyper-sphere that enfolds the sample feature space represented by a neural network, and then calculate the distance between the center of the sphere and the sample as anomaly score. Isolation forest (IForest) constructs a random forest of binary trees to isolate the data points and identifies data points with shorter average path lengths to the root as anomalies.
Isolation-based anomaly detection using nearest-neighbor ensembles (INNE) uses a multi-dimensional hypersphere to cut the data space to implement the isolation mechanism, and takes into account the local distribution characteristics of the data, the nearest neighbor distance ratio, to calculate anomaly indicators of the data. The Kmeans outlier detector (Kmeans) first finds K clusters through iteration, and then calculates the distance between the observation and the center of its nearest cluster as an outlier score. K-Nearest Neighbors (KNN) consider the anomaly score of the observation as the distance to its nearest k neighbors. Local Outlier Factor (LOF) measures the local deviation in the density of a particular data point relative to that of its neighboring points, where the local density around a given data point is estimated by the distances to its k nearest neighboring data points. One-class SVM (OCSVM) aims to find a separating hyperplane that best separates the origin from the majority of normal data points, which is also defined as the decision boundary. The PCA algorithm (PCA) maps the data onto a low-dimensional hyperplane generated by the principal components of the data, and treats points far away from this hyperplane as anomalies. Quasi-Monte Carlo Discrepancy outlier detection (QMCD) uses Wrap-around Quasi-Monte Carlo Discrepancy, a uniformity criterion which is used to assess the space filling of a number of samples in a hypercube, to calculate the discrepancy values of the sample as anomaly scores. Outlier detection based on Sampling (SP) first randomly and independently sample a subset of the data, and then view the distance between the observation and the subset as anomaly score.
Subspace Outlier Detection (SOD) explores a axis-parallel subspace spanned by the data object’s neighbors and calculates the degree how much it deviates from the neighbors in this subspace as the anomaly score. Stochastic outlier selection (SOS) employs the concept of affinity to quantify the relationship of one data point to another, where a data point is considered to be an outlier when the other data points have insufficient affinity with it.
One-Classification using Support Vector Data Description (SVDD) establish a hyper-sphere with the smallest radius in a high-dimensional space, and then calculating the distance between the sample point and the hyper-sphere as anomaly score.

\subsection{Details for relative contrast, normalized  clusteredness, and normalized adjacency}
\label{sec:rcncna}

Previous articles have discussed several indicators to measure the difference in distribution of normal points and abnormal points, including Relative Contrast (RC)~\cite{he2012difficulty}, which is defined as the ratio of the expectation of the mean distance to the expectation of nearest neighbor distance for all data points, Normalized clusteredness of abnormal points (NC)~\cite{emmott2013systematic}, which is the ratio of the average SBD of normal subsequences to the average SBD of abnormal subsequences, and Normalized adjacency of normal/abnormal cluster (NA)~\cite{paparrizos2022tsb}, which is defined as the ratio of the minimum distance between the centroids of normal and abnormal clusters to the average distance among the centroids of all normal clusters.

\begin{equation}
\begin{aligned}
& R_c=\frac{\mathbb{E}_{s \in S}\left[D_{\text {mean}}(s)\right]}{\mathbb{E}_{s \in S}\left[D_{\min }(s)\right]}, \
N_c=\frac{\mathbb{E}_{s_i, s_j \in S_{\text {nor}, i \neq j}}\left[D\left(s_i, s_j\right)\right]}{\mathbb{E}_{s_i, s_j \in S_{\text {ano }, i \neq j}}\left[D\left(s_i, s_j\right)\right]}, \ N_a=\frac{\min _{c_i \in C_{a n o}, c_j \in C_{n o r}} D\left(c_i, c_j\right)}{\mathbb{E}_{c_i, c_j \in C_{n o r}, i \neq j}\left[D\left(c_i, c_j\right)\right]}
\end{aligned}
\end{equation}

where $s$ represents time series, $D(s_i,s_j)$ is denoted as the SBD distance between two series $s_i$ and $s_j$, $S$ is the whole set, $D_{\min }(x)=\min _{s \in S-x} D(x, s)$ and $D_{\text {mean }}(x)=\mathbb{E}_{s \in S-x} D(x, s)$ are the 1-NN distance and mean distance fot series $x$, $S_{n o r}$ and $S_{a n o}$ are denoted as the set of normal sequences and the set of anomalous sequence, and $C_{n o r}$ and $C_{a n o}$ are denoted as the set of centroids of normal clusters and the set of centroids of anomalous clusters. 

\newpage
\section{Appendix for experiments}

\subsection{More results for performance of algorithms under eight representative accuracy evaluation measures}
\label{sec:moreperformancebox}
\begin{figure}[htb]
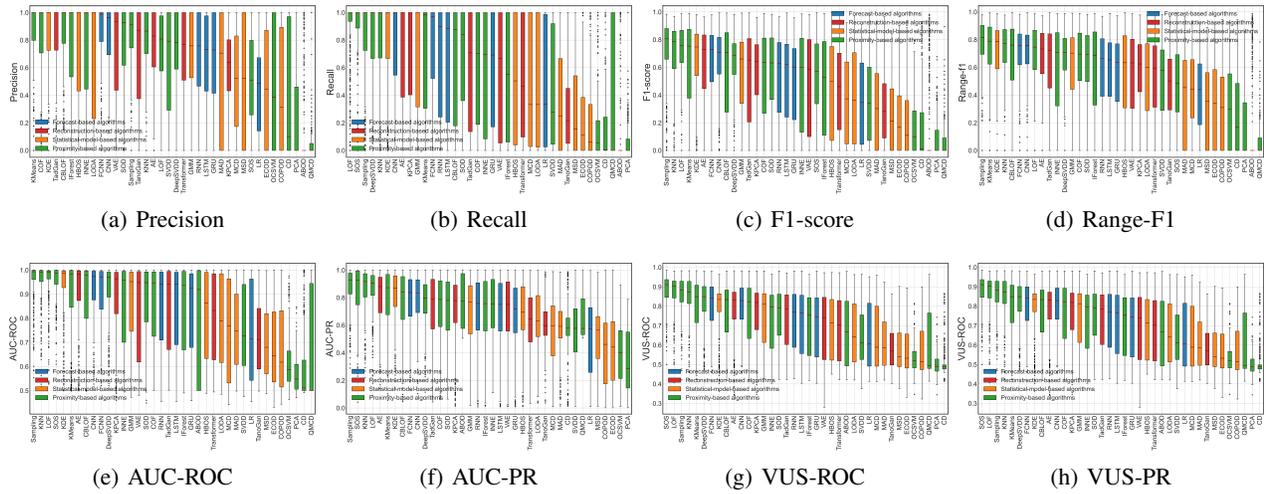

    \centering

    \subfigure[Precision]{\includegraphics[width=0.24\textwidth]{figures/appendix/boxplot-ranks-precision.pdf}}
    \subfigure[Recall]
    {\includegraphics[width=0.24\textwidth]{figures/appendix/boxplot-ranks-recall.pdf}}
    \subfigure[F1-score]{\includegraphics[width=0.24\textwidth]{figures/appendix/boxplot-ranks-f1-score.pdf}}
    \subfigure[Range-F1]
    {\includegraphics[width=0.24\textwidth]{figures/appendix/boxplot-ranks-range-f1.pdf}}

    \subfigure[AUC-ROC]{\includegraphics[width=0.24\textwidth]{figures/appendix/boxplot-ranks-auc-roc.pdf}}
    \subfigure[AUC-PR]
    {\includegraphics[width=0.24\textwidth]{figures/appendix/boxplot-ranks-auc-pr.pdf}}
    \subfigure[VUS-ROC]{\includegraphics[width=0.24\textwidth]{figures/appendix/boxplot-ranks-vus-roc.pdf}}
    \subfigure[VUS-PR]
    {\includegraphics[width=0.24\textwidth]{figures/appendix/boxplot-ranks-vus-roc.pdf}}

    \caption{Performance of algorithms under eight representative accuracy evaluation measures. The sub-caption of (a-h) indicate the evaluation measures used. 
    % \label{fig:isw2}
    }
\end{figure} 

\subsection{More results for critical differnece of algorithms under eight representative accuracy evaluation measures}
\label{sec:moreperformancecd}
\begin{figure}[htb]
    \centering
    \subfigure[Precision]{\includegraphics[width=0.24\textwidth]{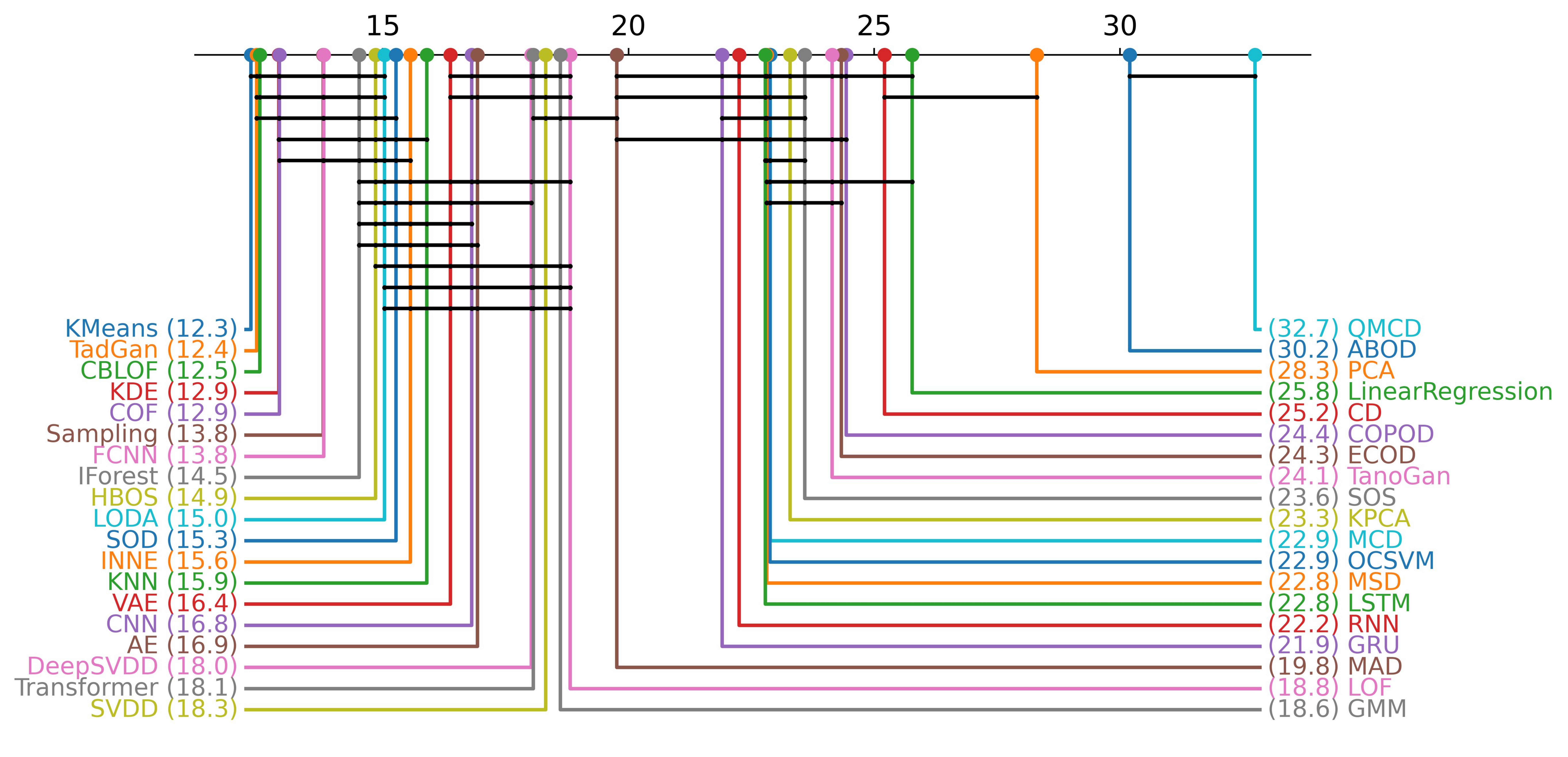}}
    \subfigure[Recall]
    {\includegraphics[width=0.24\textwidth]{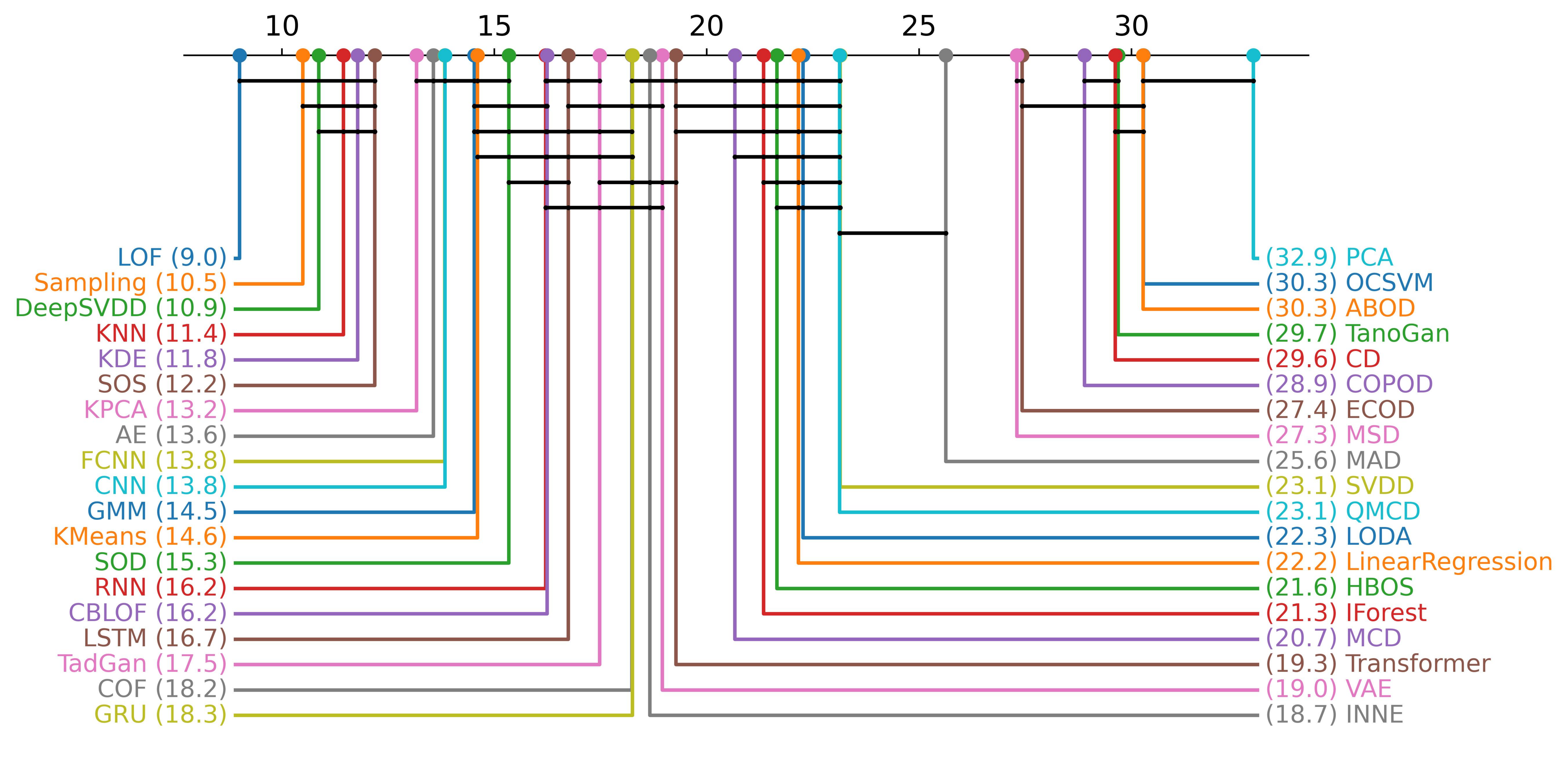}}
    \subfigure[F1-score]{\includegraphics[width=0.24\textwidth]{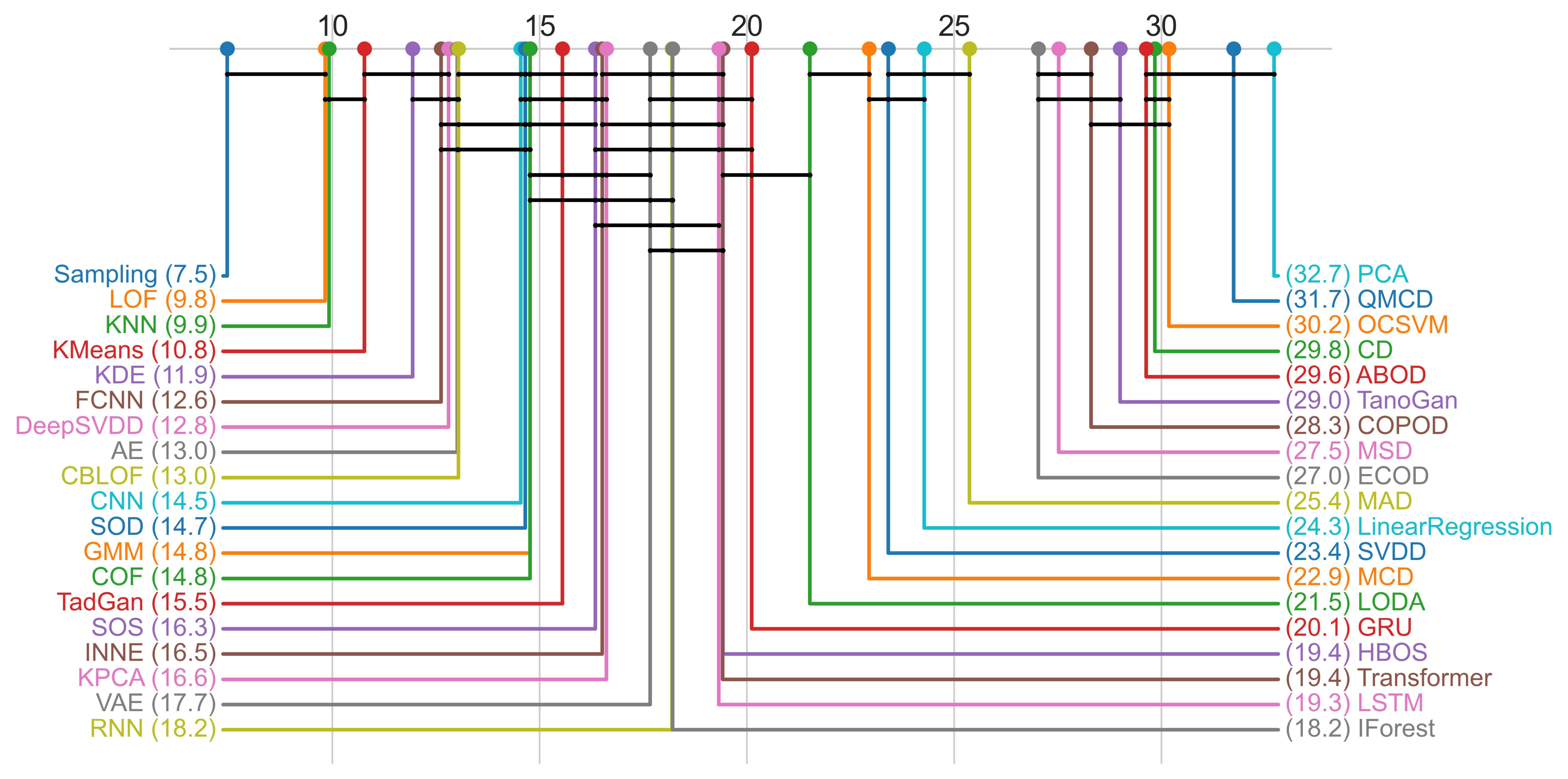}}
    \subfigure[Range-f1]
    {\includegraphics[width=0.24\textwidth]{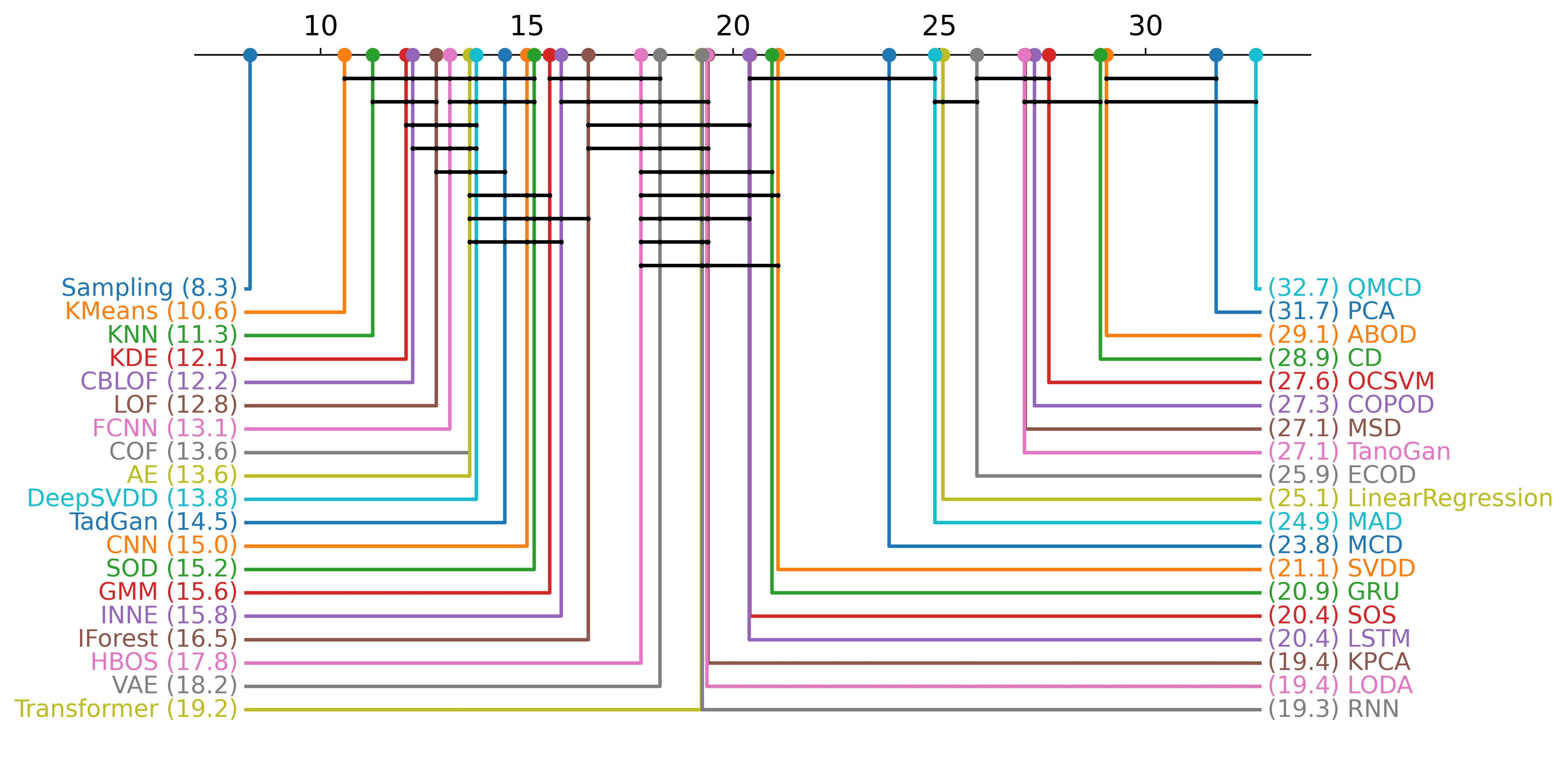}}

    \subfigure[AUC-ROC]{\includegraphics[width=0.24\textwidth]{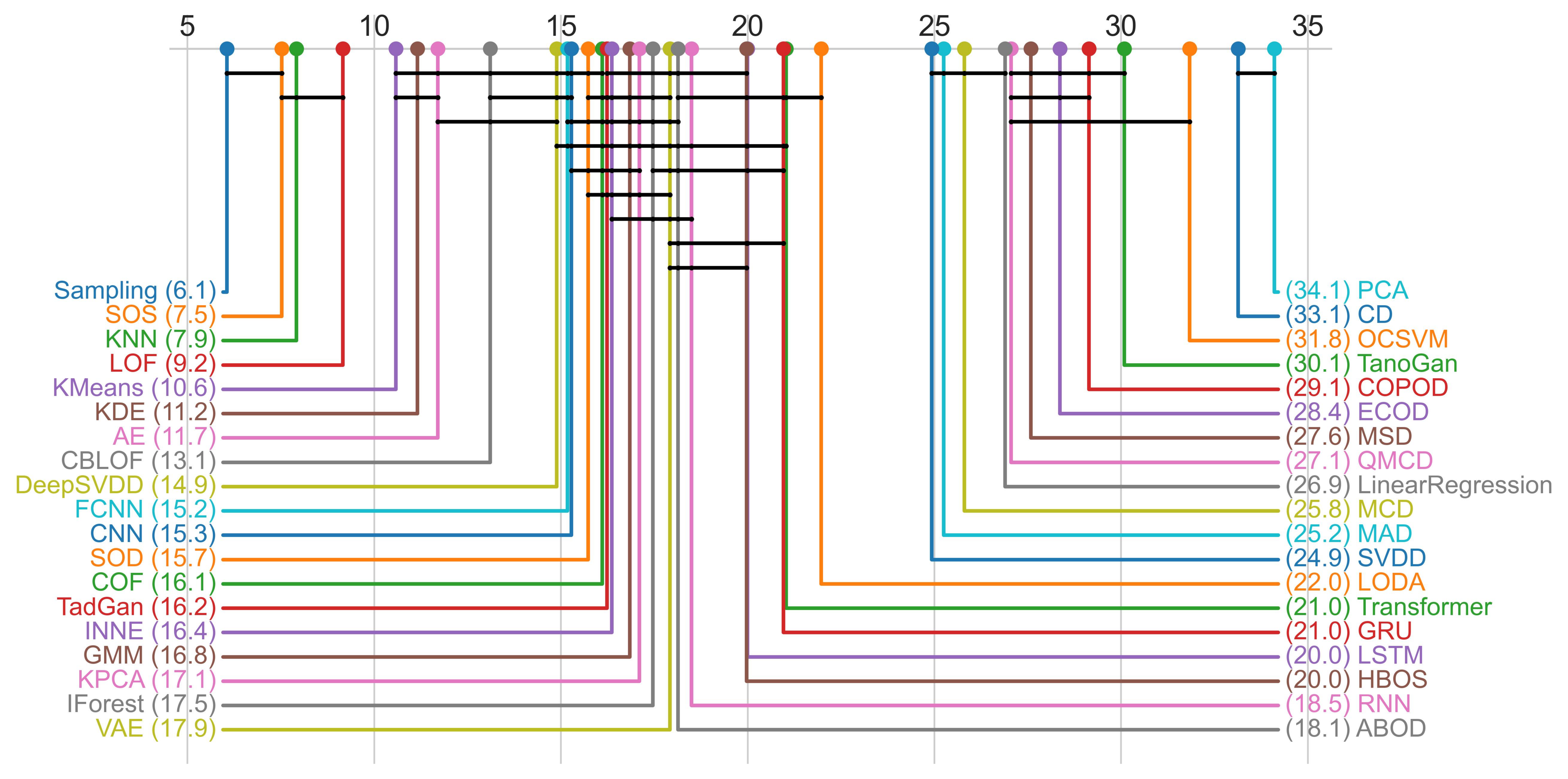}}
    \subfigure[AUC-PR]
    {\includegraphics[width=0.24\textwidth]{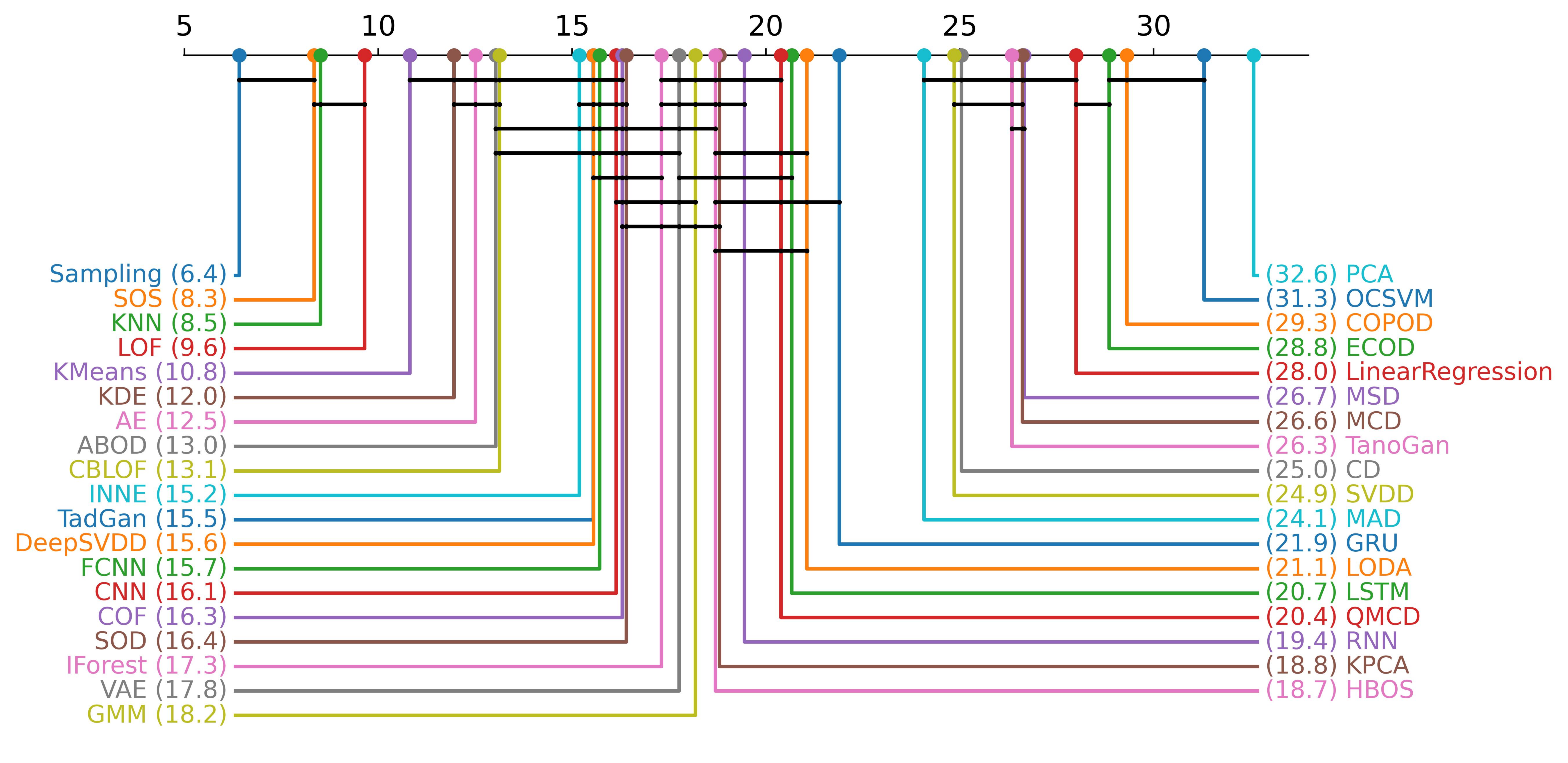}}
    \subfigure[VUS-ROC]{\includegraphics[width=0.24\textwidth]{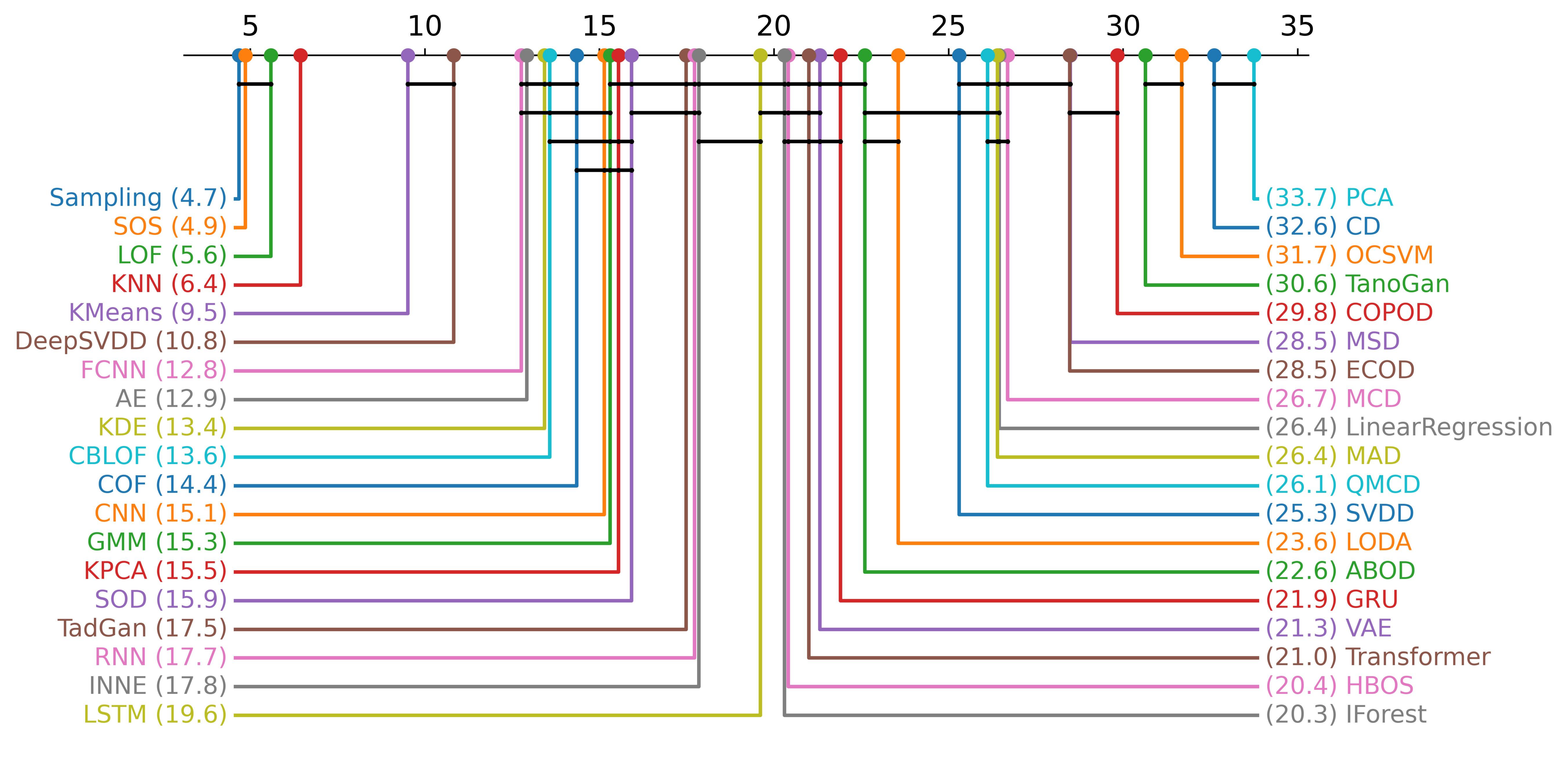}}
    \subfigure[VUS-PR]
    {\includegraphics[width=0.24\textwidth]{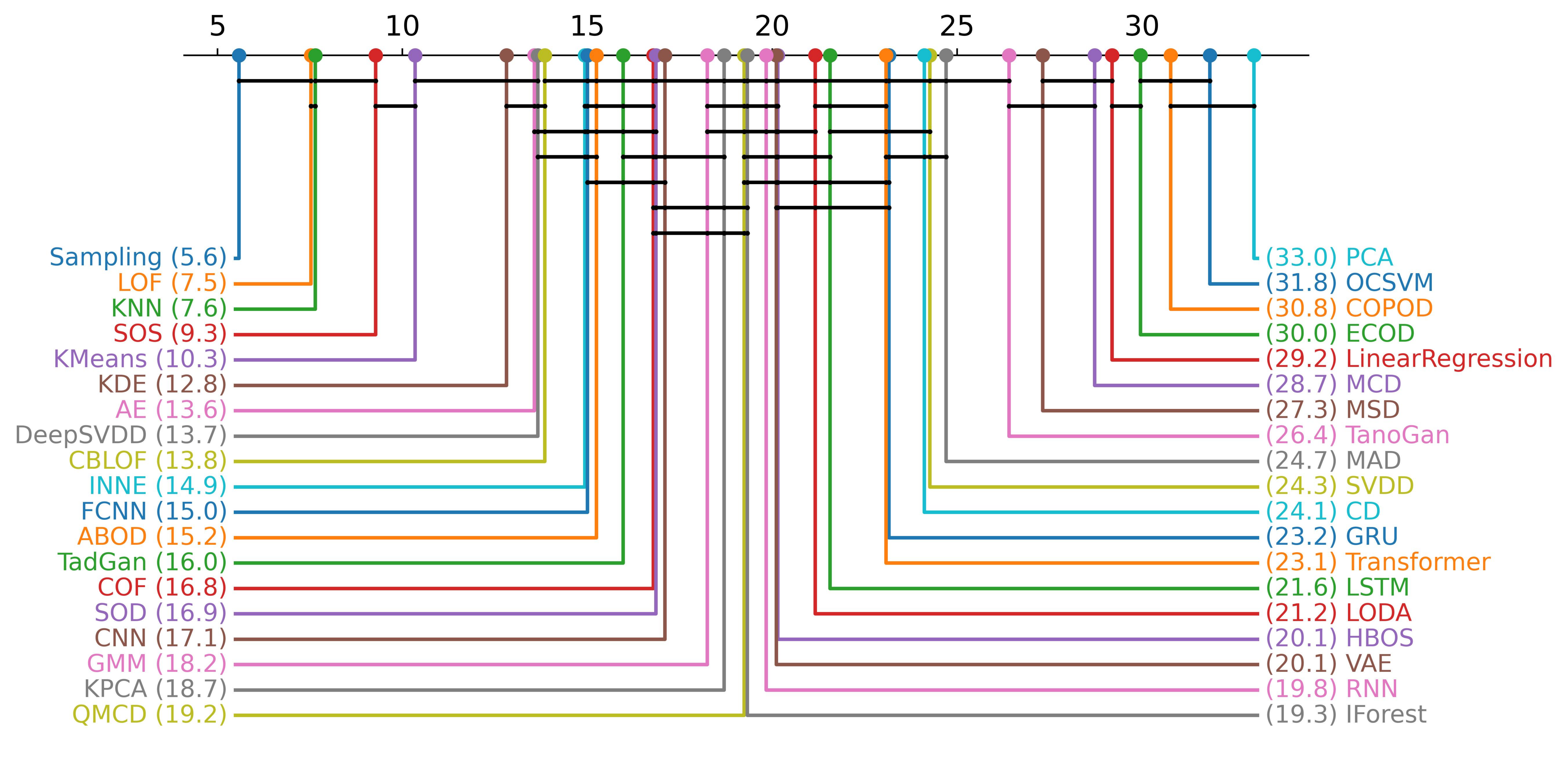}}

    \caption{Critical Difference of algorithms under eight representative accuracy evaluation measures. The sub-caption of (a-h) indicate the evaluation measures used.
    % \label{fig:isw2}
    }
\end{figure} 

\newpage
\subsection{More results for performance of four types of algorithms under eight representative accuracy evaluation measures}
\begin{figure}[htbp]
    \centering
    \subfigure[Precision]{\includegraphics[width=0.18\textwidth]{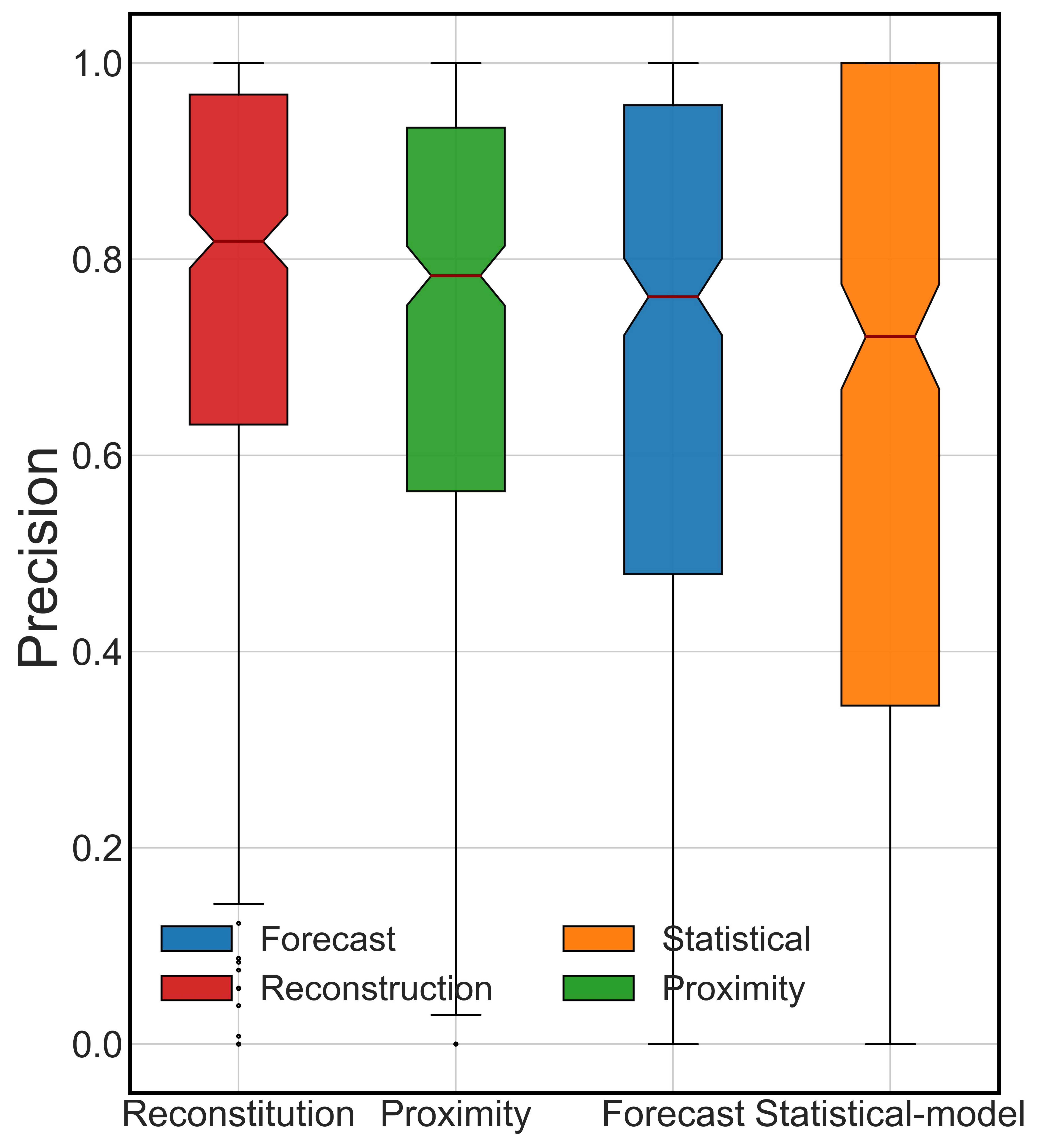}}
    \subfigure[Recall]
    {\includegraphics[width=0.18\textwidth]{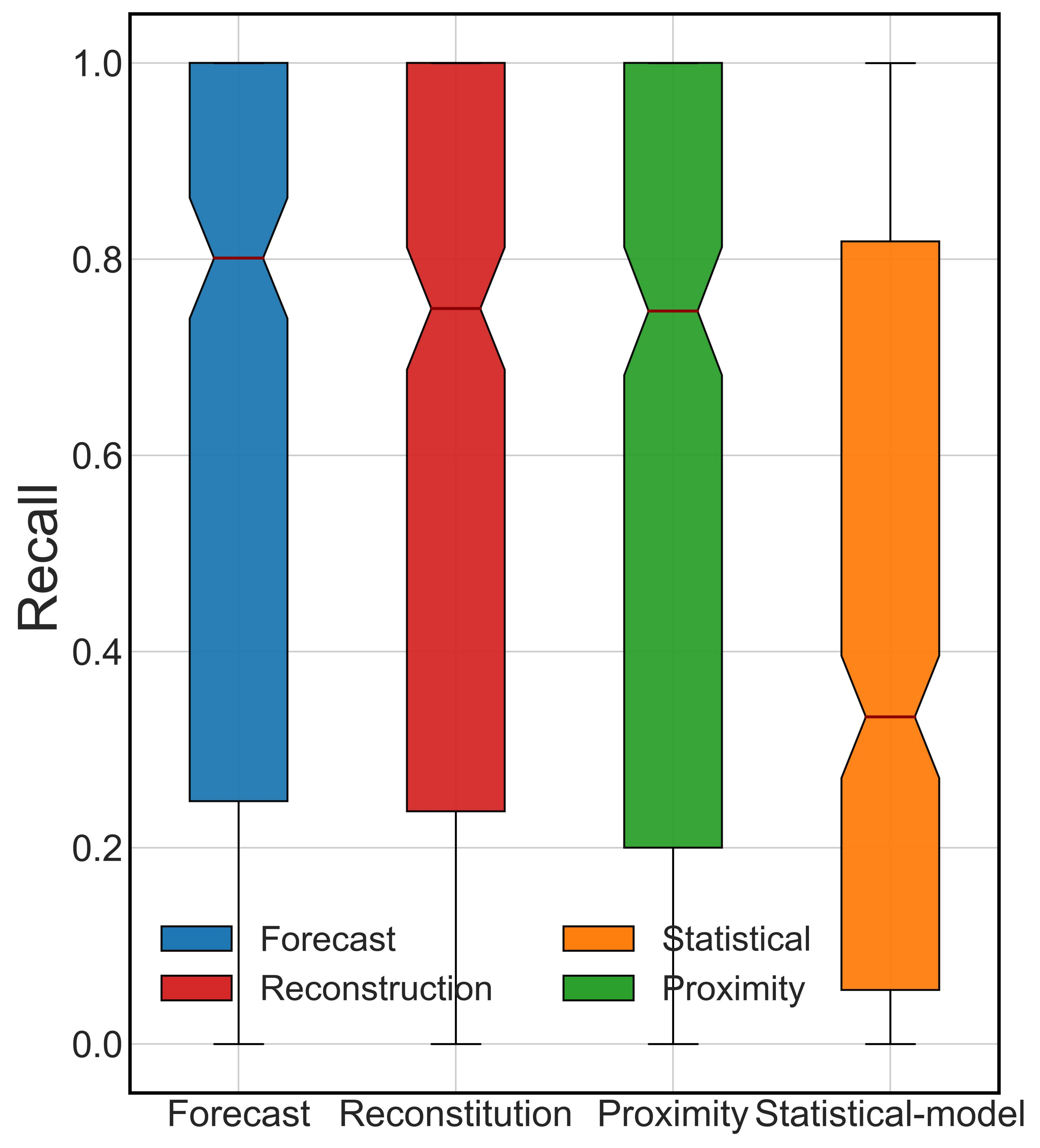}}
    \subfigure[F1-score]{\includegraphics[width=0.18\textwidth]{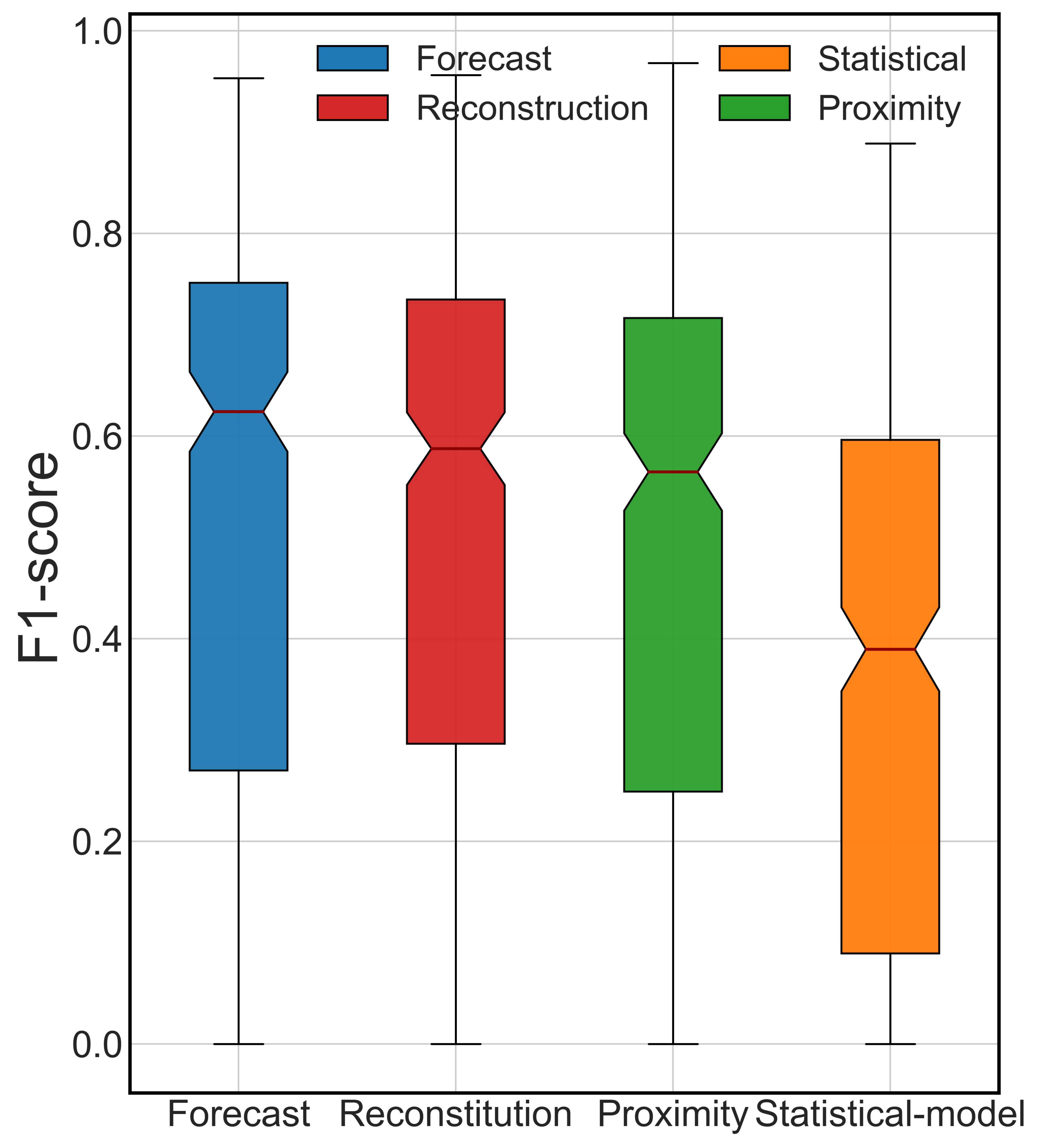}}
    \subfigure[Range-f1]
    {\includegraphics[width=0.18\textwidth]{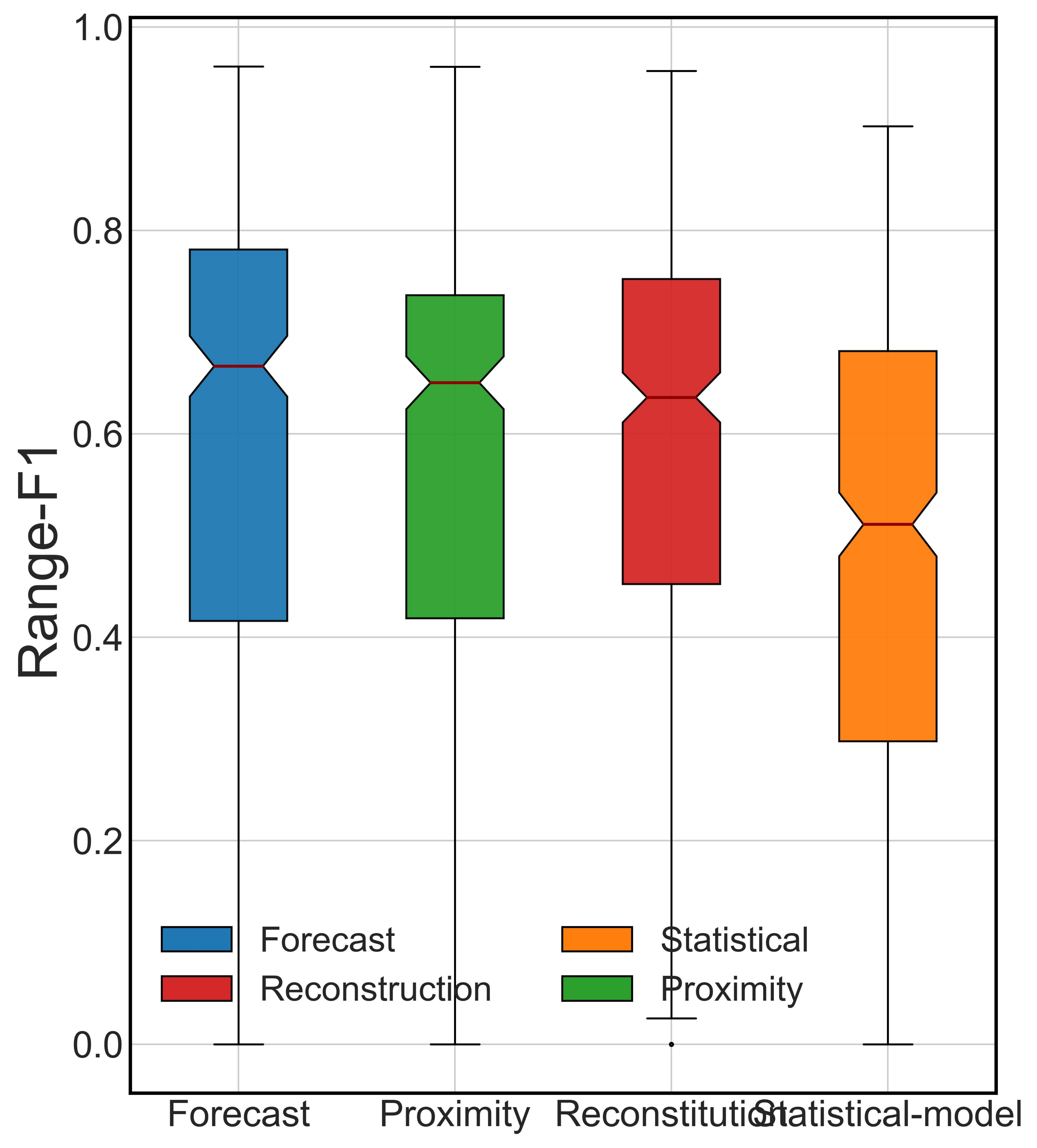}}

    \subfigure[AUC-ROC]{\includegraphics[width=0.18\textwidth]{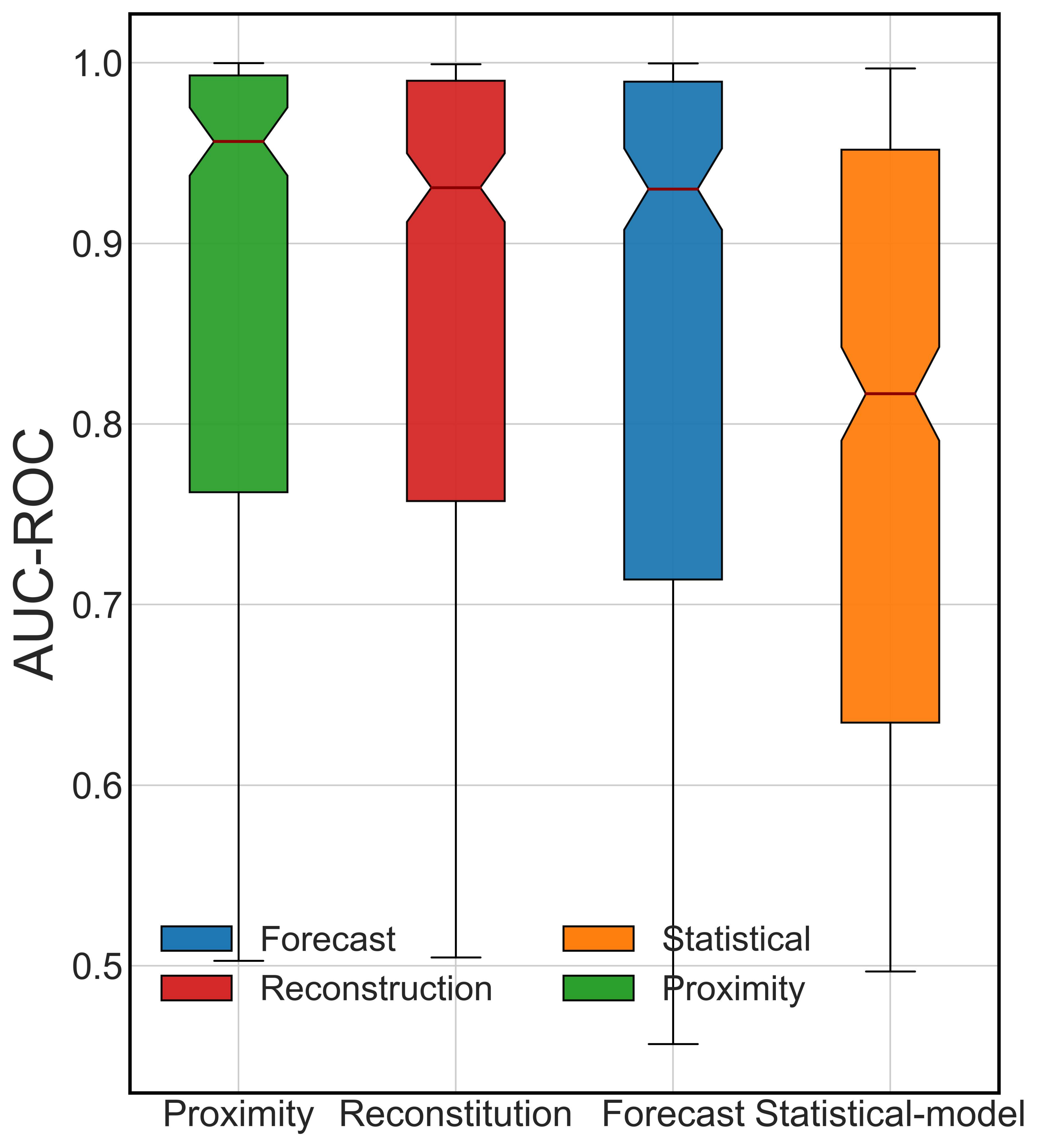}}
    \subfigure[AUC-PR]
    {\includegraphics[width=0.18\textwidth]{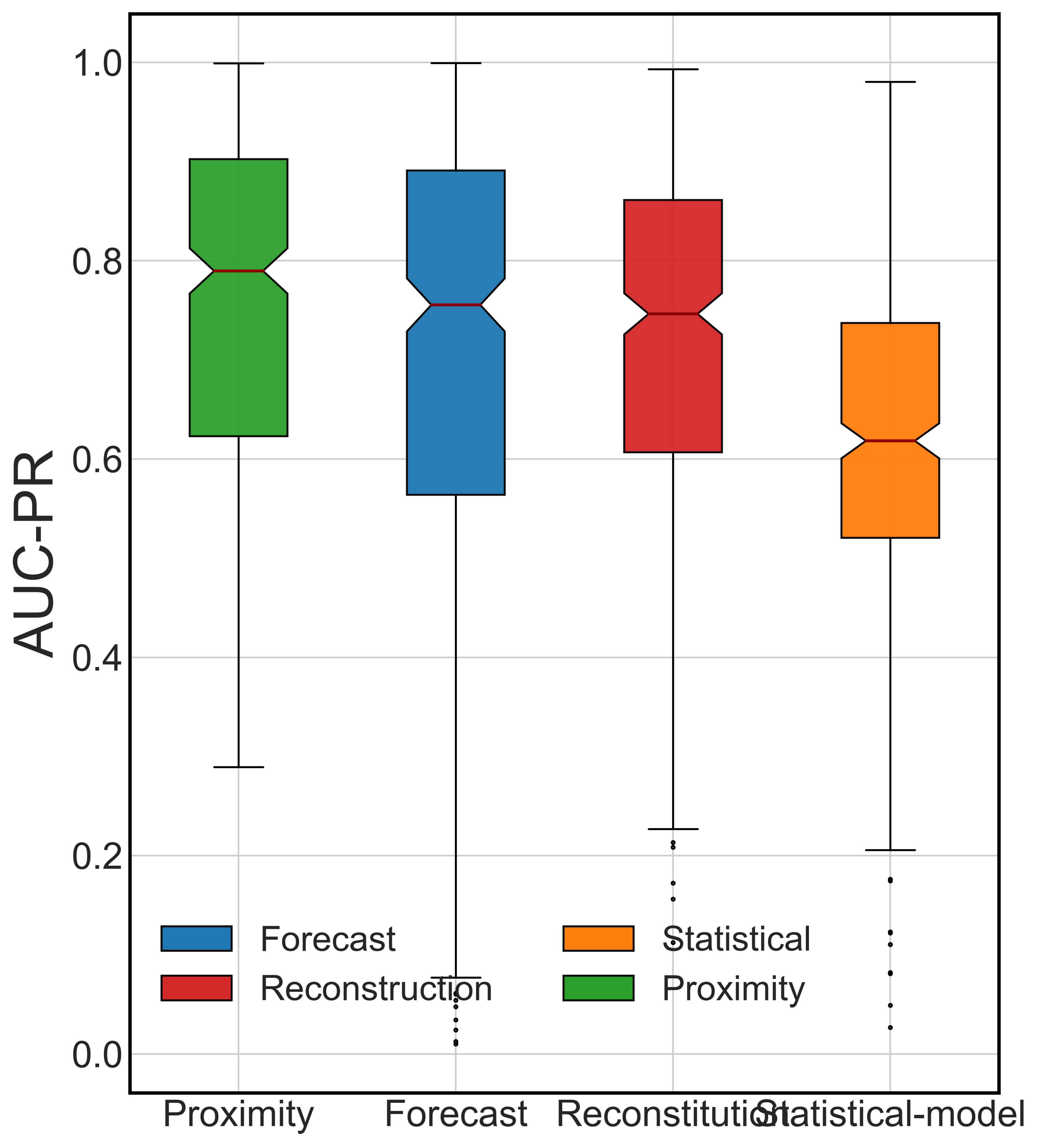}}
    \subfigure[VUS-ROC]{\includegraphics[width=0.18\textwidth]{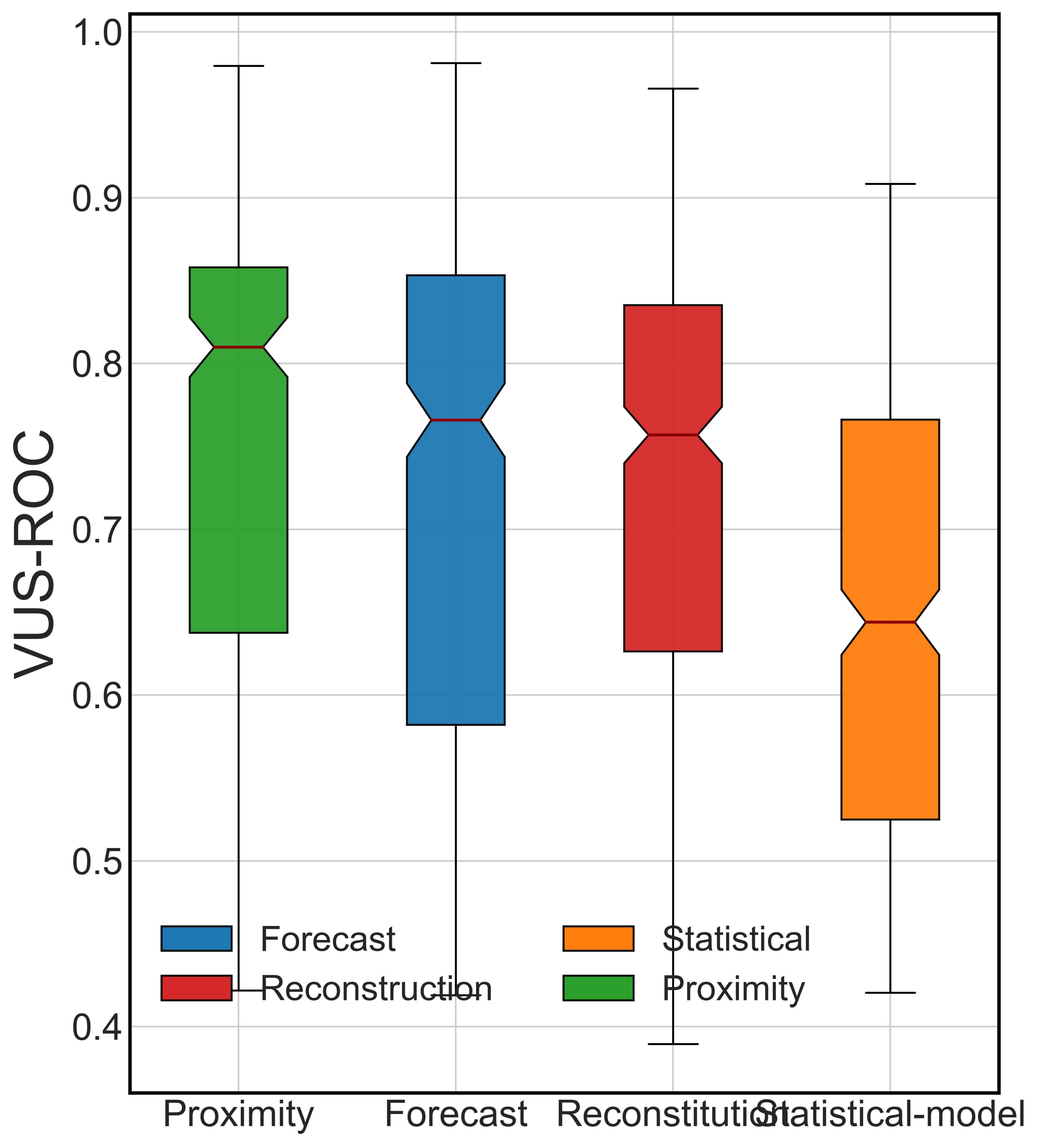}}
    \subfigure[VUS-PR]
    {\includegraphics[width=0.18\textwidth]{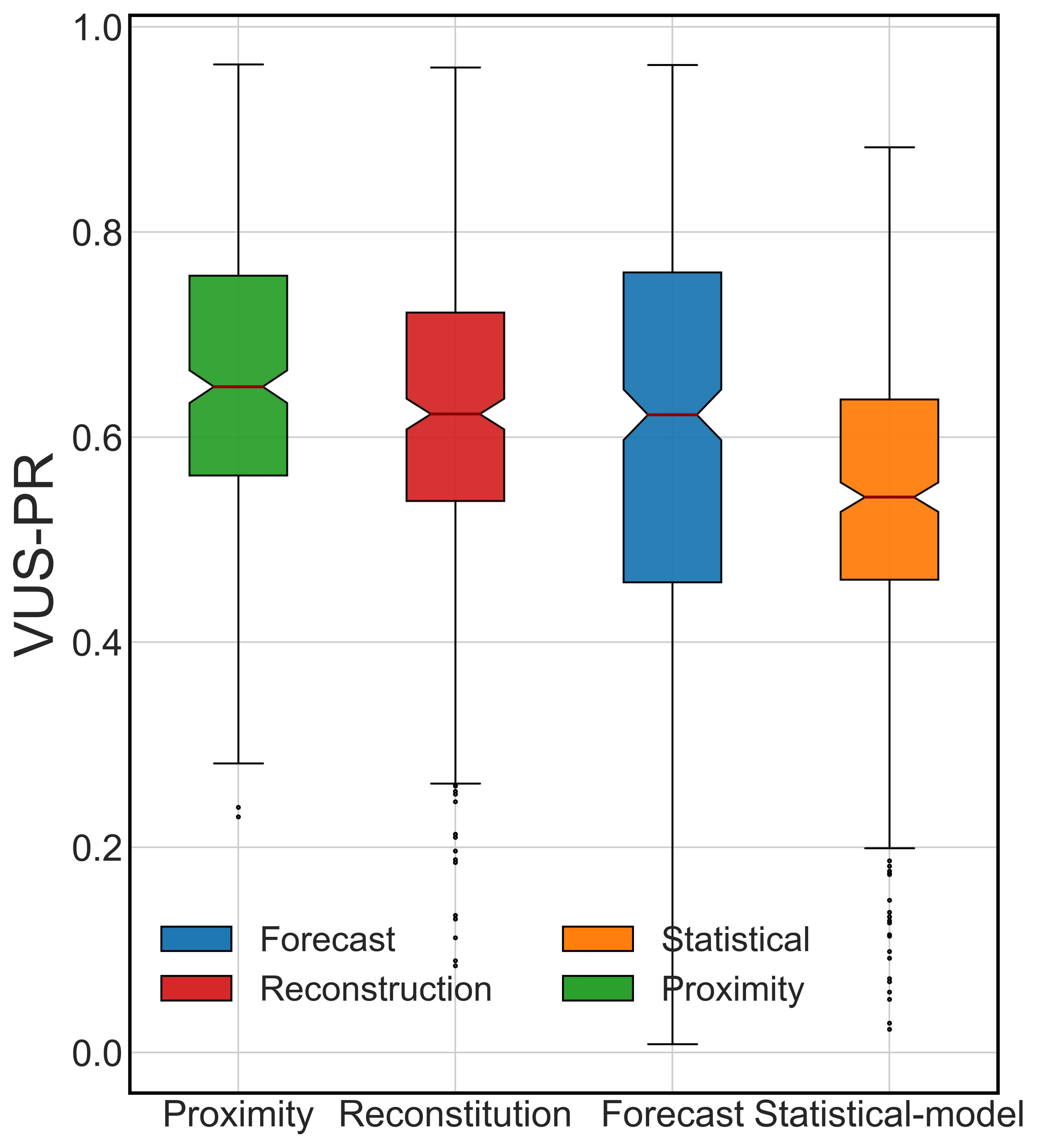}}

    \subfigure[Critical Difference under Precision]{\includegraphics[width=0.38\textwidth]{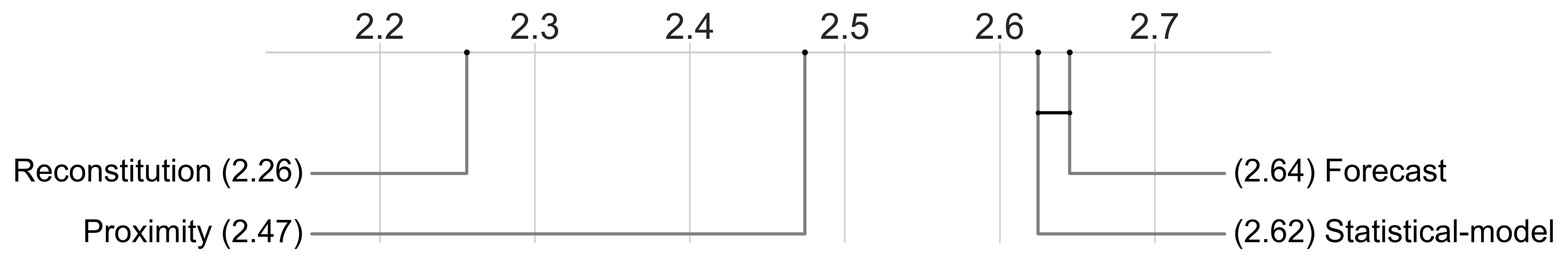}}
    \subfigure[Critical Difference under Recall]
    {\includegraphics[width=0.38\textwidth]{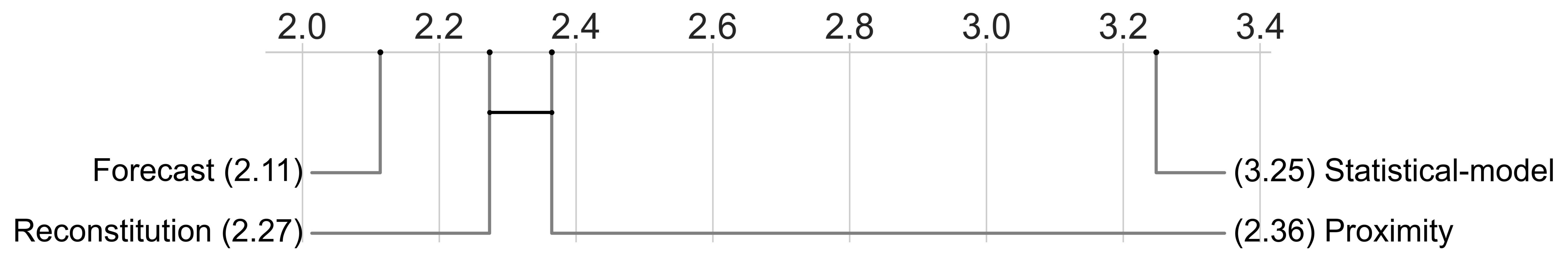}}
    
    \subfigure[Critical Difference under F1-score]{\includegraphics[width=0.38\textwidth]{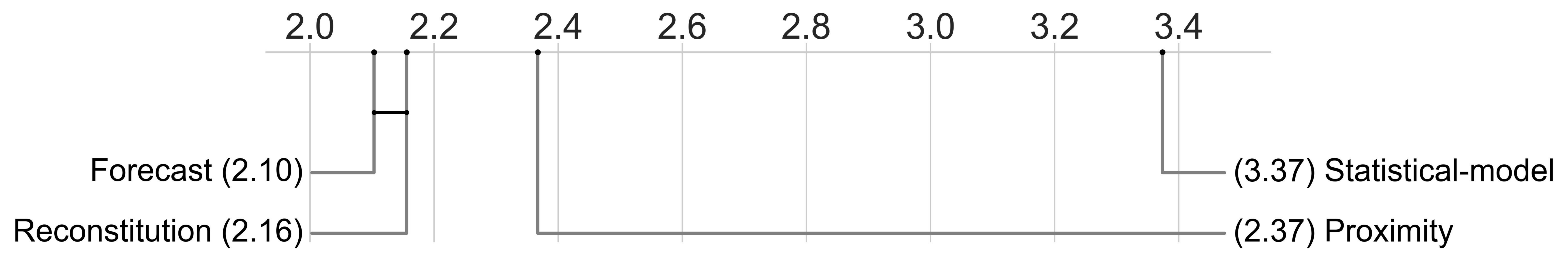}}
    \subfigure[Critical Difference under Range-f1]
    {\includegraphics[width=0.38\textwidth]{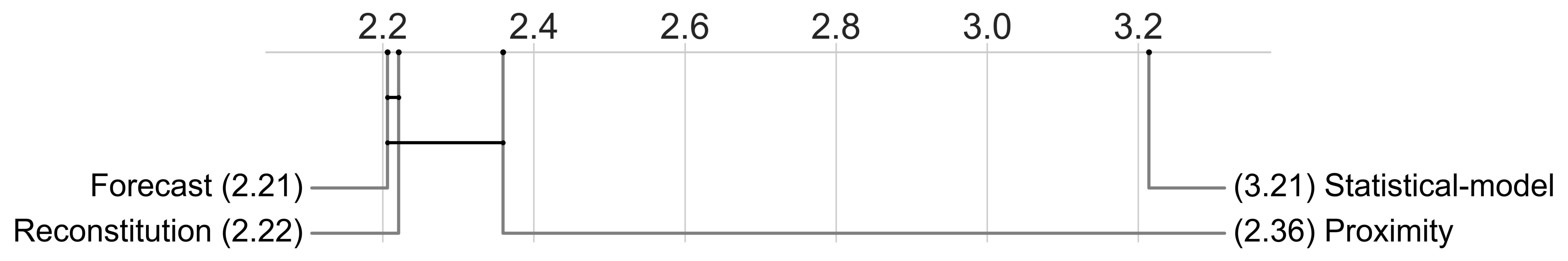}}

    \subfigure[Critical Difference under AUC-ROC]{\includegraphics[width=0.38\textwidth]{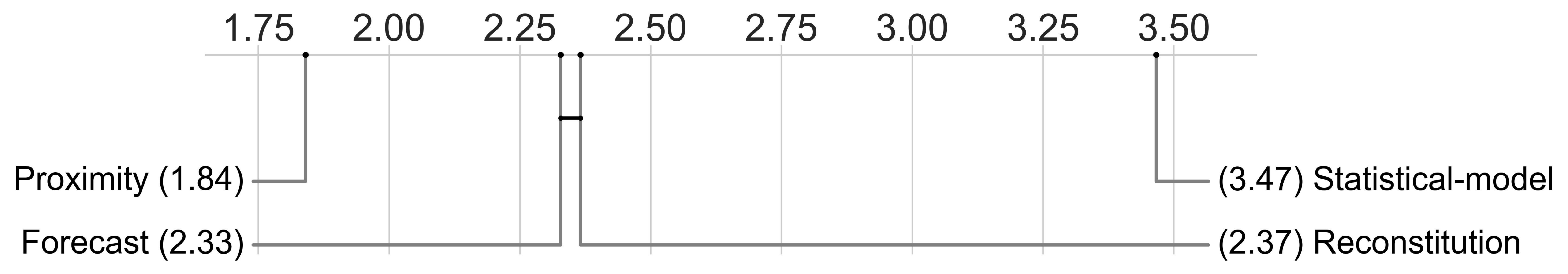}}
    \subfigure[Critical Difference under AUC-PR]
    {\includegraphics[width=0.38\textwidth]{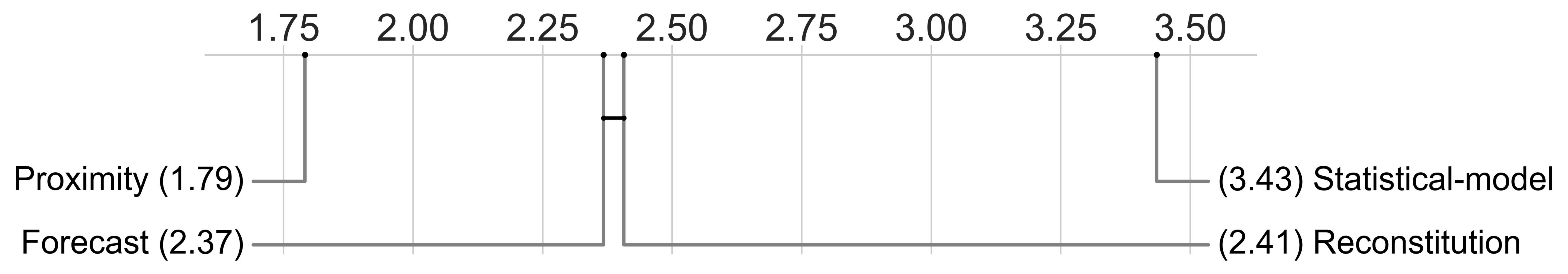}}
    
    \subfigure[Critical Difference under VUS-ROC]{\includegraphics[width=0.38\textwidth]{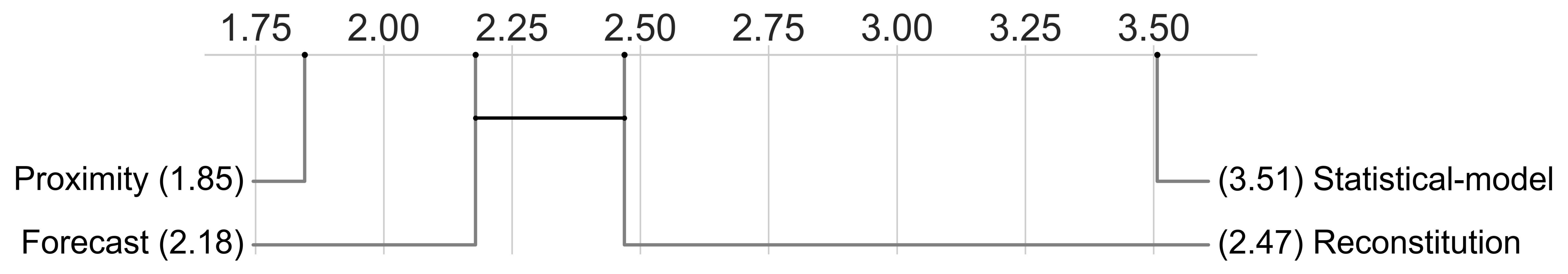}}
    \subfigure[Critical Difference under VUS-PR]
    {\includegraphics[width=0.38\textwidth]{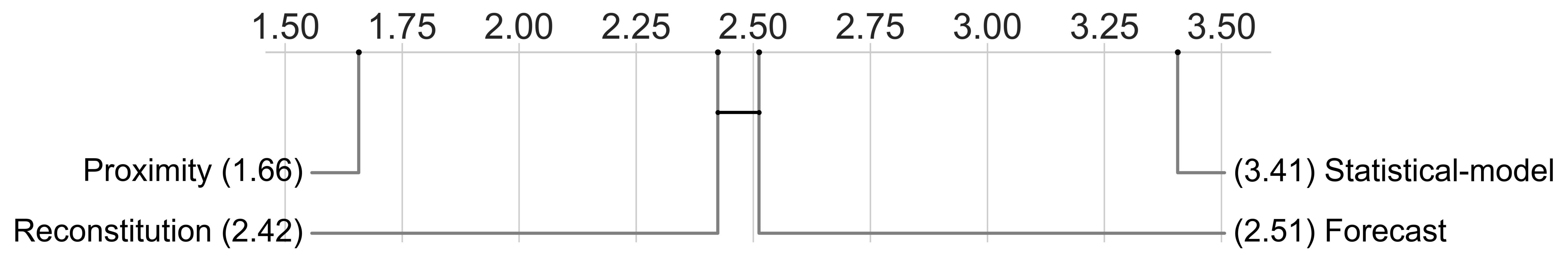}}

    \caption{Performance and Critical Difference of four types of algorithms under eight representative accuracy evaluation measures. 
    The sub-caption of (a-p) indicate the evaluation measures used.
    % \label{fig:isw2}
    }
\end{figure} 

\newpage
\subsection{More results for algorthms performance under KNC}
\label{sec:moreresultsknc}
%%%%%%%%%%%%%%%%%%%%%%%%%%%%%%%%
\begin{table}[htbp]
\renewcommand\arraystretch{1.1}
\setlength\tabcolsep{3.0pt}
\centering
\caption{Average value of eight representative accuracy evaluation measures for all the thirty-eight algorithms.}
\scalebox{0.95}{
\fontsize{6pt}{4.0pt}\selectfont
\begin{tabular}{|>{\centering\arraybackslash}m{2.5cm}||>{\centering\arraybackslash}m{1.5cm}|>{\centering\arraybackslash}m{1.5cm}|>{\centering\arraybackslash}m{1.5cm}|>{\centering\arraybackslash}m{1.5cm}|>{\centering\arraybackslash}m{1.5cm}|>{\centering\arraybackslash}m{1.5cm}|>{\centering\arraybackslash}m{1.5cm}|>{\centering\arraybackslash}m{1.5cm}|}
\toprule[0.8pt]
Metrics 1 & Precision & Recall & F1 & Range-F1 & AUC-ROC & AUC-PR & VUS-ROC & VUS-PR \\
\midrule
ABOD & 0.183 & 0.200 & 0.158 & 0.174 & 0.797 & 0.775 & 0.664 & 0.650  \\ 
AE & 0.743 & 0.729 & 0.613 & 0.655 & 0.907 & 0.802 & 0.785 & 0.680  \\ 
CBLOF & 0.799 & 0.631 & 0.562 & 0.656 & 0.879 & 0.773 & 0.771 & 0.664  \\ 
CD & 0.378 & 0.197 & 0.165 & 0.253 & 0.596 & 0.583 & 0.514 & 0.510  \\ 
CNN & 0.739 & 0.690 & 0.567 & 0.633 & 0.876 & 0.741 & 0.757 & 0.629  \\ 
COF & 0.815 & 0.591 & 0.543 & 0.650 & 0.858 & 0.724 & 0.765 & 0.624  \\ 
COPOD & 0.412 & 0.220 & 0.207 & 0.297 & 0.678 & 0.425 & 0.572 & 0.359  \\ 
DeepSVDD & 0.754 & 0.807 & 0.644 & 0.680 & 0.913 & 0.764 & 0.816 & 0.677  \\ 
ECOD & 0.452 & 0.247 & 0.232 & 0.341 & 0.686 & 0.428 & 0.580 & 0.370  \\ 
FCNN & 0.813 & 0.696 & 0.596 & 0.663 & 0.881 & 0.753 & 0.776 & 0.653  \\ 
GMM & 0.683 & 0.672 & 0.553 & 0.614 & 0.861 & 0.684 & 0.751 & 0.588  \\ 
GRU & 0.581 & 0.579 & 0.463 & 0.508 & 0.822 & 0.628 & 0.696 & 0.524  \\ 
HBOS & 0.721 & 0.488 & 0.448 & 0.549 & 0.810 & 0.686 & 0.692 & 0.583  \\ 
IForest & 0.730 & 0.520 & 0.474 & 0.568 & 0.838 & 0.707 & 0.710 & 0.595  \\ 
INNE & 0.708 & 0.581 & 0.505 & 0.577 & 0.844 & 0.740 & 0.732 & 0.646  \\ 
KDE & 0.812 & 0.776 & 0.641 & 0.690 & 0.921 & 0.811 & 0.792 & 0.688  \\ 
KMeans & 0.818 & 0.689 & 0.614 & 0.685 & 0.895 & 0.803 & 0.795 & 0.700  \\ 
KNN & 0.798 & 0.799 & 0.686 & 0.719 & 0.949 & 0.857 & 0.847 & 0.749  \\ 
KPCA & 0.602 & 0.724 & 0.560 & 0.574 & 0.883 & 0.705 & 0.765 & 0.606  \\ 
LinearRegression & 0.450 & 0.459 & 0.352 & 0.403 & 0.746 & 0.501 & 0.645 & 0.422  \\ 
LODA & 0.693 & 0.466 & 0.395 & 0.510 & 0.785 & 0.640 & 0.664 & 0.565  \\ 
LOF & 0.750 & \textbf{0.873} & 0.713 & 0.701 & 0.953 & 0.851 & 0.865 & 0.762  \\ 
LSTM & 0.578 & 0.621 & 0.485 & 0.527 & 0.827 & 0.649 & 0.713 & 0.547  \\ 
MAD & 0.582 & 0.357 & 0.320 & 0.406 & 0.752 & 0.581 & 0.616 & 0.502  \\ 
MCD & 0.511 & 0.472 & 0.366 & 0.435 & 0.751 & 0.537 & 0.631 & 0.441  \\ 
MSD & 0.493 & 0.290 & 0.259 & 0.341 & 0.704 & 0.510 & 0.583 & 0.450  \\ 
OCSVM & 0.466 & 0.153 & 0.161 & 0.311 & 0.614 & 0.387 & 0.533 & 0.355  \\ 
PCA & 0.275 & 0.066 & 0.085 & 0.177 & 0.566 & 0.332 & 0.499 & 0.314  \\ 
QMCD & 0.053 & 0.443 & 0.074 & 0.074 & 0.654 & 0.663 & 0.598 & 0.595  \\ 
RNN & 0.600 & 0.636 & 0.507 & 0.554 & 0.841 & 0.667 & 0.729 & 0.571  \\ 
Sampling & \textbf{0.841} & 0.818 & \textbf{0.718} & \textbf{0.762} & \textbf{0.956} & \textbf{0.873} & 0.858 & \textbf{0.769}  \\ 
SOD & 0.771 & 0.678 & 0.567 & 0.630 & 0.875 & 0.714 & 0.752 & 0.607  \\ 
SOS & 0.510 & 0.825 & 0.535 & 0.476 & 0.955 & 0.851 & \textbf{0.883} & 0.740  \\ 
SVDD & 0.642 & 0.421 & 0.362 & 0.491 & 0.757 & 0.567 & 0.640 & 0.513  \\ 
TadGan & 0.804 & 0.582 & 0.523 & 0.622 & 0.843 & 0.736 & 0.733 & 0.639  \\ 
TanoGan & 0.446 & 0.176 & 0.193 & 0.313 & 0.651 & 0.529 & 0.541 & 0.481  \\ 
Transformer & 0.675 & 0.510 & 0.441 & 0.519 & 0.806 & 0.620 & 0.685 & 0.531  \\ 
VAE & 0.691 & 0.559 & 0.481 & 0.541 & 0.831 & 0.692 & 0.698 & 0.576  \\ 
\bottomrule[1pt]
\end{tabular}
}
\end{table}

%%%%%%%%%%%%%%%%%%%%%%%%%%%%%%%%
\begin{table}[htbp]
\renewcommand\arraystretch{1.1}
\setlength\tabcolsep{3.0pt}
\centering
\caption{Average value of eight representative accuracy evaluation measures for all the thirty-eight algorithms when KNC$>$10}
\scalebox{0.95}{
\fontsize{6pt}{4.0pt}\selectfont
\begin{tabular}{|>{\centering\arraybackslash}m{2.5cm}||>{\centering\arraybackslash}m{1.5cm}|>{\centering\arraybackslash}m{1.5cm}|>{\centering\arraybackslash}m{1.5cm}|>{\centering\arraybackslash}m{1.5cm}|>{\centering\arraybackslash}m{1.5cm}|>{\centering\arraybackslash}m{1.5cm}|>{\centering\arraybackslash}m{1.5cm}|>{\centering\arraybackslash}m{1.5cm}|}
\toprule[0.8pt]
Metrics 1 & Precision & Recall & F1 & Range-F1 & AUC-ROC & AUC-PR & VUS-ROC & VUS-PR \\
\midrule

ABOD & 0.108 & 0.131 & 0.111 & 0.113 & 0.806 & 0.816 & 0.686 & 0.676  \\ 
AE & 0.753 & 0.952 & 0.772 & 0.738 & 0.975 & 0.914 & 0.878 & 0.806  \\ 
CBLOF & 0.841 & 0.801 & 0.708 & 0.755 & 0.913 & 0.854 & 0.840 & 0.755  \\ 
CD & 0.555 & 0.360 & 0.350 & 0.447 & 0.715 & 0.668 & 0.591 & 0.559  \\ 
CNN & 0.835 & 0.921 & 0.763 & 0.774 & 0.964 & 0.887 & 0.878 & 0.789  \\ 
COF & 0.886 & 0.765 & 0.704 & 0.771 & 0.932 & 0.851 & 0.853 & 0.747  \\ 
COPOD & 0.570 & 0.362 & 0.345 & 0.451 & 0.779 & 0.513 & 0.654 & 0.436  \\ 
DeepSVDD & 0.758 & 0.961 & 0.752 & 0.722 & 0.974 & 0.857 & 0.897 & 0.782  \\ 
ECOD & 0.584 & 0.377 & 0.359 & 0.480 & 0.765 & 0.489 & 0.645 & 0.426  \\ 
FCNN & 0.899 & 0.893 & 0.752 & 0.782 & 0.961 & 0.882 & 0.880 & 0.793  \\ 
GMM & 0.748 & 0.867 & 0.709 & 0.730 & 0.953 & 0.804 & 0.856 & 0.709  \\ 
GRU & 0.706 & 0.846 & 0.674 & 0.670 & 0.931 & 0.779 & 0.821 & 0.675  \\ 
HBOS & 0.860 & 0.645 & 0.609 & 0.716 & 0.890 & 0.807 & 0.776 & 0.688  \\ 
IForest & 0.869 & 0.750 & 0.699 & 0.759 & 0.931 & 0.872 & 0.818 & 0.727  \\ 
INNE & 0.818 & 0.821 & 0.705 & 0.725 & 0.918 & 0.871 & 0.835 & 0.768  \\ 
KDE & 0.923 & 0.960 & 0.796 & 0.817 & 0.973 & 0.919 & 0.870 & 0.805  \\ 
KMeans & 0.815 & 0.822 & 0.730 & 0.737 & 0.925 & 0.883 & 0.857 & 0.799  \\ 
KNN & 0.782 & 0.991 & 0.824 & 0.768 & 0.993 & 0.944 & 0.927 & 0.860  \\ 
KPCA & 0.661 & 0.950 & 0.702 & 0.642 & 0.965 & 0.836 & 0.874 & 0.752  \\ 
LinearRegression & 0.528 & 0.698 & 0.499 & 0.500 & 0.861 & 0.626 & 0.751 & 0.538  \\ 
LODA & 0.821 & 0.753 & 0.651 & 0.703 & 0.907 & 0.830 & 0.792 & 0.712  \\ 
LOF & 0.728 & 0.998 & 0.792 & 0.701 & 0.991 & 0.924 & 0.931 & 0.850  \\ 
LSTM & 0.695 & 0.879 & 0.692 & 0.683 & 0.941 & 0.815 & 0.844 & 0.718  \\ 
MAD & 0.768 & 0.534 & 0.481 & 0.570 & 0.836 & 0.661 & 0.699 & 0.581  \\ 
MCD & 0.612 & 0.667 & 0.495 & 0.557 & 0.839 & 0.665 & 0.717 & 0.557  \\ 
MSD & 0.637 & 0.437 & 0.386 & 0.460 & 0.777 & 0.561 & 0.649 & 0.512  \\ 
OCSVM & 0.667 & 0.292 & 0.295 & 0.487 & 0.684 & 0.442 & 0.580 & 0.406  \\ 
PCA & 0.451 & 0.120 & 0.159 & 0.292 & 0.602 & 0.318 & 0.521 & 0.316  \\ 
QMCD & 0.081 & 0.643 & 0.110 & 0.105 & 0.752 & 0.754 & 0.673 & 0.648  \\ 
RNN & 0.731 & 0.893 & 0.716 & 0.718 & 0.944 & 0.831 & 0.853 & 0.739  \\ 
Sampling & 0.807 & 0.991 & 0.839 & 0.793 & 0.995 & 0.948 & 0.931 & 0.867  \\ 
SOD & 0.844 & 0.815 & 0.724 & 0.762 & 0.952 & 0.845 & 0.851 & 0.740  \\ 
SOS & 0.557 & 0.888 & 0.624 & 0.493 & 0.990 & 0.963 & 0.933 & 0.848  \\ 
SVDD & 0.743 & 0.678 & 0.565 & 0.648 & 0.857 & 0.715 & 0.734 & 0.625  \\ 
TadGan & 0.829 & 0.764 & 0.662 & 0.717 & 0.912 & 0.845 & 0.817 & 0.743  \\ 
TanoGan & 0.687 & 0.387 & 0.384 & 0.527 & 0.752 & 0.627 & 0.618 & 0.563  \\ 
Transformer & 0.749 & 0.745 & 0.621 & 0.657 & 0.898 & 0.743 & 0.793 & 0.655  \\ 
VAE & 0.851 & 0.803 & 0.697 & 0.708 & 0.923 & 0.853 & 0.814 & 0.727  \\ 
\bottomrule[1pt]
\end{tabular}
}
\end{table}

%%%%%%%%%%%%%%%%%%%%%%%%%%%%%%%%
\begin{table}[htbp]
\renewcommand\arraystretch{1.1}
\setlength\tabcolsep{3.0pt}
\centering
\caption{Average value of eight representative accuracy evaluation measures for all the thirty-eight algorithms when 5$<$KNC$<$10}
\scalebox{0.95}{
\fontsize{6pt}{4.0pt}\selectfont
\begin{tabular}{|>{\centering\arraybackslash}m{2.5cm}||>{\centering\arraybackslash}m{1.5cm}|>{\centering\arraybackslash}m{1.5cm}|>{\centering\arraybackslash}m{1.5cm}|>{\centering\arraybackslash}m{1.5cm}|>{\centering\arraybackslash}m{1.5cm}|>{\centering\arraybackslash}m{1.5cm}|>{\centering\arraybackslash}m{1.5cm}|>{\centering\arraybackslash}m{1.5cm}|}
\toprule[0.8pt]
Metrics 1 & Precision & Recall & F1 & Range-F1 & AUC-ROC & AUC-PR & VUS-ROC & VUS-PR \\
\midrule

ABOD & 0.261 & 0.324 & 0.254 & 0.273 & 0.885 & 0.862 & 0.718 & 0.685  \\ 
AE & 0.788 & 0.920 & 0.748 & 0.759 & 0.966 & 0.905 & 0.841 & 0.748  \\ 
CBLOF & 0.856 & 0.779 & 0.683 & 0.750 & 0.949 & 0.868 & 0.828 & 0.726  \\ 
CD & 0.460 & 0.179 & 0.163 & 0.268 & 0.614 & 0.591 & 0.491 & 0.491  \\ 
CNN & 0.834 & 0.910 & 0.718 & 0.750 & 0.955 & 0.862 & 0.845 & 0.733  \\ 
COF & 0.828 & 0.756 & 0.672 & 0.743 & 0.926 & 0.793 & 0.831 & 0.667  \\ 
COPOD & 0.418 & 0.205 & 0.204 & 0.285 & 0.683 & 0.418 & 0.568 & 0.337  \\ 
DeepSVDD & 0.724 & 0.941 & 0.709 & 0.726 & 0.965 & 0.828 & 0.853 & 0.723  \\ 
ECOD & 0.433 & 0.226 & 0.212 & 0.326 & 0.691 & 0.400 & 0.570 & 0.329  \\ 
FCNN & 0.878 & 0.905 & 0.738 & 0.764 & 0.956 & 0.845 & 0.851 & 0.740  \\ 
GMM & 0.720 & 0.864 & 0.674 & 0.706 & 0.939 & 0.807 & 0.823 & 0.678  \\ 
GRU & 0.671 & 0.772 & 0.585 & 0.617 & 0.912 & 0.729 & 0.751 & 0.577  \\ 
HBOS & 0.866 & 0.640 & 0.588 & 0.673 & 0.898 & 0.760 & 0.759 & 0.636  \\ 
IForest & 0.856 & 0.709 & 0.650 & 0.728 & 0.942 & 0.834 & 0.793 & 0.684  \\ 
INNE & 0.856 & 0.743 & 0.652 & 0.723 & 0.933 & 0.817 & 0.798 & 0.691  \\ 
KDE & 0.875 & 0.945 & 0.757 & 0.783 & 0.967 & 0.895 & 0.841 & 0.752  \\ 
KMeans & 0.887 & 0.887 & 0.773 & 0.804 & 0.959 & 0.900 & 0.861 & 0.766  \\ 
KNN & 0.805 & 0.959 & 0.807 & 0.786 & 0.985 & 0.934 & 0.896 & 0.821  \\ 
KPCA & 0.640 & 0.919 & 0.683 & 0.652 & 0.953 & 0.828 & 0.835 & 0.708  \\ 
LinearRegression & 0.481 & 0.681 & 0.480 & 0.490 & 0.838 & 0.585 & 0.707 & 0.477  \\ 
LODA & 0.775 & 0.616 & 0.518 & 0.612 & 0.860 & 0.715 & 0.708 & 0.608  \\ 
LOF & 0.717 & 0.982 & 0.763 & 0.704 & 0.984 & 0.912 & 0.907 & 0.829  \\ 
LSTM & 0.693 & 0.862 & 0.636 & 0.661 & 0.931 & 0.788 & 0.792 & 0.647  \\ 
MAD & 0.645 & 0.420 & 0.384 & 0.472 & 0.797 & 0.578 & 0.630 & 0.490  \\ 
MCD & 0.610 & 0.651 & 0.474 & 0.500 & 0.846 & 0.628 & 0.679 & 0.490  \\ 
MSD & 0.547 & 0.318 & 0.299 & 0.404 & 0.725 & 0.499 & 0.583 & 0.426  \\ 
OCSVM & 0.500 & 0.163 & 0.169 & 0.323 & 0.637 & 0.395 & 0.535 & 0.352  \\ 
PCA & 0.276 & 0.056 & 0.075 & 0.174 & 0.574 & 0.347 & 0.498 & 0.315  \\ 
QMCD & 0.051 & 0.537 & 0.070 & 0.068 & 0.696 & 0.695 & 0.615 & 0.598  \\ 
RNN & 0.731 & 0.860 & 0.655 & 0.675 & 0.934 & 0.800 & 0.801 & 0.669  \\ 
Sampling & 0.843 & 0.962 & 0.833 & 0.826 & 0.987 & 0.947 & 0.902 & 0.842  \\ 
SOD & 0.816 & 0.806 & 0.653 & 0.703 & 0.936 & 0.800 & 0.794 & 0.661  \\ 
SOS & 0.533 & 0.918 & 0.593 & 0.516 & 0.986 & 0.934 & 0.913 & 0.802  \\ 
SVDD & 0.711 & 0.543 & 0.445 & 0.554 & 0.813 & 0.597 & 0.661 & 0.531  \\ 
TadGan & 0.881 & 0.839 & 0.711 & 0.760 & 0.942 & 0.860 & 0.807 & 0.716  \\ 
TanoGan & 0.635 & 0.317 & 0.332 & 0.453 & 0.745 & 0.625 & 0.597 & 0.538  \\ 
Transformer & 0.694 & 0.553 & 0.483 & 0.536 & 0.838 & 0.634 & 0.694 & 0.520  \\ 
VAE & 0.804 & 0.740 & 0.597 & 0.633 & 0.924 & 0.770 & 0.729 & 0.595  \\  
\bottomrule[1pt]
\end{tabular}
}
\end{table}

%%%%%%%%%%%%%%%%%%%%%%%%%%%%%%%%
\begin{table}[htbp]
\renewcommand\arraystretch{1.1}
\setlength\tabcolsep{3.0pt}
\centering
\caption{Average value of eight representative accuracy evaluation measures for all the thirty-eight algorithms when 2$<$KNC$<$5}
\scalebox{0.95}{
\fontsize{6pt}{4.0pt}\selectfont
\begin{tabular}{|>{\centering\arraybackslash}m{2.5cm}||>{\centering\arraybackslash}m{1.5cm}|>{\centering\arraybackslash}m{1.5cm}|>{\centering\arraybackslash}m{1.5cm}|>{\centering\arraybackslash}m{1.5cm}|>{\centering\arraybackslash}m{1.5cm}|>{\centering\arraybackslash}m{1.5cm}|>{\centering\arraybackslash}m{1.5cm}|>{\centering\arraybackslash}m{1.5cm}|}
\toprule[0.8pt]
Metrics 1 & Precision & Recall & F1 & Range-F1 & AUC-ROC & AUC-PR & VUS-ROC & VUS-PR \\
\midrule

ABOD & 0.189 & 0.190 & 0.156 & 0.175 & 0.784 & 0.759 & 0.663 & 0.656  \\ 
AE & 0.779 & 0.605 & 0.546 & 0.643 & 0.888 & 0.745 & 0.759 & 0.630  \\ 
CBLOF & 0.807 & 0.502 & 0.479 & 0.620 & 0.861 & 0.740 & 0.747 & 0.639  \\ 
CD & 0.277 & 0.082 & 0.075 & 0.163 & 0.538 & 0.535 & 0.488 & 0.487  \\ 
CNN & 0.842 & 0.684 & 0.610 & 0.692 & 0.889 & 0.738 & 0.764 & 0.628  \\ 
COF & 0.779 & 0.469 & 0.453 & 0.582 & 0.808 & 0.657 & 0.715 & 0.568  \\ 
COPOD & 0.407 & 0.182 & 0.172 & 0.281 & 0.648 & 0.426 & 0.554 & 0.372  \\ 
DeepSVDD & 0.799 & 0.699 & 0.592 & 0.662 & 0.872 & 0.714 & 0.772 & 0.631  \\ 
ECOD & 0.451 & 0.216 & 0.205 & 0.311 & 0.661 & 0.442 & 0.567 & 0.392  \\ 
FCNN & 0.864 & 0.681 & 0.614 & 0.695 & 0.882 & 0.730 & 0.772 & 0.638  \\ 
GMM & 0.673 & 0.527 & 0.457 & 0.556 & 0.802 & 0.602 & 0.694 & 0.518  \\ 
GRU & 0.653 & 0.467 & 0.414 & 0.511 & 0.783 & 0.587 & 0.659 & 0.490  \\ 
HBOS & 0.665 & 0.336 & 0.318 & 0.446 & 0.736 & 0.603 & 0.629 & 0.523  \\ 
IForest & 0.624 & 0.351 & 0.311 & 0.425 & 0.761 & 0.602 & 0.643 & 0.521  \\ 
INNE & 0.619 & 0.399 & 0.368 & 0.469 & 0.784 & 0.660 & 0.673 & 0.594  \\ 
KDE & 0.836 & 0.695 & 0.604 & 0.679 & 0.918 & 0.773 & 0.783 & 0.661  \\ 
KMeans & 0.861 & 0.574 & 0.546 & 0.675 & 0.888 & 0.770 & 0.778 & 0.667  \\ 
KNN & 0.848 & 0.722 & 0.650 & 0.730 & 0.945 & 0.843 & 0.827 & 0.722  \\ 
KPCA & 0.617 & 0.579 & 0.486 & 0.559 & 0.840 & 0.624 & 0.716 & 0.527  \\ 
LinearRegression & 0.402 & 0.258 & 0.232 & 0.326 & 0.653 & 0.413 & 0.571 & 0.356  \\ 
LODA & 0.616 & 0.261 & 0.227 & 0.377 & 0.703 & 0.519 & 0.597 & 0.486  \\ 
LOF & 0.811 & 0.843 & 0.722 & 0.751 & 0.959 & 0.843 & 0.859 & 0.748  \\ 
LSTM & 0.638 & 0.498 & 0.431 & 0.516 & 0.788 & 0.593 & 0.661 & 0.491  \\ 
MAD & 0.509 & 0.233 & 0.225 & 0.324 & 0.701 & 0.560 & 0.579 & 0.491  \\ 
MCD & 0.403 & 0.297 & 0.252 & 0.340 & 0.666 & 0.446 & 0.569 & 0.375  \\ 
MSD & 0.447 & 0.199 & 0.189 & 0.281 & 0.667 & 0.503 & 0.560 & 0.449  \\ 
OCSVM & 0.366 & 0.080 & 0.094 & 0.230 & 0.577 & 0.363 & 0.515 & 0.351  \\ 
PCA & 0.188 & 0.035 & 0.047 & 0.119 & 0.546 & 0.344 & 0.495 & 0.335  \\ 
QMCD & 0.043 & 0.280 & 0.056 & 0.060 & 0.619 & 0.626 & 0.579 & 0.584  \\ 
RNN & 0.652 & 0.499 & 0.432 & 0.530 & 0.791 & 0.600 & 0.670 & 0.503  \\ 
Sampling & 0.883 & 0.739 & 0.680 & 0.766 & 0.950 & 0.855 & 0.836 & 0.737  \\ 
SOD & 0.781 & 0.641 & 0.523 & 0.598 & 0.841 & 0.655 & 0.722 & 0.546  \\ 
SOS & 0.536 & 0.771 & 0.516 & 0.488 & 0.941 & 0.801 & 0.869 & 0.707  \\ 
SVDD & 0.611 & 0.267 & 0.251 & 0.426 & 0.704 & 0.513 & 0.606 & 0.479  \\ 
TadGan & 0.835 & 0.481 & 0.453 & 0.593 & 0.808 & 0.674 & 0.697 & 0.594  \\ 
TanoGan & 0.717 & 0.258 & 0.289 & 0.454 & 0.707 & 0.620 & 0.584 & 0.553  \\ 
Transformer & 0.670 & 0.391 & 0.350 & 0.478 & 0.761 & 0.573 & 0.642 & 0.502  \\ 
VAE & 0.619 & 0.375 & 0.348 & 0.451 & 0.763 & 0.596 & 0.648 & 0.510  \\ 
\bottomrule[1pt]
\end{tabular}
}
\end{table}

%%%%%%%%%%%%%%%%%%%%%%%%%%%%%%%%
\begin{table}[htbp]
\renewcommand\arraystretch{1.1}
\setlength\tabcolsep{3.0pt}
\centering
\caption{Average value of eight representative accuracy evaluation measures for all the thirty-eight algorithms when 1$<$KNC$<$2}
\scalebox{0.95}{
\fontsize{6pt}{4.0pt}\selectfont
\begin{tabular}{|>{\centering\arraybackslash}m{2.5cm}||>{\centering\arraybackslash}m{1.5cm}|>{\centering\arraybackslash}m{1.5cm}|>{\centering\arraybackslash}m{1.5cm}|>{\centering\arraybackslash}m{1.5cm}|>{\centering\arraybackslash}m{1.5cm}|>{\centering\arraybackslash}m{1.5cm}|>{\centering\arraybackslash}m{1.5cm}|>{\centering\arraybackslash}m{1.5cm}|}
\toprule[0.8pt]
Metrics 1 & Precision & Recall & F1 & Range-F1 & AUC-ROC & AUC-PR & VUS-ROC & VUS-PR \\
\midrule

ABOD & 0.190 & 0.168 & 0.115 & 0.140 & 0.709 & 0.640 & 0.578 & 0.561  \\ 
AE & 0.612 & 0.419 & 0.353 & 0.440 & 0.783 & 0.626 & 0.639 & 0.511  \\ 
CBLOF & 0.658 & 0.457 & 0.367 & 0.477 & 0.784 & 0.610 & 0.654 & 0.513  \\ 
CD & 0.213 & 0.201 & 0.073 & 0.129 & 0.515 & 0.538 & 0.480 & 0.502  \\ 
CNN & 0.804 & 0.551 & 0.492 & 0.614 & 0.846 & 0.696 & 0.708 & 0.568  \\ 
COF & 0.757 & 0.377 & 0.332 & 0.488 & 0.769 & 0.581 & 0.659 & 0.494  \\ 
COPOD & 0.205 & 0.113 & 0.086 & 0.129 & 0.587 & 0.305 & 0.497 & 0.249  \\ 
DeepSVDD & 0.711 & 0.639 & 0.514 & 0.606 & 0.847 & 0.652 & 0.742 & 0.560  \\ 
ECOD & 0.283 & 0.153 & 0.135 & 0.217 & 0.618 & 0.343 & 0.522 & 0.291  \\ 
FCNN & 0.826 & 0.478 & 0.443 & 0.574 & 0.817 & 0.652 & 0.689 & 0.533  \\ 
GMM & 0.578 & 0.429 & 0.365 & 0.452 & 0.748 & 0.522 & 0.622 & 0.443  \\ 
GRU & 0.510 & 0.283 & 0.265 & 0.368 & 0.701 & 0.532 & 0.574 & 0.440  \\ 
HBOS & 0.460 & 0.366 & 0.293 & 0.361 & 0.732 & 0.572 & 0.609 & 0.475  \\ 
IForest & 0.571 & 0.279 & 0.240 & 0.361 & 0.721 & 0.506 & 0.576 & 0.426  \\ 
INNE & 0.544 & 0.380 & 0.296 & 0.393 & 0.745 & 0.612 & 0.613 & 0.516  \\ 
KDE & 0.551 & 0.472 & 0.358 & 0.431 & 0.805 & 0.631 & 0.644 & 0.500  \\ 
KMeans & 0.664 & 0.477 & 0.386 & 0.494 & 0.795 & 0.639 & 0.664 & 0.545  \\ 
KNN & 0.728 & 0.496 & 0.425 & 0.560 & 0.860 & 0.676 & 0.720 & 0.562  \\ 
KPCA & 0.458 & 0.443 & 0.350 & 0.423 & 0.767 & 0.521 & 0.617 & 0.418  \\ 
LinearRegression & 0.397 & 0.219 & 0.207 & 0.301 & 0.642 & 0.387 & 0.552 & 0.310  \\ 
LODA & 0.564 & 0.258 & 0.200 & 0.362 & 0.672 & 0.502 & 0.550 & 0.449  \\ 
LOF & 0.709 & 0.638 & 0.539 & 0.615 & 0.860 & 0.697 & 0.741 & 0.591  \\ 
LSTM & 0.508 & 0.293 & 0.267 & 0.369 & 0.698 & 0.547 & 0.577 & 0.441  \\ 
MAD & 0.370 & 0.254 & 0.183 & 0.239 & 0.670 & 0.502 & 0.545 & 0.420  \\ 
MCD & 0.450 & 0.296 & 0.261 & 0.357 & 0.668 & 0.412 & 0.564 & 0.340  \\ 
MSD & 0.313 & 0.216 & 0.156 & 0.204 & 0.639 & 0.461 & 0.529 & 0.388  \\ 
OCSVM & 0.323 & 0.080 & 0.088 & 0.203 & 0.559 & 0.337 & 0.496 & 0.286  \\ 
PCA & 0.195 & 0.062 & 0.064 & 0.128 & 0.545 & 0.304 & 0.478 & 0.264  \\ 
QMCD & 0.038 & 0.358 & 0.066 & 0.068 & 0.533 & 0.560 & 0.510 & 0.532  \\ 
RNN & 0.539 & 0.305 & 0.279 & 0.399 & 0.727 & 0.543 & 0.601 & 0.424  \\ 
Sampling & 0.815 & 0.559 & 0.491 & 0.644 & 0.888 & 0.722 & 0.749 & 0.610  \\ 
SOD & 0.597 & 0.392 & 0.321 & 0.412 & 0.758 & 0.531 & 0.617 & 0.462  \\ 
SOS & 0.365 & 0.721 & 0.375 & 0.379 & 0.903 & 0.688 & 0.810 & 0.571  \\ 
SVDD & 0.470 & 0.189 & 0.175 & 0.312 & 0.645 & 0.421 & 0.545 & 0.393  \\ 
TadGan & 0.624 & 0.203 & 0.231 & 0.380 & 0.691 & 0.544 & 0.592 & 0.474  \\ 
TanoGan & 0.646 & 0.231 & 0.226 & 0.387 & 0.697 & 0.587 & 0.572 & 0.522  \\ 
Transformer & 0.546 & 0.350 & 0.300 & 0.378 & 0.722 & 0.511 & 0.602 & 0.420  \\ 
VAE & 0.454 & 0.328 & 0.278 & 0.360 & 0.716 & 0.547 & 0.589 & 0.458  \\ 
\bottomrule[1pt]
\end{tabular}
}
\end{table}

%%%%%%%%%%%%%%%%%%%%%%%%%%%%%%%%
\begin{table}[htbp]
\renewcommand\arraystretch{1.1}
\setlength\tabcolsep{3.0pt}
\centering
\caption{Average value of eight representative accuracy evaluation measures for all the thirty-eight algorithms when KNC$<$1}
\scalebox{0.95}{
\fontsize{6pt}{4.0pt}\selectfont
\begin{tabular}{|>{\centering\arraybackslash}m{2.5cm}||>{\centering\arraybackslash}m{1.5cm}|>{\centering\arraybackslash}m{1.5cm}|>{\centering\arraybackslash}m{1.5cm}|>{\centering\arraybackslash}m{1.5cm}|>{\centering\arraybackslash}m{1.5cm}|>{\centering\arraybackslash}m{1.5cm}|>{\centering\arraybackslash}m{1.5cm}|>{\centering\arraybackslash}m{1.5cm}|}
\toprule[0.8pt]
Metrics 1 & Precision & Recall & F1 & Range-F1 & AUC-ROC & AUC-PR & VUS-ROC & VUS-PR \\
\midrule

ABOD & 0.000 & 0.000 & 0.000 & 0.000 & 0.500 & 0.694 & 0.493 & 0.688  \\ 
AE & 1.000 & 0.312 & 0.380 & 0.558 & 0.729 & 0.796 & 0.682 & 0.749  \\ 
CBLOF & 1.000 & 0.334 & 0.351 & 0.548 & 0.705 & 0.737 & 0.644 & 0.689  \\ 
CD & 0.344 & 0.020 & 0.038 & 0.180 & 0.516 & 0.694 & 0.493 & 0.678  \\ 
CNN & 1.000 & 0.383 & 0.482 & 0.679 & 0.744 & 0.817 & 0.673 & 0.754  \\ 
COF & 1.000 & 0.152 & 0.252 & 0.652 & 0.656 & 0.761 & 0.617 & 0.703  \\ 
COPOD & 0.224 & 0.009 & 0.017 & 0.171 & 0.520 & 0.610 & 0.483 & 0.576  \\ 
DeepSVDD & 1.000 & 0.413 & 0.502 & 0.581 & 0.736 & 0.799 & 0.666 & 0.747  \\ 
ECOD & 0.894 & 0.035 & 0.066 & 0.430 & 0.601 & 0.685 & 0.547 & 0.634  \\ 
FCNN & 1.000 & 0.479 & 0.513 & 0.696 & 0.755 & 0.814 & 0.673 & 0.744  \\ 
GMM & 0.462 & 0.500 & 0.376 & 0.431 & 0.675 & 0.517 & 0.665 & 0.498  \\ 
GRU & 0.700 & 0.325 & 0.353 & 0.493 & 0.677 & 0.591 & 0.632 & 0.544  \\ 
HBOS & 0.000 & 0.000 & 0.000 & 0.000 & 0.513 & 0.702 & 0.494 & 0.675  \\ 
IForest & 0.500 & 0.121 & 0.193 & 0.377 & 0.660 & 0.766 & 0.620 & 0.729  \\ 
INNE & 0.500 & 0.446 & 0.410 & 0.415 & 0.729 & 0.813 & 0.674 & 0.766  \\ 
KDE & 0.500 & 0.456 & 0.400 & 0.398 & 0.758 & 0.823 & 0.674 & 0.755  \\ 
KMeans & 1.000 & 0.462 & 0.459 & 0.595 & 0.723 & 0.737 & 0.675 & 0.706  \\ 
KNN & 1.000 & 0.169 & 0.261 & 0.592 & 0.732 & 0.758 & 0.688 & 0.722  \\ 
KPCA & 0.429 & 0.469 & 0.429 & 0.414 & 0.754 & 0.826 & 0.703 & 0.767  \\ 
LinearRegression & 0.440 & 0.325 & 0.340 & 0.422 & 0.630 & 0.473 & 0.618 & 0.460  \\ 
LODA & 0.500 & 0.033 & 0.062 & 0.216 & 0.668 & 0.789 & 0.617 & 0.745  \\ 
LOF & 1.000 & 0.579 & 0.518 & 0.708 & 0.812 & 0.842 & 0.742 & 0.787  \\ 
LSTM & 0.896 & 0.521 & 0.428 & 0.542 & 0.704 & 0.695 & 0.640 & 0.653  \\ 
MAD & 0.500 & 0.275 & 0.355 & 0.421 & 0.700 & 0.814 & 0.643 & 0.763  \\ 
MCD & 0.438 & 0.413 & 0.354 & 0.408 & 0.655 & 0.481 & 0.644 & 0.467  \\ 
MSD & 0.500 & 0.294 & 0.370 & 0.428 & 0.701 & 0.814 & 0.643 & 0.763  \\ 
OCSVM & 0.500 & 0.011 & 0.022 & 0.205 & 0.535 & 0.694 & 0.510 & 0.653  \\ 
PCA & 0.000 & 0.000 & 0.000 & 0.000 & 0.524 & 0.694 & 0.497 & 0.678  \\ 
QMCD & 0.000 & 0.000 & 0.000 & 0.000 & 0.539 & 0.694 & 0.527 & 0.678  \\ 
RNN & 0.945 & 0.511 & 0.408 & 0.597 & 0.691 & 0.611 & 0.651 & 0.594  \\ 
Sampling & 1.000 & 0.316 & 0.406 & 0.671 & 0.746 & 0.776 & 0.695 & 0.730  \\ 
SOD & 1.000 & 0.989 & 0.695 & 0.798 & 0.833 & 0.772 & 0.770 & 0.731  \\ 
SOS & 0.813 & 1.000 & 0.655 & 0.698 & 0.790 & 0.761 & 0.740 & 0.711  \\ 
SVDD & 1.000 & 0.299 & 0.393 & 0.575 & 0.698 & 0.799 & 0.646 & 0.747  \\ 
TadGan & 1.000 & 0.332 & 0.367 & 0.552 & 0.686 & 0.704 & 0.633 & 0.678  \\ 
TanoGan & 0.560 & 0.036 & 0.066 & 0.338 & 0.544 & 0.718 & 0.524 & 0.688  \\ 
Transformer & 1.000 & 0.231 & 0.340 & 0.505 & 0.689 & 0.801 & 0.613 & 0.738  \\ 
VAE & 1.000 & 0.477 & 0.383 & 0.538 & 0.686 & 0.764 & 0.616 & 0.723  \\ 
\bottomrule[1pt]
\end{tabular}
}
\end{table}

%%%%%%%%%%%%%%%%%%%%%%%%%%%%%%%%
\begin{table}[htbp]
\renewcommand\arraystretch{1.1}
\setlength\tabcolsep{3.0pt}
\centering
\caption{Average value of eight representative accuracy evaluation measures for four types of algorithms.}
\scalebox{0.95}{
\fontsize{6pt}{4.0pt}\selectfont
\begin{tabular}{|>{\centering\arraybackslash}m{2.5cm}||>{\centering\arraybackslash}m{1.5cm}|>{\centering\arraybackslash}m{1.5cm}|>{\centering\arraybackslash}m{1.5cm}|>{\centering\arraybackslash}m{1.5cm}|>{\centering\arraybackslash}m{1.5cm}|>{\centering\arraybackslash}m{1.5cm}|>{\centering\arraybackslash}m{1.5cm}|>{\centering\arraybackslash}m{1.5cm}|}
\toprule[0.8pt]
Metrics 1 & Precision & Recall & F1 & Range-F1 & AUC-ROC & AUC-PR & VUS-ROC & VUS-PR \\
\midrule

Forecast-based & 0.591 & \textbf{0.628} & 0.496 & 0.540 & 0.834 & 0.658 & 0.721 & 0.559  \\ 
-based & 0.683 & 0.571 & 0.502 & 0.558 & 0.837 & 0.699 & 0.715 & 0.591  \\ 
Statistical-model-based & 0.582 & 0.466 & 0.366 & 0.435 & 0.752 & 0.581 & 0.631 & 0.502  \\ 
Proximity-based & \textbf{0.730} & 0.591 & \textbf{0.535} & \textbf{0.577} & \textbf{0.858} & \textbf{0.740} & \textbf{0.752} & \textbf{0.646}  \\ 

\bottomrule[1pt]
\end{tabular}
}
\end{table}

%%%%%%%%%%%%%%%%%%%%%%%%%%%%%%%%
\begin{table}[htbp]
\renewcommand\arraystretch{1.1}
\setlength\tabcolsep{3.0pt}
\centering
\caption{Average value of eight representative accuracy evaluation measures for four types of algorithms when KNC$>$10}
\scalebox{0.95}{
\fontsize{6pt}{4.0pt}\selectfont
\begin{tabular}{|>{\centering\arraybackslash}m{2.5cm}||>{\centering\arraybackslash}m{1.5cm}|>{\centering\arraybackslash}m{1.5cm}|>{\centering\arraybackslash}m{1.5cm}|>{\centering\arraybackslash}m{1.5cm}|>{\centering\arraybackslash}m{1.5cm}|>{\centering\arraybackslash}m{1.5cm}|>{\centering\arraybackslash}m{1.5cm}|>{\centering\arraybackslash}m{1.5cm}|}
\toprule[0.8pt]
Metrics 1 & Precision & Recall & F1 & Range-F1 & AUC-ROC & AUC-PR & VUS-ROC & VUS-PR \\
\midrule

Forecast-based & 0.719 & 0.886 & 0.704 & 0.701 & 0.943 & 0.823 & 0.849 & 0.729  \\ 
-based & 0.751 & 0.783 & 0.680 & 0.682 & 0.917 & 0.840 & 0.815 & 0.735  \\ 
Statistical-model-based & 0.748 & 0.645 & 0.495 & 0.570 & 0.839 & 0.665 & 0.717 & 0.581  \\ 
Proximity-based & 0.758 & 0.801 & 0.704 & 0.722 & 0.925 & 0.854 & 0.840 & 0.747  \\ 

\bottomrule[1pt]
\end{tabular}
}
\end{table}

%%%%%%%%%%%%%%%%%%%%%%%%%%%%%%%%%%%%%%%%%%%%%%%%%%%%%%%%%%%%%%%%

\begin{table}[htbp]
\renewcommand\arraystretch{1.1}
\setlength\tabcolsep{3.0pt}
\centering
\caption{Average value of eight representative accuracy evaluation measures for four types of algorithms when 5$<$KNC$<$10.}
\scalebox{0.95}{
\fontsize{6pt}{4.0pt}\selectfont
\begin{tabular}{|>{\centering\arraybackslash}m{2.5cm}||>{\centering\arraybackslash}m{1.5cm}|>{\centering\arraybackslash}m{1.5cm}|>{\centering\arraybackslash}m{1.5cm}|>{\centering\arraybackslash}m{1.5cm}|>{\centering\arraybackslash}m{1.5cm}|>{\centering\arraybackslash}m{1.5cm}|>{\centering\arraybackslash}m{1.5cm}|>{\centering\arraybackslash}m{1.5cm}|}
\toprule[0.8pt]
Metrics 1 & Precision & Recall & F1 & Range-F1 & AUC-ROC & AUC-PR & VUS-ROC & VUS-PR \\
\midrule

Forecast-based & 0.712 & 0.861 & 0.645 & 0.668 & 0.932 & 0.794 & 0.796 & 0.658  \\ 
-based & 0.741 & 0.789 & 0.640 & 0.643 & 0.933 & 0.799 & 0.768 & 0.651  \\ 
Statistical-model-based & 0.645 & 0.616 & 0.474 & 0.500 & 0.846 & 0.628 & 0.679 & 0.490  \\ 
Proximity-based & 0.724 & 0.756 & 0.652 & 0.704 & 0.936 & 0.828 & 0.798 & 0.685  \\ 

\bottomrule[1pt]
\end{tabular}
}
\end{table}

%%%%%%%%%%%%%%%%%%%%%%%%%%%%%%%%%%%%%%%%%%%%%%%%%%%%%%%%%%%%%%%%
\begin{table}[htbp]
\renewcommand\arraystretch{1.1}
\setlength\tabcolsep{3.0pt}
\centering
\caption{Average value of eight representative accuracy evaluation measures for four types of algorithms when 2$<$KNC$<$5}
\scalebox{0.95}{
\fontsize{6pt}{4.0pt}\selectfont
\begin{tabular}{|>{\centering\arraybackslash}m{2.5cm}||>{\centering\arraybackslash}m{1.5cm}|>{\centering\arraybackslash}m{1.5cm}|>{\centering\arraybackslash}m{1.5cm}|>{\centering\arraybackslash}m{1.5cm}|>{\centering\arraybackslash}m{1.5cm}|>{\centering\arraybackslash}m{1.5cm}|>{\centering\arraybackslash}m{1.5cm}|>{\centering\arraybackslash}m{1.5cm}|}
\toprule[0.8pt]
Metrics 1 & Precision & Recall & F1 & Range-F1 & AUC-ROC & AUC-PR & VUS-ROC & VUS-PR \\
\midrule

Forecast-based & 0.653 & 0.499 & 0.432 & 0.523 & 0.790 & 0.597 & 0.665 & 0.497  \\ 
-based & 0.693 & 0.436 & 0.401 & 0.518 & 0.785 & 0.622 & 0.673 & 0.540  \\ 
Statistical-model-based & 0.509 & 0.261 & 0.227 & 0.340 & 0.701 & 0.519 & 0.579 & 0.486  \\ 
Proximity-based & 0.624 & 0.469 & 0.453 & 0.488 & 0.808 & 0.660 & 0.715 & 0.594  \\ 

\bottomrule[1pt]
\end{tabular}
}
\end{table}

%%%%%%%%%%%%%%%%%%%%%%%%%%%%%%%%%%%%%%%%%%%%%%%%%%%%%%%%%%%%%%%%
\begin{table}[htbp]
\renewcommand\arraystretch{1.1}
\setlength\tabcolsep{3.0pt}
\centering
\caption{Average value of eight representative accuracy evaluation measures for four types of algorithms when 1$<$KNC$<$2}
\scalebox{0.95}{
\fontsize{6pt}{4.0pt}\selectfont
\begin{tabular}{|>{\centering\arraybackslash}m{2.5cm}||>{\centering\arraybackslash}m{1.5cm}|>{\centering\arraybackslash}m{1.5cm}|>{\centering\arraybackslash}m{1.5cm}|>{\centering\arraybackslash}m{1.5cm}|>{\centering\arraybackslash}m{1.5cm}|>{\centering\arraybackslash}m{1.5cm}|>{\centering\arraybackslash}m{1.5cm}|>{\centering\arraybackslash}m{1.5cm}|}
\toprule[0.8pt]
Metrics 1 & Precision & Recall & F1 & Range-F1 & AUC-ROC & AUC-PR & VUS-ROC & VUS-PR \\
\midrule

Forecast-based & 0.524 & 0.299 & 0.273 & 0.384 & 0.714 & 0.545 & 0.589 & 0.441  \\ 
-based & 0.579 & 0.339 & 0.289 & 0.383 & 0.719 & 0.546 & 0.597 & 0.466  \\ 
Statistical-model-based & 0.450 & 0.258 & 0.200 & 0.357 & 0.670 & 0.502 & 0.550 & 0.420  \\ 
Proximity-based & 0.571 & 0.380 & 0.321 & 0.393 & 0.758 & 0.610 & 0.617 & 0.516  \\ 

\bottomrule[1pt]
\end{tabular}
}
\end{table}

%%%%%%%%%%%%%%%%%%%%%%%%%%%%%%%%%%%%%%%%%%%%%%%%%%%%%%%%%%%%%%%%
\begin{table}[htbp]
\renewcommand\arraystretch{1.1}
\setlength\tabcolsep{3.0pt}
\centering
\caption{Average value of eight representative accuracy evaluation measures for four types of algorithms when KNC$<$1}
\scalebox{0.95}{
\fontsize{6pt}{4.0pt}\selectfont
\begin{tabular}{|>{\centering\arraybackslash}m{2.5cm}||>{\centering\arraybackslash}m{1.5cm}|>{\centering\arraybackslash}m{1.5cm}|>{\centering\arraybackslash}m{1.5cm}|>{\centering\arraybackslash}m{1.5cm}|>{\centering\arraybackslash}m{1.5cm}|>{\centering\arraybackslash}m{1.5cm}|>{\centering\arraybackslash}m{1.5cm}|>{\centering\arraybackslash}m{1.5cm}|}
\toprule[0.8pt]
Metrics 1 & Precision & Recall & F1 & Range-F1 & AUC-ROC & AUC-PR & VUS-ROC & VUS-PR \\
\midrule

Forecast-based & 0.920 & 0.431 & 0.418 & 0.570 & 0.697 & 0.653 & 0.645 & 0.623  \\ 
-based & 1.000 & 0.322 & 0.373 & 0.521 & 0.688 & 0.780 & 0.624 & 0.731  \\ 
Statistical-model-based & 0.500 & 0.275 & 0.354 & 0.408 & 0.668 & 0.702 & 0.643 & 0.675  \\ 
Proximity-based & 1.000 & 0.299 & 0.351 & 0.575 & 0.705 & 0.761 & 0.646 & 0.711  \\ 

\bottomrule[1pt]
\end{tabular}
}
\end{table}

\begin{figure}[h]
    \centering
    \subfigure[Algorithms ranks under eight representative accuracy evaluation measures when KNC$<$1]{\includegraphics[width=0.48\textwidth]{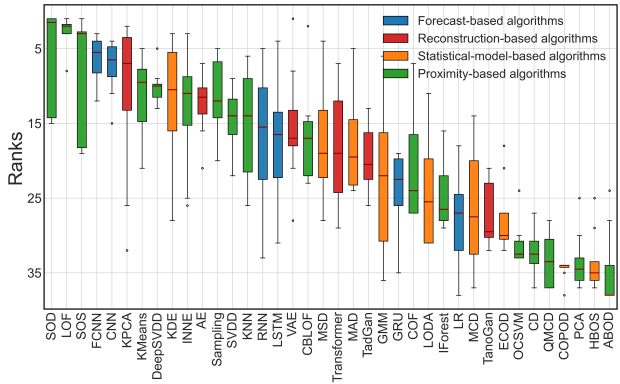}}
    \subfigure[Algorithms ranks under eight representative accuracy evaluation measures when 1$<$KNC$<$2]  
    {\includegraphics[width=0.48\textwidth]{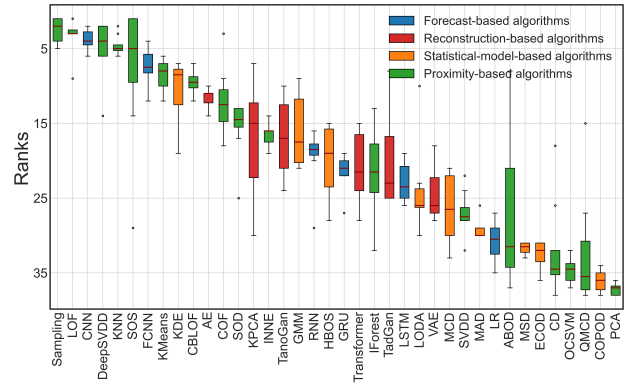}}
    
    \subfigure[Algorithms ranks under eight representative accuracy evaluation measures when 2$<$KNC$<$5]{\includegraphics[width=0.48\textwidth]{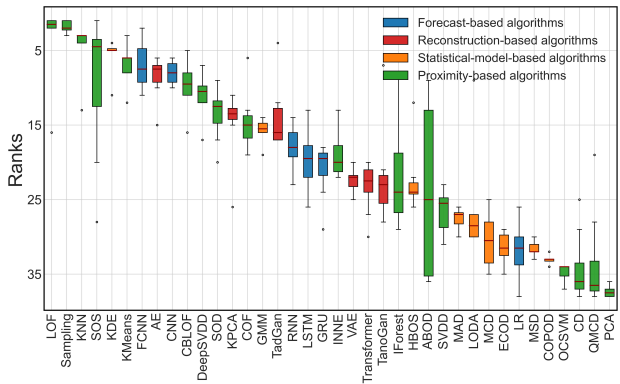}}
    \subfigure[Algorithms ranks under eight representative accuracy evaluation measures when 5$<$KNC$<$10] 
    {\includegraphics[width=0.48\textwidth]{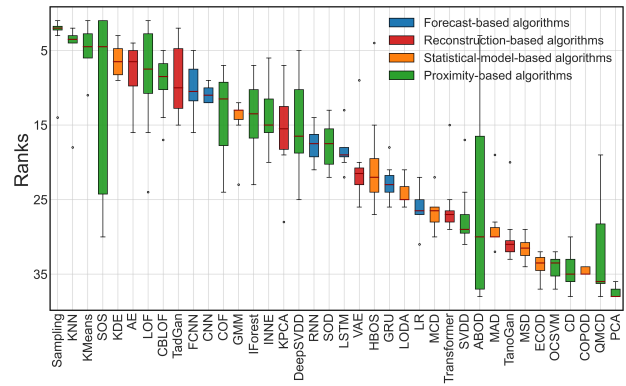}}

    \subfigure[Algorithms ranks under eight representative accuracy evaluation measures when 10$<$KNC]{\includegraphics[width=0.48\textwidth]{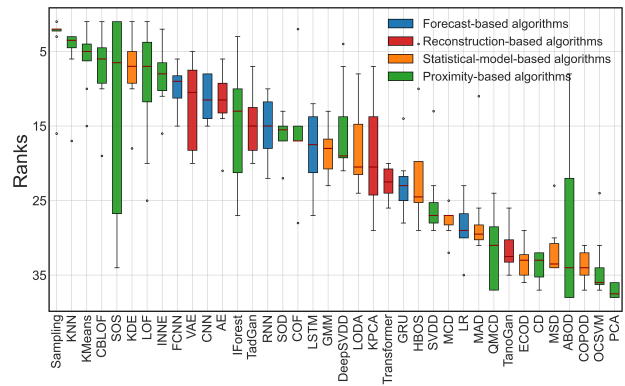}}

    \caption{Performance of algorithms under eight representative accuracy evaluation measures and different KNC.
    % \label{fig:isw2}
    }
\end{figure}

\end{document}